# Dienstplanerstellung in Krankenhäusern mittels genetischer Algorithmen

Diplomarbeit

vorgelegt am

Lehrstuhl für ABWL und Unternehmensforschung
Prof. Dr. Christoph Schneeweiß
Universität Mannheim

von

cand. rer. oec.

Uwe Aickelin

School of Computer Science
University of Nottingham
NG8 1BB   UK
uxa@cs.nott.ac.uk

Dezember 1996

# Inhaltsverzeichnis

# Abbildungsverzeichnis

# Tabellenverzeichnis

# 1 Einführung in die Problematik des Nurse Scheduling

## 1.1 Allgemeine Einführung

Der Gesundheitssektor ist in den neunziger Jahren durch die von der Politik festgesetzten Sparmaßnahmen in einer schwierigen Lage. Dies gilt sowohl für den Sektor als Ganzes, als auch für die Krankenhäuser, die durch den zusätzlichen Mangel an Fachpersonal besonderen Herausforderungen gegenüberstehen. Die zentrale Komponente der Gesundheitsversorgung in Krankenhäusern ist das Personal. Die Personalkosten machen gleichzeitig mehr als die Hälfte der Krankenhausbudgets aus.[1] Um die Situation der Krankenhäuser zu verbessern, ist daher der Personalbereich von besonderem Interesse.

Bei einer Fluktuation von bis zu 200% pro Jahr in manchen Krankenhäusern ist der als Pflegenotstand bezeichnete Mangel an qualifizierten Krankenschwestern auf viele Gründe zurückzuführen.[2] Neben der Überforderung durch zuwenig Personal und der relativ schlechten Bezahlung,[3] sind die unpopulären Schichtpläne,[4] die oft die persönlichen Interessen der Krankenschwestern nicht berücksichtigen, ein wichtiger Grund für die Unattraktivität des Berufes.

Verbesserungen der heute angewandten kurzfristigen Personaleinsatzplanung,[5] könnten sich somit in zweierlei Hinsicht sehr positiv auf die Situation der Krankenhäuser auswirken. Einerseits würden die Personalresourcen besser ausgenutzt und die angespannten Finanzen damit entlastet und andererseits wäre es durch die Einbindung von Wünschen des Personals möglich, die Attraktivität des Berufes einer Krankenschwester zu erhöhen und damit den Pflegenotstand zu mildern.

Eine solche kurzfristige Personaleinsatzplanung in Krankenhäusern muß aufgrund der Spezialisierung der Krankenschwestern auf Stationsbasis erfolgen. Die genaue Zielsetzung einer solchen Planung variiert von Station zu Station. Das grundlegende Ziel ist es, einen Einsatzplan zu finden, der zu jeder Tageszeit die Anwesenheit einer ausreichenden Anzahl entsprechend qualifizierter Krankenschwestern sicherstellt. Zusätzlich sollten die individuellen Wünsche der Krankenschwestern möglichst gut berücksichtigt werden.

---

[1] Vgl. Grütz, 1982, S. 88.
[2] Vgl. Easton, 1992, S. 159.
[3] Vgl. Worthington, 1988, S. 174-182.
[4] Vgl. Sitferd, 1992, S. 233-246.
[5] In der anglo-amerikanischen Literatur bezeichnet Nurse Scheduling den Problemkreis der Personaleinsatzplanung in Krankenhäusern.

Verschiedene Qualifikationsstufen, Teilzeitarbeitsverträge und ein schwankender Bedarf während des Tagesablaufs bedeuten, daß diese Aufgabe analytisch keinesfalls einfach zu lösen ist. Erschwert wird das Problem in der Praxis noch durch Urlaubstage, Krankheit und den Wunsch, unbeliebte Arbeitszeiten, wie nachts oder am Wochenende, möglichst gerecht unter der gesamten Belegschaft zu verteilen.

## 1.2    Fallbeispiel

### 1.2.1   Die Aufgabe

Diese Arbeit stützt sich auf Datenmaterial eines großen britischen Krankenhauses, das eine flexible, kurzfristige Einsatzplanung wünscht. Um für eine gute Arbeitsmoral zu sorgen, ist es für die Krankenhausleitung wichtig, daß persönliche Wünsche berücksichtigt werden können und die Einsatzpläne hinsichtlich unpopulärer Arbeitszeiten vom Personal als fair betrachtet werden. Außerdem muß die Einsatzplanung in der Lage sein, ein sich kurzfristig änderndes Personalangebot durch Urlaub, Krankheit, Schulungen etc. zu berücksichtigen.

Um dies zu erreichen sollen mit wöchentlichem Planungshorizont auf Stationsbasis mehrere gute Schichtpläne erstellt werden, die den Bedarf an Krankenschwestern erfüllen. Aus diesen wählt die Stationsleitung den aus ihrer Sicht optimalen aus. Bisher wurde das Problem manuell gelöst, was nicht immer für eine ausgewogene Bedarfsdeckung und die Berücksichtigung aller Wünsche sorgte und außerdem einen beträchtlichen Zeitaufwand je Planungslauf verursachte.

### 1.2.2   Das mathematische Modell

Da das Krankenhaus in einem drei Schicht System (einer Früh-, einer Spät- und einer längeren Nachtschicht) arbeitet und die Krankenschwestern in einer Woche entweder nur tagsüber oder nur nachts arbeiten dürfen,[6] läßt sich das vollständige Problem in drei einzeln zu lösende
Teilprobleme aufspalten:

1. Sicherstellen, daß für die nächste Woche genügend Personal zur Verfügung steht und
   notfalls Springer eingesetzt werden.
2. Festlegen, an welchen Tagen bzw. Nächten eine Krankenschwester arbeiten soll.
3. Aufspalten der Tag-Schichten in Früh- und Spätschichten.

Für den ersten und letzten Schritt existiert bereits eine Software, die das Problem
zufriedenstellend löst.[7] Die durch Relaxation davon verbleibende Aufgabe ist es daher, den
Krankenschwestern die richtigen Tage bzw. Nächte zuzuordnen. Den Wocheneinsatzplan einer
Krankenschwester kann man somit als 14-stelligen binären Vektor darstellen. Die ersten sieben
Elemente des Vektors sollen hierbei die Tage von Sonntag bis Samstag, die restlichen sieben die
entsprechenden Nächte repräsentieren, wobei

$$x_i = \begin{cases} 1 & \text{an Tag bzw. Nacht i wird gearbeitet} \\ 0 & \text{sonst} \end{cases} \qquad \forall\ i = 1..14$$

Ein solcher Vektor soll im folgenden Schichtmuster genannt werden. Aufgrund der Forderung,
daß Krankenschwestern in einer Woche nur tagsüber oder nur nachts arbeiten dürfen, gibt es
genau zwei Arten von Schichtmustern: Schichtmuster, bei denen nur tagsüber und
Schichtmuster, bei denen nur nachts gearbeitet wird. Ein Schichtmuster, bei dem z.B. tagsüber
an Sonntag, Montag und Mittwoch gearbeitet wird hat dann folgendes Aussehen:
(11010000000000). Wird nachts von Dienstag bis Samstag gearbeitet ist (0000000001111) das
entsprechende Schichtmuster.

Die Qualität eines Schichtmusters für eine Krankenschwester hängt von mehreren Faktoren ab:
die grundsätzliche Güte des Musters,[8] die Wünsche der Krankenschwester[9] und das in der
Woche zuvor gearbeitete Schichtmuster. Jedem Krankenschwester-Schichtmuster-Paar können
somit bestimmte Strafkosten zugeteilt werden, wobei eine optimale Kombination Strafkosten
von Null erhält. Diese Strafkosten sind keinesfalls als pagatorische Kosten aufzufassen. Sie sind
vielmehr Lenkkosten, die die Lösung des Formalmodells der richtigen Schichtmuster-

---

[6]  Dies wurde von der Krankenhausleitung so festgelegt.

[7]  Vgl. Dowsland, 1996a, S. 1-21.

[8]  Wurde von der Krankenhausleitung allgemein festgelegt, z.B. ob die freien Tage bzw. Arbeitstage an
     einem Stück sind. Für ein Beispiel siehe Anhang A4.

[9]  Die Wünsche der Krankenschwestern werden in einem Kalender, der in der Station ausliegt gesammelt
     und von der Stationsleitung aufgearbeitet. Für eine ausführliche Erklärung siehe Kapitel 0. Für ein
     Beispiel siehe Anhang A3.

Krankenschwester-Zuordnung derart „lenken" sollen, daß das zugrunde liegende Realproblem des Nurse Scheduling möglichst gut gelöst wird.[10]

Stehen die Strafkosten fest,[11] so besteht das Problem darin, diese unter den zwingenden Nebenbedingungen der Bedarfsdeckung an allen Tagen bzw. Nächten und auf allen Qualifikationsstufen zu minimieren.[12] Dies führt zu folgenden mathematischen Modell:[13]

Indizes:

$i \quad = 1,...,n$ Krankenschwesterindex

$j \quad = 1,...,m$ Schichtmusterindex

$k \quad = 1,...,14$ Tag- bzw. Nachtindex

$s \quad = 1,...,p$ Qualifikationsstufenindex

Entscheidungsvariablen:

$$x_{ij} = \begin{cases} 1 & \text{Krankenschwester i arbeitet Schichtmuster j} \\ 0 & \text{sonst} \end{cases}$$

Modellparameter:

$n \quad$ = Anzahl der Krankenschwestern der betrachteten Station

$m \quad$ = Anzahl aller theoretisch möglichen Schichtmuster

$p \quad$ = Anzahl Qualifikationsstufen der Krankenschwestern

$$a_{jk} = \begin{cases} 1 & \text{Schichtmuster j deckt Tag bzw. Nacht k ab} \\ 0 & \text{sonst} \end{cases}$$

$$q_{is} = \begin{cases} 1 & \text{Krankenschwester i ist von Qualifikationsstufe s oder höher} \\ 0 & \text{sonst} \end{cases}$$

$p_{ij} \quad$ = mit Krankenschwester i und Schichtmuster j verbundenen Strafkosten

$W_i \quad$ = Wochenarbeitszeit von Krankenschwester i

$R_{ks} \quad$ = Mindestbedarf an Krankenschwestern der Qualifikationsstufe s an Tag bzw. Nacht k

---

[10] Zum allgemeinen Prozeß der Modellbildung und zu den Lenkkosten als Modellparameter vgl. Schneeweiß, 1992, S. 74-84.

[11] Für weitere Beispiele siehe Anhang A5.

[12] Für ein Beispiel eines Wochenbedarfes siehe Anhang A1.

[13] Zur Formulierung vergleiche auch Miller, 1976, S. 857-870, der schon sehr früh ein ähnliches Modell aufstellte und Brusco, 1993, S.175-186.

Zielfunktion: Minimierung der Strafkosten:

$$\sum_{i=1}^{n}\sum_{j=1}^{m}p_{ij}x_{ij} \quad \rightarrow \quad \min! \qquad\qquad\qquad ( \ 0\text{-}1 \ )$$

Restriktionen:

1. Jede Krankenschwester arbeitet genau ein Schichtmuster:

$$\sum_{j=1}^{m}x_{ij} \quad = \quad 1 \qquad\qquad \forall i \qquad\qquad\qquad ( \ 0\text{-}2 \ )$$

2. Das gearbeitete Schichtmuster entspricht der Wochenarbeitszeit der Krankenschwester:

$$\sum_{j=1}^{m}\sum_{k=1}^{14}a_{jk}\,x_{ij} \ = \ W_{i} \qquad\qquad \forall i \qquad\qquad\qquad ( \ 0\text{-}3 \ )$$

3. Für jeden Tag und für jede Nacht stehen genügend ausreichend qualifizierte Krankenschwestern zur Verfügung:

$$\sum_{s=1}^{p}\sum_{j=1}^{m}\sum_{i=1}^{n}q_{is}\,a_{jk}\,x_{ij} \ \geq \ R_{ks} \qquad\qquad \forall k \qquad\qquad\qquad ( \ 0\text{-}4 \ )$$

Die Komplexität des Problems deutet auf eine heuristische Lösungsmethode hin. Dafür spricht auch, daß die Krankenhausleitung mehrere gute und grundsätzlich verschiedene Lösungen wünscht. Dies wird durch den gewissen Grad an Zufall, dem Lösungen von Heuristiken zugrunde liegen, ermöglicht. Außerdem soll die Lösungsmethode schnell und flexibel sein, um wöchentlich auf den PCs der einzelnen Stationen angewendet werden zu können.

### 1.2.3   Das Datenmaterial

Eine typische Station des zu betrachtenden Krankenhauses hat 20 bis 25 Krankenschwestern, die für eine Betreuung rund um die Uhr sorgen müssen. Dazu wird der Tag in drei Schichten aufgeteilt: eine Früh-, eine Spät- und eine längere Nachtschicht. Für die möglichen Arbeitszeiten wurden als Konvention sieben Stufen eingerichtet, wobei feinere Abstufungen durch Pausen und leicht variierte Anfangs- und Endzeiten auf Stationsebene erzielt werden. Durch die längeren Nachtschichten arbeitet demnach eine voll angestellte Krankenschwester entweder fünf Tag- oder vier Nachtschichten. Durch Teilzeitverträge und Urlaubstage gibt es folgende Möglichkeiten der Wochenarbeitszeit:

- Stufe 1: 100% einer Vollzeitstelle entspricht 5 Tag- oder 4 Nachtschichten.

- Stufe 2: 80% einer Vollzeitstelle entspricht 4 Tag- oder 3 Nachtschichten.

- Stufe 3: 60% einer Vollzeitstelle entspricht 3 Tag- oder 3 Nachtschichten.

- Stufe 4: 50% einer Vollzeitstelle entspricht 3 Tag- oder 2 Nachtschichten.

- Stufe 5: 40% einer Vollzeitstelle entspricht 2 Tag- oder 2 Nachtschichten.

- Stufe 6: 20% einer Vollzeitstelle entspricht 1 Tag- oder 1 Nachtschicht.

- Stufe 7: 0% einer Vollzeitstelle entspricht Urlaub oder Krankheit.

Auf diese Weise erhält man für eine Vollzeit arbeitende Krankenschwester 21 mögliche Tag-Schichtmuster und 35 mögliche Nacht-Schichtmuster,[14] für eine 80% arbeitende Krankenschwester entsprechend 35 mögliche Tag-Schichtmuster und 35 mögliche Nacht-Schichtmuster.[15]

Aufgrund ihrer Ausbildung und Erfahrung werden die Krankenschwestern in drei verschiedene Qualifikationsstufen eingeteilt, wobei die am höchsten qualifizierten Krankenschwestern in Stufe eins eingeordnet sind. Für jede Qualifikationsstufe gibt es eine minimale Anzahl an Krankenschwestern, die mindestens pro Schicht anwesend sein muß. Der Bedarf an weniger qualifizierten Krankenschwestern kann durch höher qualifizierte abgedeckt werden.[16]

Die Wünsche der Krankenschwestern nach freien Tagen werden in einem auf der jeweiligen Station ausliegenden Tagebuch gesammelt. Die Stationsleitung vereinheitlicht diese dann auf eine Skala von 0 (Indifferenz für diesen Tag) bis 4 (sehr starker Wunsch, nicht an diesem Tag zu arbeiten).[17] Die Bewertung der generellen Güte eines Schichtmuster wurde von der Krankenhausleitung aufgrund der Zusammensetzung der freien Tage und der Arbeitstage des Schichtmusters auf einer Skala von 0 (sehr gutes Schichtmuster) bis 4 (sehr schlechtes Schichtmuster) vorgenommen.[18]

Diese generelle Güte wird durch Verknüpfung der Einzelwünsche und den grundsätzlichen Wünschen der Krankenschwestern nach Tag- bzw. Nachtarbeit zu einer vorläufigen Schichtmuster-Krankenschwester Strafkosten-Tabelle zusammengefaßt. Diese wird von der

---

[14] Berechnet als $\binom{7}{5} = 21$ bzw. $\binom{7}{4} = 35$.

[15] Für eine vollständige Liste aller möglichen Schichtmuster siehe Anhang A4.

[16] Für ein Beispiel für typische Wochenarbeitszeiten und Qualifikationen einer Station siehe Anhang A2.

[17] Für ein Beispiel zu den Tageswünschen siehe Anhang A3. Für ein Beispiel zu den grundsätzlichen Wünschen nach Tag- bzw. Nachtarbeit siehe Anhang A2.

[18] Für ein Beispiel siehe Anhang A4.

Stationsleitung nochmals geringfügig überarbeitet, um z.B. das Schichtmuster der Vorwoche einer Krankenschwester zu berücksichtigen und anschließend mit den Krankenschwestern durchgesprochen. Die endgültige Tabelle enthält für jedes Krankenschwester-Schichtmuster-Paar Werte von 0 (sehr gute Kombination) bis 100 (sehr schlechte Kombination).[19]

## 1.2.4 Lösungsansätze

Ein beliebter Ansatz zur Lösung von Personaleinsatzproblemen sind zyklische Einsatzpläne, bei denen jede(r) Angestellte wiederholt das selbe Muster von bestimmten Wocheneinsatzplänen arbeitet.[20] Diese eher im mittelfristigen Planungsbereich anzusiedelnden Ansätze gehen jedoch von einem konstanten Personalstamm aus, wobei jede(r) Mitarbeiter(in) jederzeit verfügbar ist. Da bei diesem Ansatz keine kurzfristigen Änderungen oder Wünsche berücksichtigt werden können, ist er für das zu betrachtende Problem zu unflexibel.

Gute Erfahrungen wurden bisher mit einigen spezifischen Heuristiken gemacht, die jedoch konkret auf einen sehr speziellen Fall zugeschnitten waren und daher auch nicht die geforderte Flexibilität besitzen.[21] Flexibel und erfolgreich bei der Lösung des Nurse Scheduling Problems sind Ansätze mit Simulated Annealing und Tabu Search, auf die an dieser Stelle jedoch nicht weiter eingegangen werden soll.[22]

Ein weiterer vielversprechender Ansatz sind genetische Algorithmen.[23] Diese bisher eher wenig benutzte Klasse von Heuristiken zeigte sich in vielen Fällen Tabu Search und Simulated Annealing in Lösungsqualität und Lösungsgeschwindigkeit überlegen.[24] Genetische Algorithmen wurden außerdem bereits sehr erfolgreich zur Lösung des verwandten Problems der Stundenplanerstellung eingesetzt.[25]

---

[19] Für ein Beispiel einer endgültigen Schichtmuster-Krankenschwester Strafkosten-Tabelle siehe Anhang A5.

[20] Vgl. Schneeweiß, 1996, S. 15-27 und Rosenbloom, 1987, S. 19-23.

[21] Vgl. Nooriafshar, 1995, S. 50-61 und Randhawa, 1993, S. 837-844.

[22] Vgl. Dowsland, 1996a, S. 1-21.

[23] Eine ausführliche Beschreibung von genetischen Algorithmen erfolgt in Kapitel 2.

[24] Vgl. Kurbel, 1995, wo ebenfalls zuerst Simulated Annealing und dann genetische Algorithmen angewendet werden sowie Dowsland, 1996b, S. 550-561 und Heistermann, 1994, S.80ff für weitere Vorteile der genetischen Algorithmen im Vergleich zu Simulated Annealing.

[25] Zum Problematik der Stundenplanerstellung mittels genetischer Algorithmen vgl. Chambers, 1995, S.219ff.

Beides legt die Schlußfolgerung nahe, daß genetische Algorithmen sehr gut zur Lösung des Nurse Scheduling Problems geeignet sind. Das Ziel dieser Diplomarbeit ist es, die Möglichkeiten genetischer Algorithmen im Rahmen des Nurse Scheduling Problems aufzuzeigen.

# 1.3    Weitere Vorgehensweise

Kapitel 0 gibt zunächst ein Einblick in genetische Algorithmen, bevor in Kapitel 0 das eigentliche Programm zur Lösung des Nurse Scheduling Problems vorgestellt wird. Erste Experimente mit dem Programm werden in Kapitel 0 durchgeführt, um einerseits Werte für die verschiedenen genetischen Parameter zu erhalten und andererseits Schwachstellen aufzuzeigen.

Grundsätzlich scheidet eine Formulierung des Nurse Scheduling Problems als lineares Programm durch die Komplexität und die Forderung nach einer flexiblen Lösungsmethode, die zukünftig auch nichtlineare Zielfunktionen verarbeiten kann aus. Zusätzlich steht dem die Forderung nach mehreren guten, grundsätzlich verschiedenen Lösungen, die auf den PCs der einzelnen Stationen erzielt werden können, entgegen.

Trotzdem wird im Rahmen von Kapitel 0 zunächst ein entsprechendes lineares Programm in Anlehnung an das Model aus Kapitel 0 aufgestellt. Dieses wird dann mittels CPLEX, einer kommerziellen Optimierungssoftware, bearbeitet, um Anhaltspunkte über die Qualität der später mittels genetischen Algorithmen erzielten Lösungen zu gewinnen.

Kapitel 0 versucht, die in Kapitel 0 gefundenen Schwachstellen nach und nach durch geeignete Maßnahmen zu beseitigen und letztlich zu einem optimal angepaßten Programm zu gelangen. Im Rahmen diese Kapitels werden auch erste Erweiterungen des vereinfachten Grundmodells vorgeschlagen. Kapitel 0 rundet die Diplomarbeit mit einer Zusammenfassung und einem Ausblick auf mögliche nächste Schritte ab.

# 2    Einführung in genetische Algorithmen

## 2.1    Definitionen

Bevor mit der eigentlichen Einführung in genetische Algorithmen begonnen wird, sollen zunächst die wichtigsten Fachbegriffe definiert werden.

Unter *Element* versteht man im folgenden die Ausprägung einer Entscheidungsvariablen, z.B. welche Schicht arbeitet Krankenschwester i. Ein solches auch als *Einzelmerkmal* bezeichnetes Element wird in der Natur *Gen* genannt. Die Zusammenfassung aller Einzelmerkmale, und damit eine mögliche Lösung, wird als *Individuum* bezeichnet. Dies entspricht einem *Chromosom* in der Natur. Die Gesamtheit aller Individuen, die vom genetischen Algorithmus gleichzeitig bearbeitet werden, nennt man *Bevölkerung*.

Unter einem *Alphabet* wird im folgenden der zulässige Wertebereich eines Elementes verstanden, dies könnte z.B. die Indizes aller theoretisch möglichen Schichtmuster sein. Mit *Kodierung* wird die Umwandlung der Darstellung eines Individuums von dessen natürlichem Alphabet in ein für die genetischen Algorithmen besser zu bearbeitendes bezeichnet, z.B. die Umwandlung von Skalaren in das binäre System.

Eine komplette Schleife eines genetischen Algorithmus wird *Generation* genannt. Die Individuen der aktuellen Generation heißen *Eltern*. Alle neu erzeugten Individuen dieser Generation und damit Kandidaten für die Eltern-Position der nächsten Generation heißen *Kinder*.

Die *Güte* eines Individuums drückt dessen Qualität bezüglich des zu lösenden Problems aus und kann z.B. ein abgewandelter Zielfunktionswert des mathematischen Modells aus Kapitel 0 sein. Sie wird in *Strafkosten* gemessen, die minimiert werden sollen. Als *Selektion* wird die Auswahl derjenigen Eltern einer Generation, die Kinder erzeugen werden, bezeichnet. Sie ist proportional zur Güte der Individuen und soll damit Darwins Gedanken „survival of the fittest" repräsentieren.

Das Kernstück der genetischen Algorithmen ist das sogenannte *Crossover*. Bei diesem Vorgang werden durch *Rekombination* von Eltern Kinder erzeugt. Dazu werden, analog zu der Vereinigung zweier Chromosome in der Natur, die Eltern zunächst an einer als *Crossover-Punkt* bezeichneten Stelle aufgespalten und anschließend überkreuz zu zwei neuen Kindern wieder zusammengefügt. Im Gegensatz zur Natur sind auch mehrere Crossover-Punkte denkbar. Als *uniformes Crossover* wird schließlich das Aufspalten der Eltern in alle Einzelmerkmale und anschließendes zufälliges Zuordnen jedes Elementes zu einem der beiden Kinder bezeichnet.

Die *Mutation* der genetischen Algorithmen ist an den gleichnamigen Vorgang in der Natur angelehnt. Mit kleiner Wahrscheinlichkeit wird dabei ein einzelnes Element eines Individuums zufällig verändert. Es ist ebenfalls möglich, daß Kinder als Klone von Eltern erzeugt werden, dies wird *Reproduktion* genannt.

Welcher der *genetischen Operatoren*, d.h. Crossover, Mutation oder Reproduktion zum Zuge kommt, bleibt dem Zufall zusammen mit den als Parameter vorgegebenen *Crossover-*, *Mutations-* und *Reproduktions-Wahrscheinlichkeiten* überlassen.

Ist eine komplette Generation an Kindern erzeugt, so entscheidet der als *Substitution* bezeichnete Mechanismus, welche der Eltern bzw. Kinder die nächste Generation erreichen. Eine Strategie, bei der auf jeden Fall die Besten der Eltern überleben wird *Elitismus* genannt.

## 2.2    Allgemeine Einführung

Genetische Algorithmen sind stochastische Suchalgorithmen, die auf die natürlichen Selektions- und Mutationsmechanismen aufbauen.[26] Frei nach Darwins „survival of the fittest" werden von einer Anfangsbevölkerung, die aus relativ schlechten Lösungen besteht, über mehrere Generationen hinweg bessere Kinder bzw. Lösungen erzeugt.

Dies erfolgt durch geeignete Rekombination von alten Teillösungen, dem sogenannten Crossover. Ein systematisches Vorgehen wird dabei mit einem zufälligen verbunden: Tendenziell werden bessere Lösungen häufiger zur Zeugung von Kindern ausgewählt. Die genaue Auswahl bleibt aber letztendlich dem Zufall überlassen. Weitere Möglichkeiten neue Lösungen zu erzeugen sind die Mutation und die Rekombination. Die Mutation dient dazu, verloren gegangene Teillösungen eventuell wieder einzuführen. Die Reproduktion entspricht dem Klonen einer Lösung.

---

[26] Vgl. Goldberg, 1989, S. 1f und Michalewicz, 1992, S.13-18.

Um dieses Vorgehen zu ermöglichen, werden Lösungen analog der in der Natur vorkommenden Chromosomen dargestellt. Dies könnte z.B. ein Vektor der Länge L sein, wobei jedes Element des Vektors das Schichtmuster einer der L Krankenschwestern beschreibt. Faßt man eine Anzahl solcher Chromosomen zusammen, so erhält man die oben erwähnte Bevölkerung.

Von klassischen Optimierungsmethoden unterscheiden sich genetische Algorithmen somit in folgenden Punkten:[27]

- Sie arbeiten nicht mit den Variablen selbst, sondern mit einem als Vektor dargestellten Variablen-Chromosom.
- Sie starten ihre Suche nicht von einem Punkt, sondern von einer Menge von Punkten aus.
- Sie benutzen nur die Güte einer Lösung als Information, keine Erweiterungen oder Ableitungen daraus wie z.B. Gradienten.
- Sie gelangen nicht deterministisch, sondern stochastisch, von einer Lösung zu einer anderen.

Trotz einer Vielzahl von erfolgreichen Anwendungen sind genetische Algorithmen noch nicht sehr weit verbreitet innerhalb des Operational Research.[28] Gründe hierfür sind die Schwierigkeit, eine geeignete Kodierung für Lösungen und gute Parameterwerte für die genetischen Operatoren zu finden und Nebenbedingungen in geeigneter Weise einzubauen.[29]

Während eine geeignete Kodierung im vorliegenden Problem relativ einfach zu finden sein wird, erweist sich die fehlende Einbindungsmöglichkeit von Nebenbedingungen als das zentrale Problem, dem ein Großteil des 0. Kapitels gewidmet ist. Mit der Suche nach guten Parameterwerten wird sich Kapitel 0 befassen, wobei die Schwierigkeiten hier in deren gegenseitigen Beeinflussung liegen.[30]

---

[27] Vgl. Goldberg, 1989, S. 7.
[28] Für eine Auflistung siehe Bäck, 1993 in dem über 200 Anwendungen aus 16 verschiedenen Bereichen aufgeführt werden und Ross, 1994.
[29] Vgl. Dowsland, 1996b, S. 552-554.
[30] Vgl. z.B. zum Zusammenhang zwischen Bevölkerungsgröße und Art des Crossovers Schwefel, 1991, S. 38ff.

## 2.3    Aufbau eines genetischen Algorithmus

### 2.3.1   Überblick

Im folgenden soll zunächst ein kurzer Überblick über den Ablauf eines konventionellen genetischen Algorithmus gegeben werden, der dann in den folgenden Abschnitten konkretisiert wird (vgl. Abb. 0-1).

In der Initialisierungsphase werden die nötigen Parameterwerte wie Bevölkerungsgröße N, Mutations- p(M), Crossover- p(C) und Reproduktionswahrscheinlichkeit p(R) sowie die Anzahl an Generationen über die der Algorithmus laufen soll, festgelegt. Zusätzlich wird in der Initialisierungsphase eine zufällige Ausgangsbevölkerung erzeugt.

Nachdem die Güte für alle Individuen berechnet ist, wird die Bevölkerung der nächsten Generation erzeugt. Eine Zufallszahl zusammen mit den entsprechenden Wahrscheinlichkeiten legt fest, ob eine Mutation, eine Reproduktion oder ein Crossover stattfindet, welches anschließend ausgeführt wird und neue Individuen erzeugt.

Dieser Schritt wird so lange wiederholt, bis die neue Bevölkerung aufgefüllt ist. In der anschließenden Substitutionsphase wird bestimmt, welche Individuen dieser Generation, d.h. der alten Eltern und deren Kindern, die Eltern der nächsten Generation sein werden. Ist der Algorithmus über die geforderte Anzahl von Generationen gelaufen, bricht er ab und das beste Individuum der letzten Generation wird als Lösung ausgegeben.

### 2.3.2   Das Kodierungsproblem

Bevor mit der Anwendung des genetischen Algorithmus gestartet werden kann, muß eine geeignete Kodierung für die Lösungen bzw. Individuen gefunden werden, die den Anforderungen der genetischen Operatoren Rechnung trägt. Dafür wurde früher ein ausschließlich binäres Alphabet verwendet. Da dies nicht immer ohne weiteres möglich und auch nicht immer von Vorteil ist,[31] werden heute meist beliebige Alphabete benutzt.[32] Zur Erläuterung diene folgendes Beispiel:

---

[31] Vgl. Fogarty, 1994, S. 38ff.
[32] Zur Vereinfachung sind die Beispiele dieses Kapitels in einem binären Alphabet gehalten. Die tatsächliche Kodierung des Nurse Scheduling umfaßt dann ein größeres Alphabet.

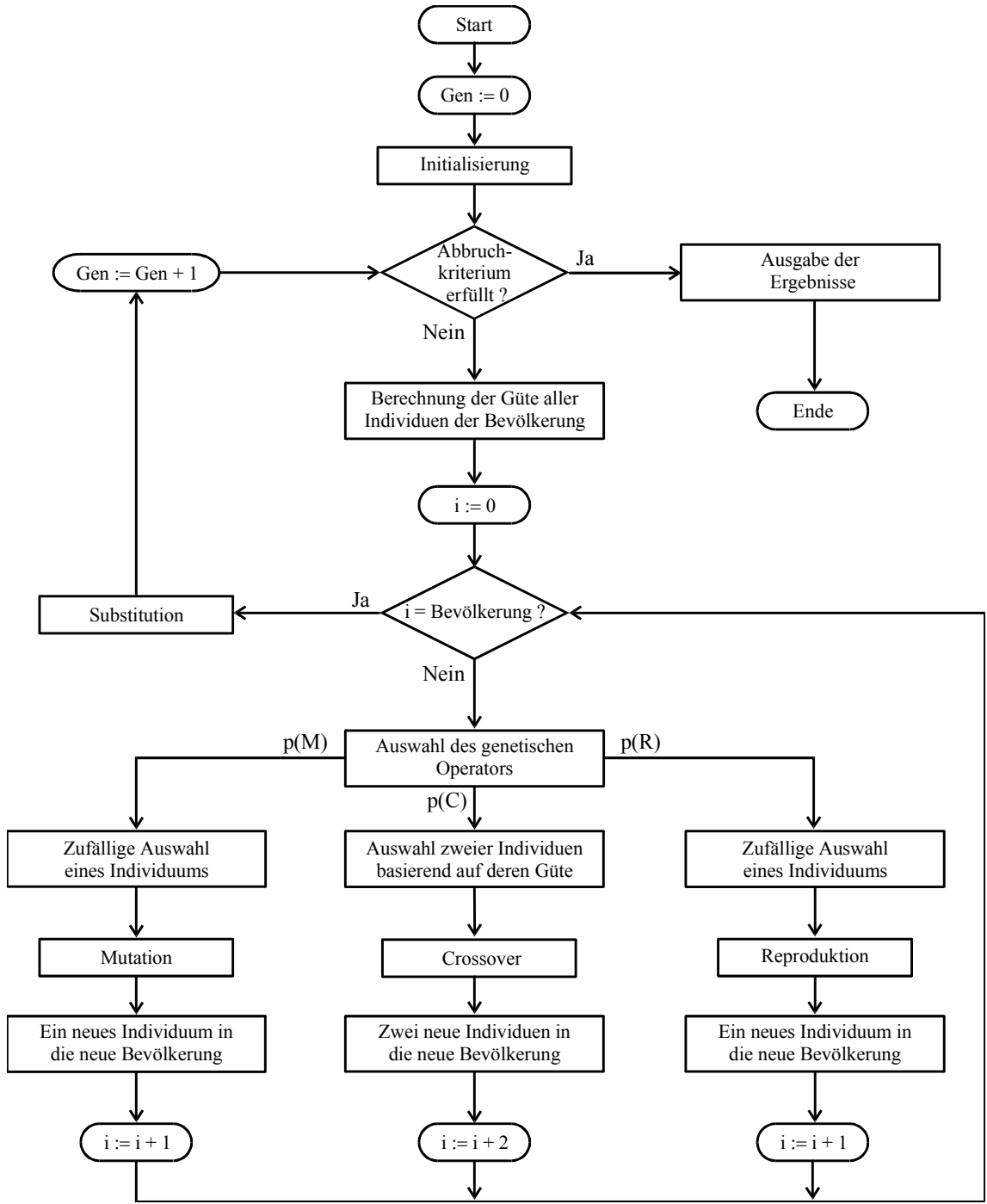

**Abb. 0-1:** Systematischer Aufbau eines konventionellen genetischen Algorithmus.[33]

Es soll die Funktion $f(x) = x^2 - 20x + 4$ minimiert werden. Zwei Startlösungen seien $x_1 = 2$ und $x_2 = 5$. Um nun die genetischen Operatoren, wie z.B. das Crossover anzuwenden, ist eine

---

[33]  Vgl. Koza, 1993, S. 29.

Kodierung der Lösungen im Zehnersystem wenig geeignet, da sich Skalare nicht spalten lassen. Wandelt man beide Lösungen in ihre entsprechenden Binärwerte um, also in $x_1 = (10)$ und $x_2 = (101)$, und wählt als Obergrenze für zu betrachtende x-Werte 128 bzw. als Untergrenze 0, so lautet die vollständige binäre Kodierung $x_1 = (00000010)$ und $x_2 = (00000101)$. Mit dieser Kodierung ist es jetzt möglich, Individuen aufzuspalten und wieder neu zusammenzusetzen.

Grundsätzlich muß die Kodierung so erfolgen, daß sie eindeutig ist und die Güte der Individuen einfach berechnet werden kann. Zusätzlich sollte man die Arbeitsweise von Mutation, Selektion, Reproduktion und Crossover berücksichtigen, um die beste problemspezifische Repräsentation zu finden. So ist es z.B. möglich, einen Teil der Nebenbedingungen bereits implizit in der Kodierung zu verankern.[34]

### 2.3.3  Initialisierung

Die Schwierigkeiten bei der Initialisierung liegen weniger bei der Erzeugung einer zufälligen Anfangsbevölkerung als vielmehr bei der Wahl von geeigneten Parameterwerten. Obwohl versucht worden ist, problemunabhängige Regeln bzw. Heuristiken zu finden,[35] herrscht in der Literatur Einigkeit darüber, daß sämtliche Parameterwerte nur problemspezifisch festlegbar sind.[36] Dies wird dazu führen, daß ein großer Teil des 0. Kapitels der Bestimmung optimaler Parameterwerte für das Nurse Scheduling Problem gewidmet ist. Im folgenden sind zunächst einige in der Literatur vorgeschlagene Heuristiken und deren Schwachpunkte aufgeführt.

Größe der Bevölkerung (N):

Bei der Wahl der Größe der Bevölkerung ist zu bedenken, daß die Laufzeit des Algorithmus mit ihrer Erhöhung wesentlich ansteigt, da entsprechend mehr Kinder erzeugt und deren Güte berechnet werden muß. Andererseits nimmt die Qualität der Lösungen im Allgemeinen mit einer größeren Bevölkerung und einer damit verbundenen größeren Vielfalt zu.[37] Typische Werte liegen zwischen N = 50 und N = 1000.

---

[34] Vgl. hierzu Kapitel 0, indem ausführlich auf die problemspezifische Kodierung des Nurse Scheduling eingegangen wird.
[35] Vgl. Yao, 1995, S. 49-60.
[36] Vgl. Koza, 1993, S. 114ff, Goldberg, 1989, S.71, Chambers, 1995, S.112f und Bäck, 1996,S. 113f.
[37] Vgl. Männer, 1994, S. 131.

Eine in der Literatur anhand von Beispielen durch Interpolation gefundene Heuristik für eine optimale Bevölkerungsgröße bei konstant vorgegebener Rechenzeit ist:

$$N = 1 + (10{,}28 - 12{,}07S + 7{,}30S^2)\sqrt{L} \cdot \ln(L) \cdot \left( \frac{1}{\sqrt{p_0}} - 1 \right) \qquad\qquad ( \ 0\text{-}1 \ )^{38}$$

- L sei die Anzahl der Elemente eines Individuums, z.B. L = 20.
- $p_0$ sei die Wahrscheinlichkeit, daß ein Element in der Startbevölkerung bereits den optimalen Wert aufweist, z.B. $p_0$ = 0,02.
- S drückt die Intensität der Selektion aus und ist ein Maß dafür, wie schnell der genetische Algorithmus auf ein lokales Optimum konvergiert. S hängt vor allem von der verwendeten Substitutions-Strategie ab und kann nur geschätzt werden, was die Verwendung der Heuristik sehr problematisch macht. Ein größeres S bedeutet dabei eine schnellere Konvergenz. Typische Werte liegen der Literaturquelle nach zwischen S = 0,8 und S = 1,4, wobei je nach verwendeter Selektion und Substitution eine wesentlich größere Spanne möglich ist.[38]

Mutationswahrscheinlichkeit p(M):

Üblicherweise liegt p(M) im Bereich zwischen p(M) = 0,001 und p(M) = 0,05.[39] Zwei durch Interpolation anhand von Beispielen gefundene Heuristiken sind:

$$p(M) = \frac{1}{L} \qquad\qquad\qquad\qquad ( \ 0\text{-}2 \ )^{40}$$

$$p(M) = \frac{1{,}75}{N\sqrt{L}} \qquad\qquad\qquad\qquad ( \ 0\text{-}3 \ )^{41}$$

Bei einer Bevölkerungsgröße von N = 1000 und für L = 20 Elemente pro Individuum ergibt dies p(M) = 0,05 für ( 2 ) und p(M) = 0,0004 für ( 3 ). An dieser Widersprüchlichkeit sieht man nochmals die Problematik der Allgemeingültigkeit solcher anhand von speziellen Beispielen gefundener Heuristiken.

Reproduktionswahrscheinlichkeit p(R) und Crossoverwahrscheinlichkeit p(C):

---

[38] Vgl. Stender, 1995, S. 41.
[39] Zum Vergleich: die Mutationsrate eines menschlichen Gens liegt bei ca. 0,000007.
[40] Vgl. Schwefel, 1991, S.26f.
[41] Vgl. Bäck, 1996, S.198f.

Die Reproduktionswahrscheinlichkeit liegt in der Literatur zwischen p(R) = 0 und p(R) = 0,5.[42] Für die Crossoverwahrscheinlichkeit wird entsprechend p(M) + p(R) + p(C) ≡ 1 ein Wert zwischen p(C) = 0,99 und p(C) = 0,5 angesetzt. Für beide Wahrscheinlichkeiten sind keine Heuristiken bekannt.

### 2.3.4   Selektion

Die Selektion soll für die Erfüllung des Darwinschen Leitsatzes „survival of the fittest" sorgen. Die schlechteren Individuen sollen seltener zur Zeugung von Kindern herangezogen werden, damit sich die Besseren durchsetzen können. Zuerst muß dazu jedem Individuum eine Güte zugewiesen werden. Problematisch hierbei ist die Einbeziehung von Nebenbedingungen, was in klassischen genetischen Algorithmen nicht vorgesehen ist. Grundsätzlich gibt es drei Möglichkeiten trotzdem Nebenbedingungen zu berücksichtigen:

1. Keine Verwendung von unzulässigen Lösungen.
2. Umwandlung von unzulässigen Lösungen in zulässige Lösungen.
3. Verwendung und gleichzeitige Bestrafung von unzulässigen Lösungen.

Der erste Ansatz setzt voraus, daß alle Individuen in der Anfangslösung zulässig sind und dies ebenfalls für alle durch die genetischen Operatoren erzeugten neuen Individuen gilt. Sollten unzulässige Individuen erzeugt werden, so müssen diese sofort durch zulässige ersetzt werden. Dieser Ansatz ist nur bei einer sehr großen Anzahl von zulässigen Lösungen durchführbar, da ansonsten zuviel Rechenzeit mit unzulässigen Individuen verschwendet wird. Aufgrund der insgesamt eher wenigen zulässigen Lösungen des Nurse Scheduling Problems wird er hier deshalb nicht weiter verfolgt.[43]

Beim zweiten Ansatz wird versucht, unzulässige Individuen so zu modifizieren, daß sie alle Nebenbedingungen erfüllen.[44] Eine mögliche Vorgehensweise ist dabei, unzulässige Kinder so lange an ihre zulässigen Eltern anzugleichen, bis sie ebenfalls wieder zulässig sind.[45] Für das Nurse Scheduling Problem ist dieser Ansatz jedoch aufgrund der komplexen Nebenbedingungen und der insgesamt geringen Zahl an zulässigen Lösungen ebenfalls ungeeignet.

---

[42]  Vgl. Koza, 1993, S. 25, Goldberg, 1989, S. 71 und Michalewicz, 1992, S. 21.
[43]  Für eine ausführliche Beschreibung siehe Chambers, 1995, S. 317-327.
[44]  Vgl. hierzu Eiben, 1994, S. 1-12, Eiben, 1995a, S. 1f und Eiben 1995b, S. 1ff.
[45]  Zu diesem als Backtracking bezeichnetem Verfahren siehe Kennedy, 1993, S.37f.

Der dritte Ansatz, in der Literatur als Strafkostenverfahren bezeichnet, läßt unzulässige Individuen zwar zu, bestraft sie aber für die Nichteinhaltung von Nebenbedingungen.[46] Hierzu werden alle unzulässigen Über- bzw. Unterschreitungen der Nebenbedingungen zum Zielfunktionswert addiert, bei einem Maximierungsproblem entsprechend subtrahiert. Eventuell wird das Verletzen verschiedener Nebenbedingungen unterschiedlich gewichtet.[47] Diese Strafkosten sind wiederum keine tatsächlich auftretenden Kosten, sondern lediglich Lenkparameter, die versuchen, die Lösung in den zulässigen Bereich zu „lenken".

Der Ansatz entspricht einer Mehrfachzielsetzung, wobei die zu gewichtenden Ziele das Orginalziel und die Nebenbedingungen sind. Die gewählte Form der additiven Wertaggregation setzt eine starke Präferenzunabhängigkeit bezüglich der einzelnen Ziele, Vollständigkeit und Transitivität aller Alternativen, sowie eine gegenseitige Substituierbarkeit der einzelnen Attributausprägungen voraus.[48] Ohne dies näher zu untersuchen, soll hier vereinfachend von der Erfüllung dieser Kriterien ausgegangen werden.

Dieser, in der anglo-amerikanischen Literatur als Penalty Function bezeichnete Ansatz, wurde auf viele stark restriktive Probleme mit Erfolg angewendet.[49] Wandelt man das mathematische Modell aus Kapitel 0 entsprechend um und führt die Gewichte $g_1$, $g_2$ und $g_3$ für die drei Nebenbedingungen ein, so erhält man folgende Formulierung:

$$\sum_{i=1}^{n}\left[\sum_{j=1}^{m}\left(p_{ij}x_{ij}\right) + \left(g_1 \left| 1 - \sum_{j=1}^{m} x_{ij} \right|^{+}\right) + \left(g_2 \sum_{j=1}^{m}\sum_{k=1}^{14}\left| a_{jk}x_{ij} - W_i \right|^{+}\right)\right.$$
$$\left. + \left(g_3 \sum_{j=1}^{m}\sum_{s=1}^{p}\sum_{k=1}^{14} \left| \min\left[ q_{is}a_{jk}x_{ij} - R_{ks};0\right] \right|^{+}\right)\right] \rightarrow \min!$$

( 0-4 )[50]

Die eigentliche Selektion erfolgt dann proportional zum Gütemaß.[51] Zur Verdeutlichung dient folgendes Beispiel, bei dem eine Minimierung der Güte der Individuen angestrebt wird:

| Individuum | Kodierung | Güte [Strafkosten] | Selektions-Wahrscheinlichkeit |
|---|---|---|---|

[46] Vgl. Domschke, 1991, S. 180.

[47] Zur Zielgewichtung siehe Schneeweiß, 1991, S.95-99 und Domschke, 1991, S.46f.

[48] Vgl. Schneeweiß, 1991, S.125ff.

[49] Vgl. Michalewicz, 1992, S.97ff und Goldberg, 1989, S.85f.

[50] $\left|...\right|^{+}$ steht für den Betrag der entsprechenden Funktion.

[51] Vgl. Goldberg, 1989, S. 10ff.

| 1 | 001010010 | -48 | -48/-325  = 15 % |
| 2 | 101000100 | -123 | -123/-325 = 38 % |
| 3 | 111010101 | -55 | -55/-325  = 17 % |
| 4 | 111000010 | -99 | -99/-325  = 30 % |
| Summe | | -325 | 100% |

**Tabelle 0-1:** Ein Beispiel zur Erläuterung der Selektion bei Minimierung der Güte.

Obwohl die Auswahl letztlich dem Zufall überlassen bleibt, handelt es sich keinesfalls um eine willkürliche Methode, da die Wahrscheinlichkeit, daß ein Individuum ausgewählt wird, direkt proportional zu dessen Güte ist. Bessere, d.h. „fittere" Individuen werden demnach im Schnitt häufiger zur Erzeugung von Nachkommen herangezogen.

Diese von herkömmlichen genetischen Algorithmen benutzte Art der Selektion proportional zum Gütemaß ist jedoch aus mehreren Gründen problematisch: Einerseits werden wenige gute, in der Startbevölkerung vorhandene Individuen die Entwicklung anfangs stark dominieren und andererseits wird es gegen Ende, wenn alle Individuen in etwa die selbe Güte aufweisen, zu einer rein zufälligen Auswahl der Eltern kommen, was man als mangelnden Selektionsdruck bezeichnet. Außerdem ist eine solche relative Güte als Selektionsmaß nicht verschiebungsinvariant bezüglich der absoluten Güte eines Individuums.

Um diese Probleme zu lösen, wird eine Skalierung der Güte der Individuen vorgeschlagen.[52] Ein häufig verwendetes Verfahren ist die lineare Skalierung mit skalierter Güte = a · Güte + b, wobei a und b so festgelegt werden, daß ein Individuum mit durchschnittlicher Güte auch eine durchschnittliche skalierte Güte besitzt. Eine Verschiebungsinvarianz wird durch eine solche lineare Skalierung jedoch nicht erreicht und die Probleme der Dominanz bzw. des mangelnden Selektionsdrucks werden ebenfalls nicht vollständig gelöst. Je nach Parameterwahl für a und b und der damit verbunden Ent- bzw. Verzerrung kann nur entweder die Anfangsdominanz oder der gegen Ende auftretende mangelnde Selektionsdruck behoben werden.

Besser ist daher eine Auswahl der Individuen nach deren Rang.[53] Dazu wird jedem Individuum ein Rang in der Bevölkerung aufgrund dessen Güte zugewiesen, wobei das schlechteste Individuum Rang 1, das zweit schlechteste Rang 2 etc. bekommt. Der Rang eines Individuums entspricht dann seiner relativen Selektionswahrscheinlichkeit.

---

[52] Vgl. Goldberg, 1989, S.76-79 und Bäck, 1996, S.167-169.
[53] Zu den Rang-Verfahren siehe Bäck, 1996, S. 169-172.

Bei dieser Art der Selektion ist in jeder Phase des Algorithmus sichergestellt, daß die Chance, zum Crossover ausgewählt zu werden, für die besten Individuen deutlich höher ist als für durchschnittliche Individuen. Gleichzeitig wird die anfängliche Dominanz guter Individuen verringert.[54] So erzeugt ein Individuum im Schnitt die folgende Anzahl an Kindern:

$$\text{Gesamtzahl zu erzeugender Kinder} \cdot \frac{\text{Rang des Individuums}}{\text{Rangsumme aller Individuen}} = N \cdot \frac{\text{Rang}}{(N+1)N/2}$$

Das beste Individuum mit Rang = N erzeugt also im Durchschnitt $N \cdot \frac{2N}{N^2 + N} = \frac{2N}{N+1} \approx 2\,\text{Kinder}$.

Ein durchschnittliches Individuum mit Rang = N/2 erzeugt $N/2 \cdot \frac{2N}{N^2 + N} = \frac{N}{N+1} \approx 1\,\text{Kind}$.

### 2.3.5  Crossover

Das Crossover ist der wichtigste Operator der genetischen Algorithmen und wird mit einer Wahrscheinlichkeit von p(C) ausgeführt. Prinzipiell nimmt er zwei oder mehrere Individuen der jetzigen Generation, die sogenannten Eltern, und erzeugt aus ihnen Individuen für die nächste Generation, die sogenannten Kinder. Dies wird durch Aufspalten der Eltern und anschließendes Zusammenfügen zu Kindern erreicht. Je nach Anzahl der Spalt-Punkte spricht man von 1-Punkt, 2-Punkt-, etc., bis n-Punkt-Crossover. Zusätzlich gibt es noch das uniforme Crossover.[55]

Ursprünglich wurde im Bereich der genetischen Algorithmen mit nur zwei Eltern gearbeitet. Neuere Untersuchungen zeigen jedoch, daß in Abhängigkeit vom zu optimierenden Problem, eine größere Anzahl (bis zu 10 Eltern) von Vorteil sein kann.[56] Doch wie bei anderen Parametern lassen sich hierfür ebenfalls keine problemunspezifischen Aussagen treffen. Zur Vereinfachung werden sämtliche Beispiele dieses Kapitel mit zwei Eltern durchgeführt, lassen sich aber ohne weiteres auf beliebig viele Eltern erweitern.

Die einfachste Art des Crossovers ist das 1-Punkt-Crossover. Dazu werden beide Eltern beim selben Element geteilt und die vier Teile über Kreuz zu zwei neuen Kindern wieder zusammengefügt, d.h. Kind 1 bekommt den linken Teil des einen und den rechten Teil des

---

[54] Für weitere Vorteile dieses Verfahrens und einen empirischen Vergleich mit anderen Verfahren siehe Fogarty, S. 83-87.

[55] Zu den Auswirkungen der verschiedenen Crossover Arten auf das Nurse Scheduling Problem siehe Kapitel 0 und 0.

[56] Vgl. Männer, 1994, S. 78ff.

anderen Eltern-Individuums. Bei Kind 2 ist es entsprechend umgekehrt. Der Crossover-Punkt, d.h. der Punkt, an dem die Eltern in zwei Teile geschnitten werden, wird zufällig gewählt. Zur Veranschaulichung dient folgendes Beispiel:

| 0 | 0 | 0 | 0 | 1 | 0 | Eltern-Individuum 1 |
| 1 | 1 | 1 | 0 | 1 | 1 | Eltern-Individuum 2 |
|   | ^ | ^ |   |   |   | Crossover-Punkt |
| 0 | 0 | 1 | 0 | 1 | 1 | Kind 1 |
| 1 | 1 | 0 | 0 | 1 | 0 | Kind 2 |

**Abb. 0-2:** 1-Punkt Crossover.

Entsprechend funktioniert ein 2-Punkt- bis n-Punkt-Crossover. Es werden zufällig 2 bis n Crossover-Punkte festgelegt, die die Eltern somit in sechs bis $(2n + 2)$ Teilstücke teilen. Die so entstandenen Teilstücke werden zu Kindern zusammengesetzt, indem abwechselnd ein Teilstück von jedem Elternteil gewählt wird. Offensichtlich begrenzt die Zahl der Elemente L eines Individuums n auf $n_{max} = L - 1$. Für ein beispielhaftes 4-Punkt-Crossover siehe Abb. 0-3.

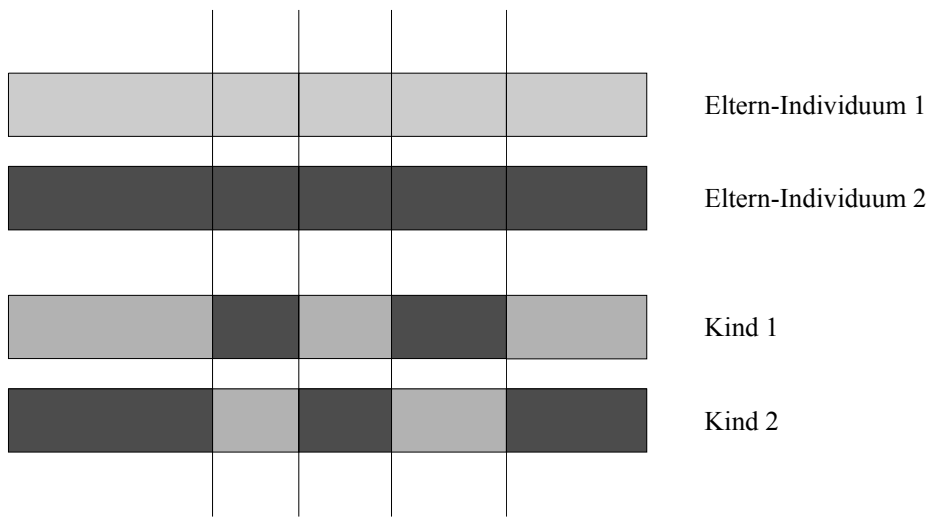

**Abb. 0-3:** Schematische Darstellung des 4-Punkt-Crossovers.

Uniformes Crossover ähnelt $n_{max}$-Punkt-Crossover. Wie bei diesem werden auch hier die Eltern in die maximal mögliche Zahl von Teilen, also in alle Einzelmerkmale aufgespalten. Anstatt nun aber abwechselnd ein Teil von jedem Eltern-Individuum pro Kind zu nehmen, wird jedes Teil zufällig einem Kind zugeordnet. Für ein Beispiel siehe Abb. 0-4. Da beim uniformen Crossover die Freiheitsgrade am höchsten sind, wird es bei genetischen Algorithmen sehr häufig verwendet. Eine allgemeine Aussage, welches Crossover-Verfahren das beste ist, läßt sich jedoch ähnlich wie für die Parameterwerte nur problemspezifisch treffen.[57]

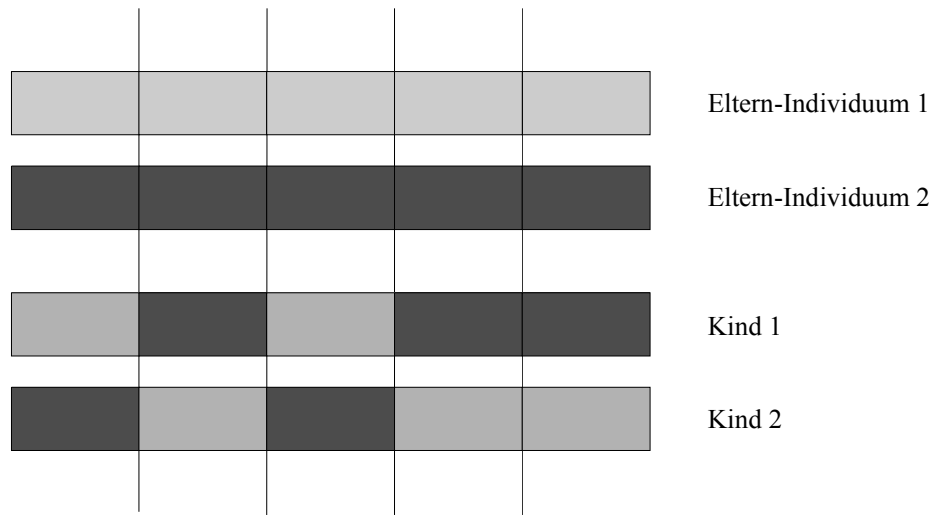

**Abb. 0-4:** Schematisches Darstellung des uniformen Crossovers.

## 2.3.6 Mutation und Reproduktion

Die beiden weiteren Operatoren, mit denen genetische Algorithmen Kinder erzeugen, sind Mutation und Reproduktion. Die Mutation der genetischen Algorithmen ist der Mutation in der Natur nachempfunden. Mit der in der Initialisierungsphase festgelegten Mutations-wahrscheinlichkeit p(M) ändert ein Element eines Individuums zufällig seinen Wert, d.h. bei binärer Kodierung von 0 auf 1 oder umgekehrt.

Im Gegensatz zu den Algorithmen der Evolutions-Strategie-Klasse, die über kein Crossover verfügen und Kinder lediglich über Mutationen erzeugen, ist die primäre Aufgabe der Mutation innerhalb der genetischen Algorithmen nicht die Verbesserung eines Individuums.[58] Vielmehr dient sie zur Wiedereinführung verloren gegangener oder in der Startbevölkerung nicht vorhandener Elemente, die eventuell Teil der optimalen Lösung sein könnten.[59]
Bei der Reproduktion, die mit einer Wahrscheinlichkeit von p(R) ausgeführt wird, handelt es sich um ein identisches Kopieren eines Individuums, das dann unverändert in die neue

---

[57] Vgl. Chambers, 1995, S.113.
[58] Für eine ausführliche Erläuterung dieser Algorithmen und ihrer weiteren Unterschiede zu den genetischen Algorithmen siehe Bäck, S. 66ff.
[59] Vgl. Bäck S. 113ff.

Generation übernommen wird. Dieses Klonen soll das Überleben mancher Eltern simulieren und damit ein zu schnelles Konvergieren des genetischen Algorithmus verhindern.

### 2.3.7   Substitution und Abbruchkriterien

Wurde durch Reproduktion, Mutation und Crossover eine neue Generation geschaffen, so stellt sich die Frage was mit den Individuen der alten Generation geschehen soll. Die übliche Vorgehensweise dabei ist, die alte Generation komplett durch die neue zu ersetzen. Varianten dazu sind z.B. der Zweikampf zwischen einem Eltern-Individuums und dessen eigenem Kind um den Platz in der nächsten Bevölkerung oder das pauschale Ersetzen der X% schlechtesten Eltern durch die entsprechende Anzahl von Kindern.

Bei der Wahl der Substitution muß zwischen einer breiten und einer tiefen Suche abgewägt werden. Eine Substitution, die auch schlechten Individuen eine Chance zum Überleben gibt, kann eine größere Vielfalt von Individuen erzeugen. Diese breit angelegte Suche hat den Nachteil, daß sie relativ lange braucht, um sehr gute Individuen zu erzeugen. Eine Substitution, die dagegen wenige gute Individuen bevorzugt, wird schneller auf eine Lösung konvergieren. Dabei kann sich aber um ein lokales Optimum handeln, da die Suche nicht breit genug gewählt wurde.

Die Anwendung des totalen Ersetzens der Eltern führt zu einer sehr breiten Suche. Problematisch dabei ist, daß die bislang beste gefundene Lösung wieder verloren gehen kann. Behält man die X% besten der Eltern-Generation und wählt X hoch, so bewirkt dies eine sehr intensive Suche nahe dieser Individuen. Dies kann den Nachteil des vorzeitigen Konvergierens auf ein lokales Optima haben. Ist außerdem der Lösungsraum durch viele Nebenbedingungen stark beschränkt, so wird durch eine solche tiefe Suche eventuell keine zulässige Lösung erreicht. X sollte daher nicht allzu hoch gewählt werden.

Ein Mittelweg bildet die Zweikampf-Strategie. Einerseits überleben tendenziell die besseren Lösungen, andererseits haben auch schlechtere Individuen eine gewisse Chance. Trotzdem kann es auch bei Anwendung dieser Strategie zu einer Dominanz weniger kommen, falls die erzeugten Kinder zu schlecht sind. Um auf jeden Fall sicherzustellen, daß die bislang beste Lösung nicht wieder verloren geht, kann eine Elitismus-Strategie angewendet werden. Dazu wird die jeweils beste Lösung einer Generation unverändert in die nächste Generation übernommen.[60]

Typische Abbruchkriterien sind entweder das Überschreiten einer vorgegeben Gesamtanzahl von Generationen bzw. eines Zeitlimits oder das Überschreiten einer Anzahl von Generationen, über die es keine Verbesserung mehr gab. Letzteres setzt allerdings ein Speichern der bislang besten Lösung voraus.

# 2.4    Schemata-Theorem und Building-Block-Hypothese

Um zu verstehen, warum genetische Algorithmen Funktionen tatsächlich optimieren und nicht etwa eine zufällige Suche darstellen, wurden das Schemata-Theorem und die Building-Block-Hypothese entworfen.[61] Sie wurden aus der Beobachtung entwickelt, daß gleiche Bestandteile guter Individuen wichtige Schlüsse auf die notwendige Richtung der Optimierung zulassen.

Um Individuen vergleichbar zu machen wird dazu das Konzept der Schemata herangezogen. Ein Schema ist dabei definiert als ein Individuum, bei dem bestimmte Einzelmerkmale durch ein Ersatzzeichen * ersetzt sind. Ein Ersatzzeichen steht dann für ein beliebig zulässiges Element des gewählten Alphabetes. Hat man z.B. das Schema (10110*) bei einem binären Alphabet, so repräsentiert es zwei Individuen, nämlich (101100) und (101101).

Ein gutes Beispiel für das Denken in Schemata stellt das Spiel „Mastermind" dar, bei dem man einen vierstelligen Zahlencode in möglichst wenigen Zügen erraten muß. Nach jedem Rateversuch nennt der Gegner nur die Anzahl der richtig erratenen Ziffern und die Anzahl der richtig erratenen Ziffern, die zusätzlich an der richtigen Position stehen. Rät man z.B. 4286 und erhält als Antwort, daß exakt eine Ziffer an der richtigen Stelle steht, so bleiben die Schemata (4***), (*2**), (**8*) und (***6) als mögliche Lösungen übrig. Ein * steht hierbei für eine beliebige Zahl aus dem Alphabet von 0-9.

Je nach Anzahl r der Ersatzzeichen in einem Schemata repräsentiert es $(k + 1)^r$ verschiedene Individuen, wobei k die Größe des verwendeten Alphabetes angibt, z.B. gilt für eine

---

[60] Zur allgemeinen Problematik der Substitution beim Nurse Scheduling siehe Kapitel 0, zum Thema Elitismus siehe Kapitel 0. Zum Elitismus vgl. auch Koza, 1993, S.113.

[61] Für eine ausführliche Herleitung des Schemata-Theorems siehe Goldberg, 1989, S. 28ff und Michalewicz, 1992, S.41ff.

Binärkodierung k = 2 und für jedes Element gibt es folglich drei Möglichkeiten: 0, 1 oder *. Ein einzelnes aus L Elementen bestehendes Individuum enthält allerdings nur $k^L$ verschiedene Schemata, da die Ersatzzeichen bei der Verarbeitung durch die genetischen Operatoren durch einen Wert des Alphabetes ersetzt sein müssen.

Zwar kann die Anzahl unterschiedlicher Schemata in einer Bevölkerung der Größe N nicht exakt ermittelt werden, sie muß aber zwischen $k^L$ und $N \cdot k^L$ liegen. Dies zeigt, daß bereits bei kleinen Bevölkerungen eine wahre Informationsfülle in den Individuen steckt. Es läßt sich beweisen, daß bei der Bearbeitung von i Individuen $i^3$ Schemata durch den genetischen Algorithmus verarbeitet werden.[62] Dieser als implizite Parallelität bezeichnete Zusammenhang zeigt die große Rechenkraft genetischer Algorithmen und erklärt, wie die Informationsfülle genutzt werden kann.[63]

Der Einfluß der genetischen Operatoren auf die Schemata ist unterschiedlich: Bei der Selektion werden bekanntlich bessere Individuen, d.h. bessere Schemata vorgezogen. Dies ist eine Entwicklung, die sich durch die daraus erzeugten Kinder exponentiell in zukünftigen Generationen fortsetzt. Kommt es dann zum Crossover, so kann es passieren, daß lange Schemata aufgespalten werden und verloren gehen. Man kann jedoch zeigen, daß im Durchschnitt mindestens soviel gute Schemata neu entstehen, wie durch das Crossover zerstört werden.[64] Je kürzer jedoch die Schemata, desto höher die Chance, daß sie intakt an ein Kind weitergereicht werden und sich folglich noch häufiger vermehren. Während die Reproduktion nur den Status quo erhält, tritt die Mutation zu selten auf, um ins Gewicht zu fallen.

Zusammen liefert dies die Schlußfolgerung, daß kurze und gute Schemata sich über die Generationen hinweg exponentiell vermehren. Diese Art von Schemata werden Building-Blocks genannt. Anschaulich gesprochen versucht der genetische Algorithmus demnach aus kleinen und guten Bausteinen eine optimale Lösung zusammenzusetzen.[65]

---

[62]  Für einen exakten Beweis siehe Goldberg, 1989, S. 20f.
[63]  Vgl. Heistermann, 1994, S.40f zum Begriff der impliziten Parallelität.
[64]  Für einen ausführlichen Beweis siehe Goldberg, 1989, S. 28ff.
[65]  Für eine ausführliche empirische Untersuchung vgl. Goldberg, 1989, S. 41ff und S. 373ff.

# 3    Das Programm

## 3.1    Einführung

Das Programm zum Lösen des Nurse Scheduling Problems wurde in Turbo Pascal 7.0 für DOS geschrieben und auf einem PC mit 100Mhz Pentium Prozessor implementiert. Turbo Pascal wurde aufgrund der Verständlichkeit des Quelltextes und seiner Modularität als Programmiersprache ausgewählt.[66] Dieses Kapitel enthält eine Beschreibung der wichtigsten Bestandteile des Programmes.[67]

Vor der eigentlichen Programmierung muß das Kodierungsproblem für die Individuen gelöst werden. Wie aus Kapitel 0 bekannt, muß die Darstellung einer Lösung bzw. eines Individuums an die Arbeitsweise der genetischen Operatoren angepaßt werden. In Anlehnung an die mathematische Formulierung aus Kapitel 0 soll ein Individuum den kompletten Schichtplan einer Station darstellen. Dies bedeutet, daß ein Individuum so viele Elemente wie die Zahl der Krankenschwestern auf der Station enthalten muß. Jedes Element repräsentiert dann das von einer Krankenschwester gearbeitete Schichtmuster.

Dazu werden zuerst alle theoretisch möglichen Schichtmuster aufgestellt und mit einem Index versehen.[68] Anschließend werden für jede Krankenschwester alle aufgrund ihrer wöchentlichen Arbeitszeit zulässigen Schichtmuster bestimmt. Für eine Vollzeit arbeitende Krankenschwester wären dies z.B. die Tag-Schichtmuster 1 bis 21 und die Nacht-Schichtmuster 22 bis 56. Das Alphabet eines Elements jedes Individuums wird je nach repräsentierter Krankenschwester auf diesen Wertebereich eingeschränkt.

Diese Kodierung hat den Vorteil, daß die erste und die zweite Nebenbedingung des mathematischen Modells aus Kapitel 0, wonach jede Krankenschwester nur ein Schichtmuster arbeiten darf und dieses Schichtmuster ihrer Wochenarbeitszeit entsprechen muß, implizit erfüllt sind. Die durch Vereinfachung von ( 2-4 ) verbleibende Zielfunktion ist daher:

$$\sum_{i=1}^{n}\left[\sum_{j=1}^{m}\left(p_{ij}x_{ij}\right)+\left(g_{Bedarf}\sum_{j=1}^{m}\sum_{s=1}^{p}\sum_{k=1}^{14}\left|\min\left[q_{is}a_{jk}x_{ij}-R_{ks};0\right]\right|^{+}\right)\right]\rightarrow\min! \qquad (\,0\text{-}1\,)$$

---

[66]  Zu Turbo Pascal siehe Hennefeld, 1995.
[67]  Für ein Listing der Hauptroutine siehe Anhang B.
[68]  Siehe Anhang A4.

Der Wert der dritten Nebenbedingung, also der Bedarfsdeckung, sowie der Zielfunktionswert kann bei der gewählten Kodierung durch einfaches Summieren ermittelt werden. Die geforderte Eindeutigkeit und Einfachheit der Kodierung sind somit gegeben. Zur Erläuterung soll folgender Beispielschichtplan für eine Station mit zehn Krankenschwestern dienen:

$$(2; 32; 153; 213; 198; 94; 74; 3; 45; 142).$$

Da das Schichtmuster mit dem Index 2 laut der Liste aller Schichtmuster (11110100000000) entspricht, bedeutet dies für die erste Krankenschwester, daß sie in dieser Woche Vollzeit Tag-Schichten arbeitet.[69] Sie wird am Sonntag, Montag, Dienstag, Mittwoch und Freitag zum Einsatz kommen.

Die vom Programm benötigen Eingabedaten sind:

- Die Nachfrage nach Krankenschwestern pro Qualifikationsstufe und Tag bzw. Nacht (siehe Anhang A1).
- Die möglichen Arbeitszeiten und Qualifikationsstufen der einzelnen Krankenschwestern (siehe Anhang A2).
- Die Wünsche der Krankenschwestern bezüglich Tag- bzw. Nachtarbeit und den einzelnen Wochentagen (siehe Anhang A3).
- Eine Auflistung aller theoretisch möglichen Schichtmuster (siehe Anhang A4).
- Eine Zusammenstellung aller möglichen Krankenschwester-Schichtmuster-Paare und deren Strafkosten (siehe Anhang A5).

## 3.2    Beschreibung der Programmbestandteile

### 3.2.1  Initialisierung

---

[69] Für eine Liste aller möglichen Schichtmuster siehe Anhang A4.

Bevor die eigentliche Initialisierung erfolgen kann, müssen für alle Parameter Werte festgelegt werden:[70]

- Größe der Bevölkerung N, z.B. N = 1000.
- Mutationswahrscheinlichkeit p(M), z.B. p(M) = 1%.
- Reproduktionswahrscheinlichkeit p(R), z.B. p(R) = 5%.
- Art des Crossovers, z.B. uniform.
- Anzahl der Eltern, z.B. zwei.
- Art der Substitution der Eltern, z.B. totale Substitution.
- Art des Abbruchkriteriums, z.B. nach zwei Minuten Laufzeit.
- Gewicht für den zur Einbindung der Nebenbedingung der Bedarfsdeckung nötigen Strafkostenansatz, z.B. Gewicht für die Bedarfsdeckung $g_{Bedarf}$ = 10.

Sind die Parameter festgelegt und die Eingabedaten vom Programm eingelesen wird eine zufällige Anfangsbevölkerung der Größe N erzeugt:

Pseudocode der Erzeugung der Startbevölkerung:

$\forall$ *Individuen i = 1,..., Größe der Bevölkerung tue*
    $\forall$ *Krankenschwestern k= 1,..., Anzahl_Krankenschwestern tue*
        *Individuum i [Schichtmuster für k] = ein zufälliges für k zulässiges Schichtmuster.*

## 3.2.2  Selektion

Die Güte eines Individuums wird nach Gleichung (3-1) wie folgt berechnet: Die Summe aus den Krankenschwester-Schichtmuster-Strafkosten und den gewichteten Bedarfsunterdeckungen für alle Tage bzw. Nächte und auf allen Qualifikationsstufen. Es muß dabei darauf geachtet werden, daß höher qualifizierte Krankenschwestern solche mit niedriger Qualifikation ersetzten können, d.h. bei einer Berechnung des Tages-Angebotes an Krankenschwestern einer Qualifikationsstufe zählen alle Tages-Angebote an Krankenschwestern einer höheren Qualifikationsstufe mit.

Da es in der endgültigen Lösung auf keinen Fall zu einer Bedarfsunterdeckung kommen darf, muß das Gewicht der Bedarfsdeckung entsprechend hoch gewählt werden. Die exakte Höhe des

---

[70]  Diese Werte sind zunächst rein intuitiv gewählt.

Gewichtes hängt von der Datenkonstellation einer Woche ab und liegt etwa zwischen 1 und 100. Je geringer der Überschuß an Krankenschwestern über den minimalen Bedarf hinaus, desto höher ist tendenziell das Gewicht. Eine genaue Analyse der Zusammenhänge zwischen dem Gewicht der Bedarfsdeckung und der Zulässigkeit einer Lösung erfolgt in den Kapiteln 0 und 0.

Pseudocode der Berechnung der Güte:

*∀ Individuen i = 1,..., Größe der Bevölkerung tue*
    *Begin*
        *∀ Qualifikationsstufen q = 1,..., 3 tue*
            *Begin*
                *∀ Tage bzw. Nächte t = 1,..., 14 tue*
                    *Begin*
                    *Berechne das Angebot(k) an Krankenschwestern der Stufe q.*
                    *Wenn Angebot(k,q) größer als Nachfrage(k,q)*
                    *Dann Angebot(k,q) = Nachfrage(k,q).*
                    *Ende.*
                *Wochenangebot(q) = Σ der Tages-Angebote(q) über k.*
                *Nachfrageüberhang(q) = Wochennachfrage(q) - Wochenangebot(q).*
                *Ende.*
        *N(i) = Σ Nachfrageüberhang N(q) über q.*
        *P(i) = Σ Krankenschwester-Schichtmuster-Strafkosten.*
        *Güte von Individuum i G(i) = P(i) + g_{Bedarf} · N(i).*
        *Ende.*

Die eigentliche Selektion erfolgt, wie in Kapitel 0 ausführlich beschrieben, nicht proportional zur Güte, sondern zum Rang der Individuen innerhalb der Bevölkerung. Dies hat den Nachteil, daß eine rechenzeitintensive Sortierung aller Individuen nach deren Güte notwendig ist.

Pseudocode der Selektion:

*Sortiere alle Individuen anhand ihrer Güte und weise entsprechenden Rang zu.*
*∀ Kinder k = 1,..., Anzahl zu erzeugender Kinder tue*
    *Begin*
        *∀ j = 1,..., Anzahl der Eltern pro Kind tue*
            *Begin*
            *i := 0. Summe := 0.*
            *Zahl := Zufällig aus der Summe der Ränge aller Individuen.*
            *Wiederhole*
                *i := i + 1.*
                *Summe := Summe + Rang von Individuum[i].*
            *Bis (Summe >= Zahl).*
            *Eltern-Individuum[j] := Individuum [i].*
            *Ende.*

*Wähle Zufallszahl zwischen [0;1] und vergleiche mit p(C), p(M) und p(R).*
*Führe Crossover, Mutation oder Reproduktion durch und erzeuge Kind k.*
*Ende.*

### 3.2.3 Crossover

Wurde zur Erzeugung eines Kindes der Crossover-Operator gewählt, so besteht die Aufgabe darin, die Eltern in die für das gewählte Crossover-Verfahren nötige Anzahl von Teilstücken zu zerschneiden. Dazu müssen zuerst die Crossover-Punkte bestimmt werden. Anschließend werden die Teilstücke umgruppiert und wieder zu Kindern zusammengesetzt. Zur Vereinfachung der Darstellung wird lediglich ein Pseudocode für die Erzeugung eines Kindes mit zwei Eltern angeführt.

Pseudocode für 1-Punkt-Crossover mit zwei Eltern:

*Crossover-Punkt := Wähle zufällig eine Krankenschwester.*
*∀ Krankenschwestern k = 1,..., Crossover-Punkt tue*
    *Kind-Schichtmuster[k] := Elternteil_1-Schichtmuster[k].*
*∀ k = Crossover-Punkt + 1,..., Anzahl der Krankenschwestern tue*
    *Kind-Schichtmuster[k] := Elternteil_2-Schichtmuster[k].*

Pseudocode für uniformes Crossover mit zwei Eltern:

*∀ Krankenschwestern k = 1,..., Anzahl der Krankenschwestern tue*
    *Werfe eine faire Münze.*
    *Wenn Münze = Kopf*
    *Dann Kind-Schichtmuster[k] :=Elternteil_1-Schichtmuster[k].*
    *Sonst Kind-Schichtmuster[k] :=Elternteil_2-Schichtmuster[k].*

### 3.2.4 Mutation und Reproduktion

Die Mutation soll für die Wiedereinführung verlorengegangener bzw. in der Startbevölkerung nicht vorhandener Schichtmuster sorgen. Wurde die Mutation als genetischer Operator ausgewählt, so wird das Kind zuerst gleich dem Eltern-Individuum gesetzt. Anschließend werden ein oder mehrere Schichtmuster des Kindes zufällig verändert. Durch die gewählte Implementierung ist es theoretisch auch möglich, daß kein Schichtmuster des Kindes verändert wird. Dies würde dann ausnahmsweise nicht zu einer Mutation sondern zu einer Reproduktion führen.

Damit die durch die spezielle Kodierung erzielte implizite Erfüllung der Nebenbedingung aus Kapitel 0 erhalten bleibt, kann ein Schichtmuster nur im Rahmen des für das zu verändernde Element zulässigen Alphabetes, d.h. im Rahmen der für die entsprechende Krankenschwester zulässigen Wochenarbeitszeit, verändert werden. Andernfalls wäre es möglich, daß eine Krankenschwester ein für sie nicht zulässiges Schichtmuster arbeiten müßte.

Pseudocode der Mutation eines Individuums:

*Mutations-Chance := 1 / Anzahl der Krankenschwestern.*
*∀ Krankenschwestern k = 1,..., Anzahl der Krankenschwestern tue*
    *Begin*
    *Kind-Schichtmuster[k] := Eltern-Schichtmuster[k].*
    *Wenn Zufallszahl zwischen [0;1] < Mutations-Chance*
    *Dann Kind-Schichtmuster[k] =    ein zufällig ausgewähltes und für diese*
                                   *Krankenschwester zulässiges Schichtmuster.*
    *Ende.*

Die Reproduktion entspricht dem Klonen eines Individuums und simuliert das intakte Überleben einiger Eltern-Individuen.

Pseudocode der Reproduktion eines Individuums:

*∀ Krankenschwestern k = 1,..., Anzahl der Krankenschwestern tue*
    *Kind-Schichtmuster[k] := Eltern-Schichtmuster[k].*

## 3.2.5  Substitution und Abbruchkriterien

Bei der Substitution wird darüber entschieden, wer die Eltern-Individuen der nächsten Generation sein sollen. Sind dies nur die Kinder dieser Generation, so spricht man von einer totalen Substitution. Tritt ein Kind-Individuum um den Platz in der nächsten Generation gegen sein eigenes Eltern-Individuum zum Zweikampf an und es kommt nur der Bessere weiter, so spricht man von einer Zweikampf-Substitution. Eine weitere Variante ist das Weiterkommen der X% besten Eltern und einer entsprechenden Anzahl von Kindern, um die neue Generation
aufzufüllen. Dieses Verfahren soll X%-Substitution genannt werden.

Pseudocode der totalen Substitution:

$\forall$ *Individuen i = 1,..., Größe der Bevölkerung tue*
    *Eltern-Individuum i der nächsten Generation := Kind-Individuum i.*

Pseudocode der Zweikampf-Substitution:

$\forall$ *Individuen i = 1,..., Größe der Bevölkerung tue*
    *Wenn Güte von Kind-Individuum[i] < Güte von Eltern-Individuum[i]*
    *Dann Eltern-Individuum[i] der nächsten Generation = Kind-Individuum[i].*
    *Sonst Eltern-Individuum[i] der nächsten Generation = Eltern-Individuum[i].*

Pseudocode der X%-Substitution:

$\forall$ *Individuen i := 1,..., Größe der Bevölkerung tue*
    *Wenn der Rang von Eltern-Individuum[i] <= Größe der Bevölkerung $\cdot$ X%*
    *Dann Eltern-Individuum[i] der nächsten Generation = Kind-Individuum[i].*
    *Sonst Eltern-Individuum[i] der nächsten Generation = Eltern-Individuum[i].*

Als Abbruchkriterium für den genetischen Algorithmus ist eine Obergrenze für die Zahl der
Generationen festgelegt. Alternativ dazu ist der Abbruch nach einer gewissen Laufzeit in
Sekunden möglich.

Pseudocode der Abbruchbedingungen:

*Generationen := 0.*
*Laufzeit := 0.*
*Wiederhole*
    *kompletter genetischer Algorithmus.*
    *Generation := Generation + 1.*
    *Laufzeit := Laufzeit + Rechenzeit für die letzte Generation.*
*Bis (Generation > Maximale Generationenanzahl) oder (Laufzeit > Maximale Laufzeit).*

# 4    Erste Analysen

## 4.1    Ergebnisse von CPLEX

Formuliert man das Problem analog zur mathematischen Formulierung aus Kapitel 0 als ein ganzzahliges lineares Programm,[71] so ist CPLEX für einige Datenreihen in der Lage, die optimale Lösung innerhalb weniger Sekunden zu liefern. Die CPLEX dabei zur Verfügung stehende Hardware entspricht ebenfalls etwa einem PC mit 100 MHz Pentium Prozessor. Jedoch ist CPLEX nicht für alle Datenreihen in der Lage, in akzeptabler Zeit optimale Lösungen zu liefern.

Die Gründe hierfür liegen in der unterschiedlichen Größe des Lösungsraumes der Datenreihen. Bei denjenigen Datenreihen, die nur wenige zulässige Lösungen aufwiesen, war CPLEX durch die angewandte Cutting-Plane-Technik mit anschließendem Branch-and-Bound sehr schnell. Bei Datenreihen mit vielen zulässigen Lösungen und bei Datenreihen, die sehr unterschiedliche Nebenbedingungen aufwiesen, zeigte sich der genetische Algorithmus in der Regel schneller.

Besonders bei ungleichförmiger Nachfrage nach Krankenschwestern konnte der genetische Algorithmus die Probleme besser als CPLEX lösen. Soweit CPLEX in der Lage war eine Datenreihe zu lösen ist dies im folgenden ausdrücklich erwähnt, um die Leistung des genetischen Algorithmus damit zu vergleichen. Die für Kapitel 0 bis 0 gewählte Datenreihe wurde von CPLEX innerhalb einer Minute optimal gelöst und ist daher für Vergleichszwecke geeignet.

## 4.2    Grundsätzliche Vorgehensweise der Analysen

Für alle Testläufe der Kapitel 0 und 0 werden, außer bei deren Variation, folgende intuitiv gewählte Parameter-Werte benutzt: Mutationsrate $p(M) = 1$ %, Reproduktionsrate $p(R) = 5$ %, uniformes Crossover mit zwei Eltern, totale Substitution der alten Generation und als Gewicht für die Bedarfsdeckung $g_{Bedarf} = 20$. Alle Ergebnisse wurden auf einem PC mit 100Mhz Pentium Prozessor unter DOS erzielt.

---

[71]  Für ein verkürztes CPLEX-Eingabefile siehe Anhang C.

Um zu statistisch guten Werten zu gelangen, werden alle Testreihen für jede Bevölkerungsgröße bzw. für jeden einzelnen Parameter-Wert zwanzigfach wiederholt. Für jede Bevölkerungsgröße bzw. für jeden Parameter-Wert werden dabei aus Vergleichsgründen stets die selben zwanzig Ausgangslösungen gewählt.

Die in diesem Kapitel gewählte Datenreihe entspricht der in Anhang A aufgeführten. Es handelt sich dabei um 21 Krankenschwestern aus drei Qualifikations- und sieben Wochenarbeitszeitstufen. Es mußte täglich ein Tages- bzw. Nachtbedarf von zwei bzw. einer Krankenschwester auf der höchsten, zwei bzw. einer Krankenschwester auf der mittleren und fünf bzw. einer Krankenschwestern auf der niedrigsten Qualifikationsstufe erfüllt werden.

# 4.3   Variation der Parameter

## 4.3.1   Bevölkerungsgröße

Der erste Testlauf des Programmes soll Aufschluß über dessen Geschwindigkeit geben. Dazu wurde bei verschieden hoher Bevölkerung die Zeit gemessen, die der Algorithmus für die Berechnung einer komplett neuen Generation, d.h. Selektion, Mutation, Reproduktion, Crossover und Substitution, benötigt. Die Obergrenze von 4000 für die Bevölkerung ergab sich aus dem zur Verfügung stehenden Speicher des verwendeten PCs.

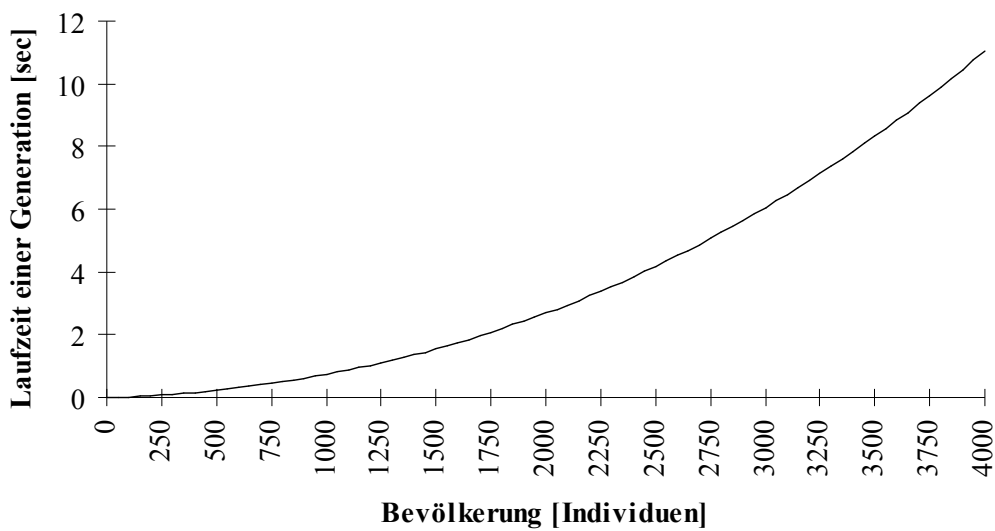

**Abb. 0-1:**   Laufzeit des Algorithmus für eine Generation bei Variation der Größe der Bevölkerung.

Der grundsätzliche Anstieg der Laufzeit mit wachsender Bevölkerung in Abb. 0-1 läßt sich mit der steigenden Anzahl der zu erzeugenden Kinder und der dazu nötigen proportionalen Anzahl an genetischen Operationen erklären. Die quadratische Komponente des Anstiegs hängt mit dem für die Selektion nötigen Sortieralgorithmus zusammen. Aufgrund der hohen Laufzeit für Bevölkerungen über 2000 werden diese im folgenden nicht weiter betrachtet.

### 4.3.2 Laufzeit des Algorithmus

Die nächste Versuchsreihe soll den Zusammenhang zwischen der Lösungsqualität und der dem Algorithmus zur Verfügung stehenden Laufzeit aufzeigen. Außerdem wird untersucht, ob eine größere Bevölkerung stets zu einer besseren Lösungsqualität führt. Dazu wurde das Programm jeweils ein, zwei und drei Minuten lang bei verschieden großen Bevölkerungen ausgeführt.

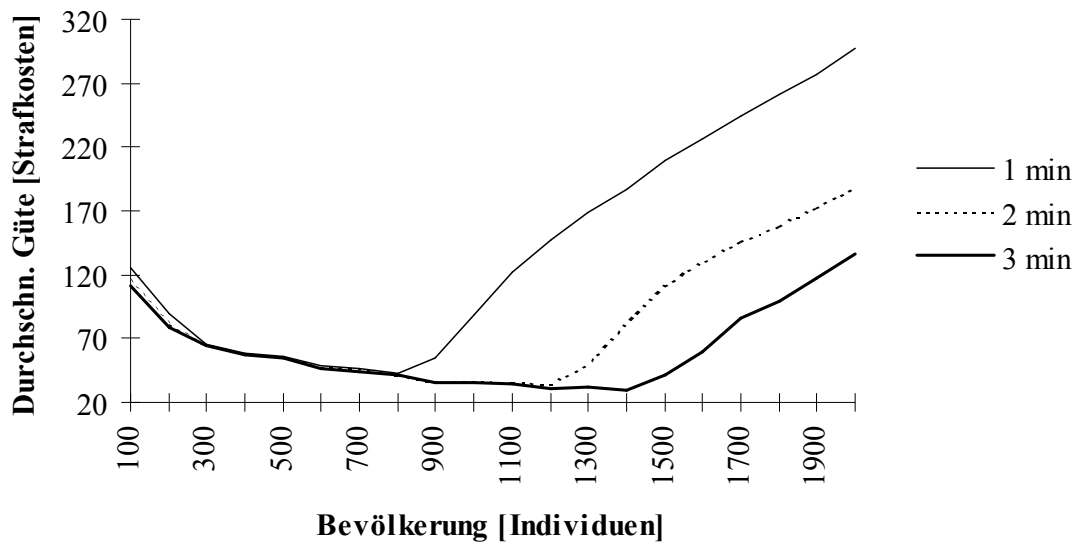

**Abb. 0-2:**  Durchschnittliche Güte aller Individuen der letzten Generation bei Variation der Gesamtlaufzeit des genetischen Algorithmus und der Größe der Bevölkerung.

Zuerst fällt auf, daß die Güte der Individuen mit steigender Bevölkerung sowohl in Abb. 0-2 wie auch in Abb. 0-3 besser wird. Da jedoch der Algorithmus bei steigender Bevölkerung mehr Zeit für eine Generation braucht und die Laufzeit begrenzt ist, bricht diese Entwicklung bei einer gewissen Bevölkerungshöhe ab. Der Algorithmus hatte dann nicht mehr genug Rechenzeit zur Verfügung, um seinen optimalen Wert zu erreichen.

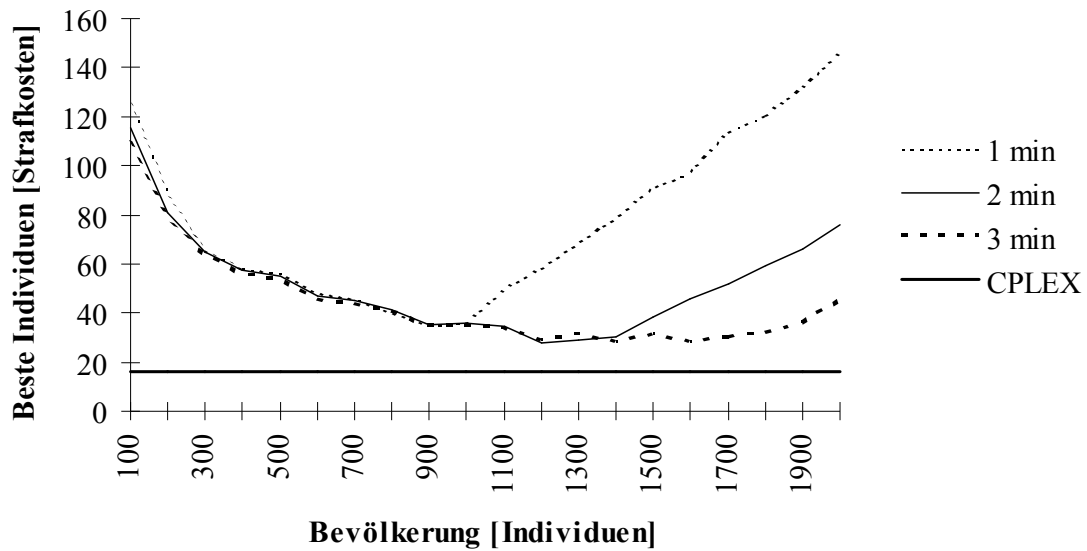

**Abb. 0-3:** Vergleich der Ergebnisse von CPLEX mit der durchschnittlichen Güte des besten Individuums der letzten Generation bei Variation der Gesamtlaufzeit des genetischen Algorithmus und der Größe der Bevölkerung.

Betrachtet man nur den Durchschnitt der besten Individuen der letzten Generation, so ist für eine Laufzeit von einer Minute eine Bevölkerung von 900, für zwei Minuten eine Bevölkerung von 1200 und für drei Minuten eine Bevölkerung von 1600 optimal. Längere Laufzeiten bringen bei jeweils kleineren Bevölkerungen keine Verbesserungen.[72] Insgesamt fällt weiterhin auf, daß der insgesamt beste Wert bei zwei Minuten Laufzeit erreicht wird. Eine Verlängerung der Laufzeit bei gleichzeitiger Vergrößerung der Bevölkerung ist nicht vorteilhaft. Ein Grund hierfür ist, daß die jeweils beste Lösung einer Generation nicht gespeichert wird und dadurch bei längerer Laufzeit wieder verloren gehen kann.

Vergleicht man die durchschnittliche Güte aller Individuen der letzten Generation in Abb. 0-2 mit der durchschnittlichen Güte des besten Individuums der letzten Generation in Abb. 0-3, so fällt auf, daß falls der Algorithmus genügend Zeit hat, bzw. die Bevölkerung genügend klein ist, beide Werte annähernd gleich sind. Dies läßt sich mit der Dominanz der besten Schemata erklären: Durch deren überproportionale Ausbreitung besteht die gesamte Bevölkerung nach ausreichender Zeit nur noch aus gleichen und sehr ähnlichen Individuen. Je kleiner die Bevölkerung desto schneller ist dieser Prozeß abgeschlossen.

---

[72] Beide Beobachtungen decken sich interessanterweise mit Ergebnissen von R. Nakano in Männer, 1994, S.136ff, der ebenfalls eine optimale Bevölkerung von knapp über 1000 für eine ähnliche Laufzeitbegrenzung und ein entfernt ähnliches Problem erhielt.

Vergleicht man das optimale Ergebnis von CPLEX mit den Ergebnissen des genetischen Algorithmus in Abb. 0-3, so schneidet letzterer trotz intuitiv gewählter Parameter-Werte relativ gut ab. Jedoch sind durch die Einbindung der Nebenbedingungen mittels Gewichten in die Zielfunktion nicht alle ermittelten Lösungen zulässig. Wie man aus Abb. 0-4 ersehen kann, sind dies selbst im besten Fall nur weniger als die Hälfte aller Lösungen.

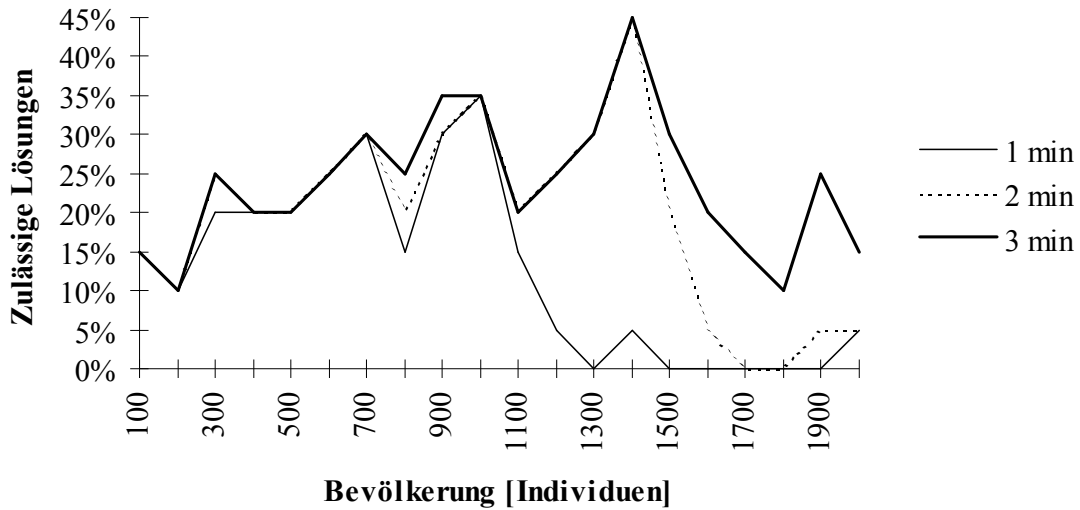

**Abb. 0-4:**   Prozentualer Anteil zulässiger Lösungen der besten Individuen der letzten Generation bei Variation der Gesamtlaufzeit des genetischen Algorithmus und der Größe der Bevölkerung.

In den folgenden Abschnitten wird versucht, durch eine Optimierung der einzelnen Parameterwerte die Qualität der Lösungen zu verbessern. Durch obigen Zusammenhang zwischen den Durchschnitten der Bevölkerung und dem Durchschnitt der besten Individuen der letzten Generation werden sich die Untersuchungen meist auf letzteren beschränken. Weiterhin wird im folgenden aufgrund der insgesamt beschränkten Bearbeitungszeit der Diplomarbeit mit einer Laufzeit von 2 Minuten und einer Referenzbevölkerung von 1200 gearbeitet, welche für diese Laufzeit die besten Werte erzielte.

### 4.3.3   Mutationsrate

Der erste zu optimierende Parameter ist die Mutationsrate. Bei einer groben Variation zeigt sich in Abb. 0-5, daß kleine Werte sowohl für den Durchschnitt der Güte aller Individuen, wie auch der besten Individuen der letzten Generation von Vorteil sind.

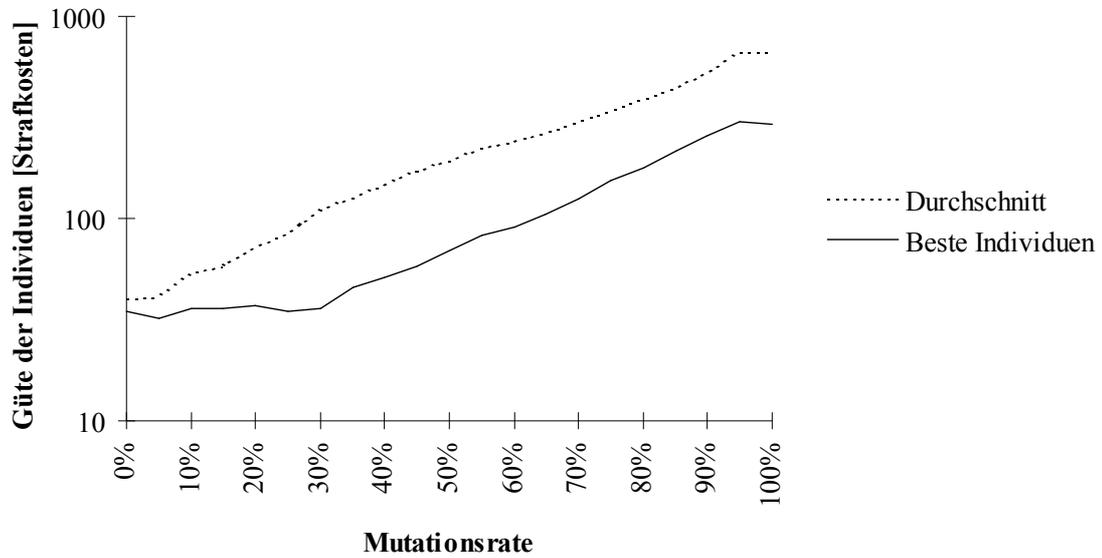

**Abb. 0-5:**  Durchschnittliche Güte eines Individuums der letzten Generation bei grober Variation der Mutationsrate.

Variiert man die Mutationsrate feiner innerhalb des interessanten Bereiches zwischen 0% und 5%, so zeigt sich in Abb. 0-6, daß die Unterschiede nur geringfügig. Die vermutete Mutationsrate von 1% ist ein guter Wert.

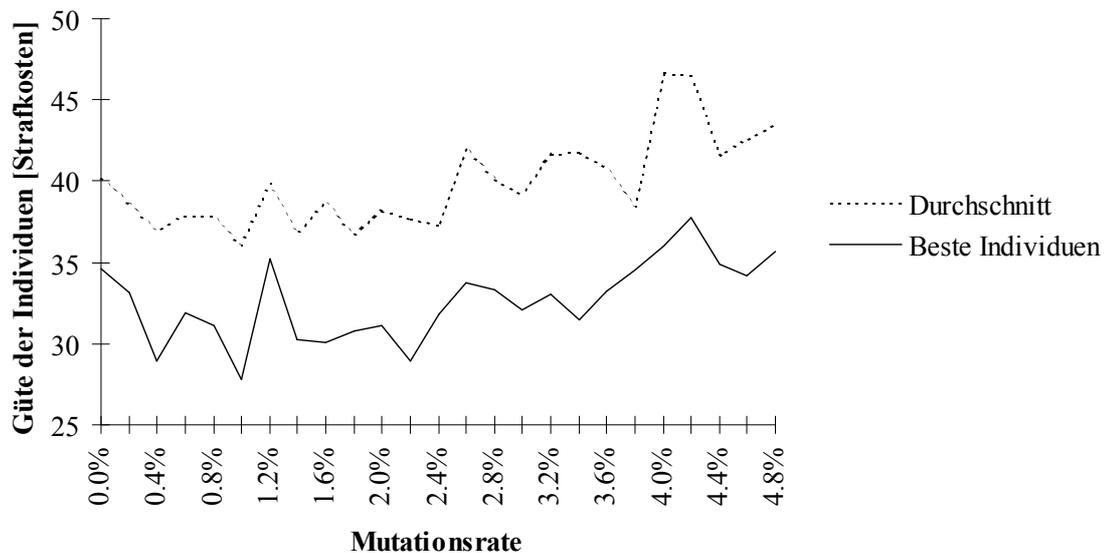

**Abb. 0-6:**  Durchschnittliche Güte eines Individuums der letzten Generation bei feiner Variation der Mutationsrate.

### 4.3.4   Reproduktionsrate

Eine grobe Variation der Reproduktionsrate in Abb. 0-7 zeigt einen ähnlichen Verlauf wie die der Mutationsrate in Abb. 0-5.

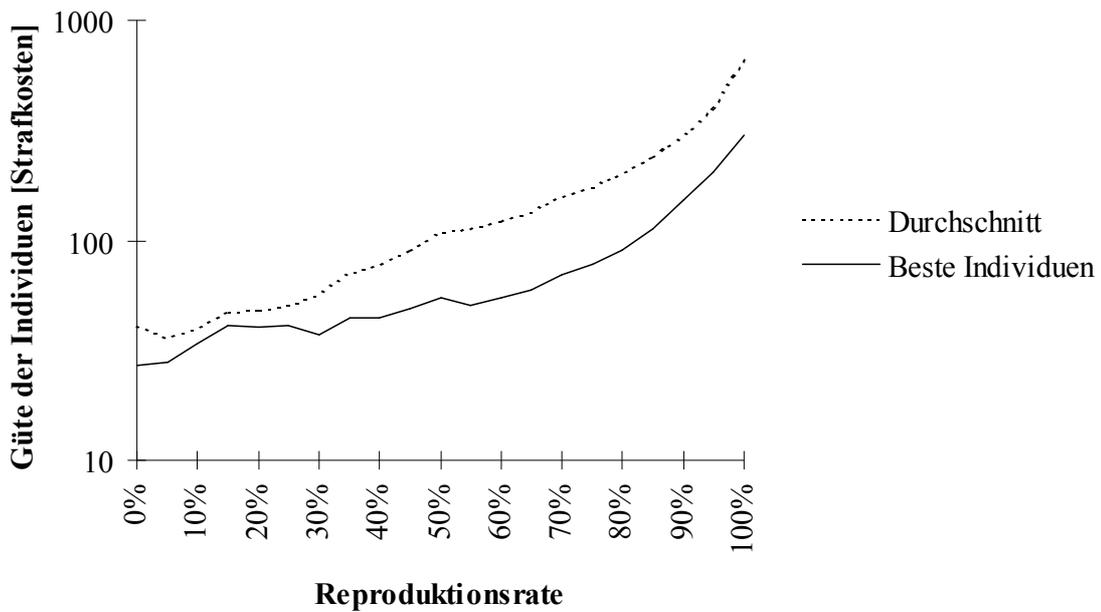

**Abb. 0-7:**   Durchschnittliche Güte eines Individuums der letzten Generation bei grober Variation der Reproduktionsrate.

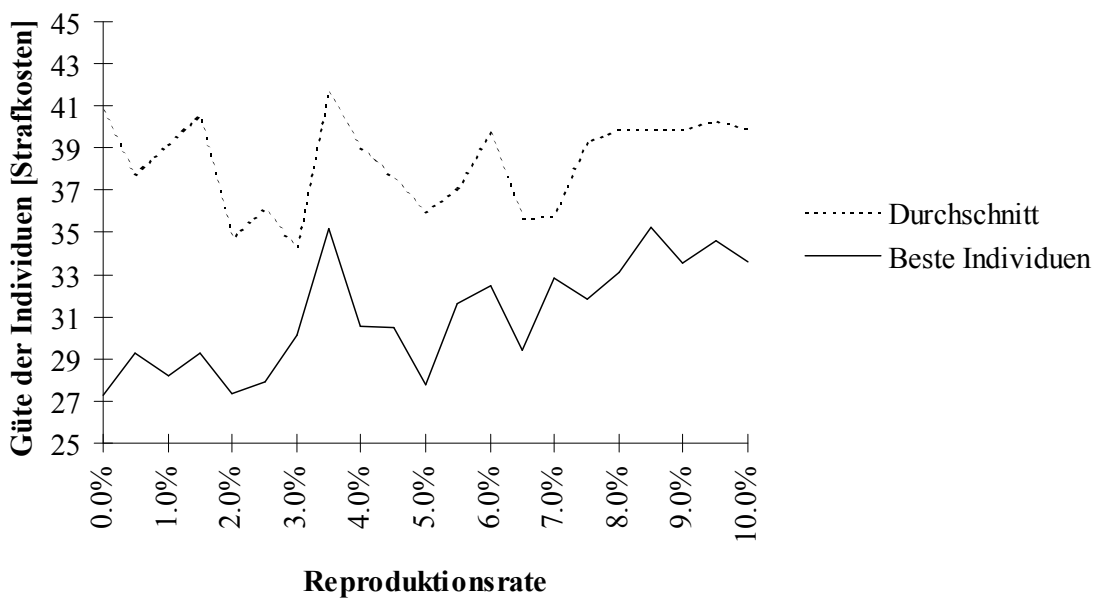

**Abb. 0-8:**   Durchschnittliche Güte eines Individuums der letzten Generation bei feiner Variation der Reproduktionsrate.

Variiert man die Reproduktionsrate innerhalb des interessanten Bereiches zwischen 0% und 10%, so sieht man in Abb. 0-8, daß der beste Wert etwas unterhalb der geschätzten 5% bei ca. 2% auftritt. Wie schon bei der Mutationsrate sind auch hier die Unterschiede gering.

### 4.3.5   Gewicht der Bedarfsdeckung

Da sich die Crossover-Wahrscheinlichkeit aus der Wahrscheinlichkeit von Mutation und Reproduktion ergibt, bleibt als letzter zu variierender Parameter das Gewicht der Bedarfsdeckung. Es soll im folgenden untersucht werden, wie sich eine Veränderung des Gewichtes auf die Lösungsqualität auswirkt und ob ein größeres Gewicht der Bedarfsdeckung, d.h. eine stärkere Berücksichtigung der Nebenbedingungen innerhalb der Zielfunktion, für mehr zulässige Lösungen sorgt.

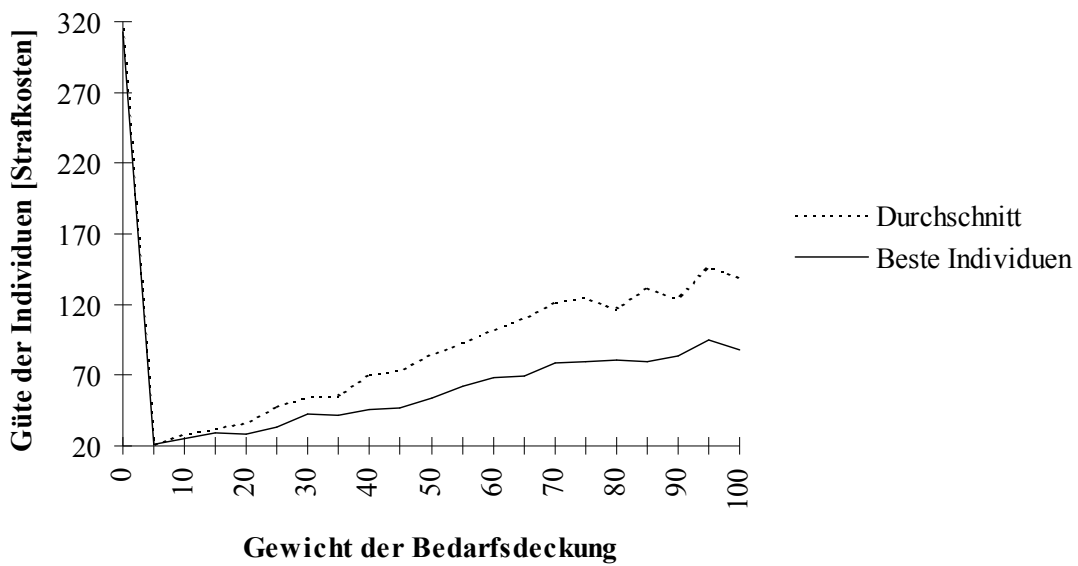

**Abb. 0-9:**   Durchschnittliche Güte eines Individuums der letzten Generation bei grober Variation des Gewichtes der Bedarfsdeckung. Zur Erstellung des Schaubilds erfolgte eine Umrechnung mit g = 20.

Eine grobe Variation des Gewichtes in Abb. 0-9 zeigt, daß gute Gewichte relativ klein sein müssen. Eine weitere, feinere Variation in Abb. 0-10 ergibt, daß ein gutes Gewicht bei ca. 5 und damit wesentlich unterhalb des geschätzten Wertes von 20 liegt.

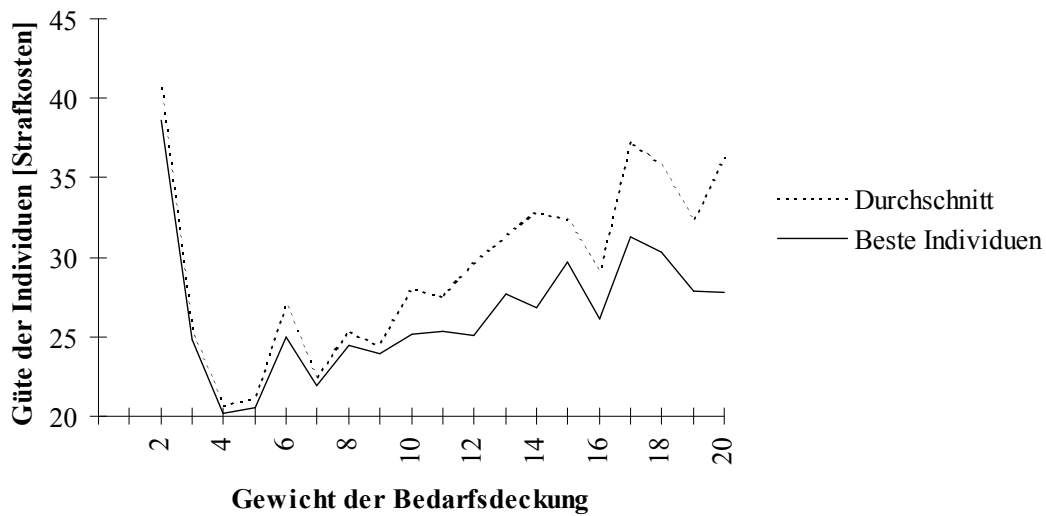

**Abb. 0-10:** Durchschnittliche Güte eines Individuums der letzten Generation bei feiner Variation des Gewichtes der Bedarfsdeckung. Für das Schaubild erfolgte eine Umrechnung mit g = 20.

Wie aus Abb. 0-11 ersichtlich, ist der prozentuale Anteil an zulässigen Lösungen stark vom gewählten Gewicht für die Bedarfsdeckung abhängig. Überraschenderweise steigt jedoch der prozentuale Anteil an zulässigen Lösungen nicht monoton mit einem steigenden Gewicht an, sondern hat bei einem Wert von etwa fünf sein Maximum.

Dies liegt daran, daß ein zu hohes Gewicht dazu führt, daß geringfügig unzulässige Lösungen gegenüber anderen unzulässigen Lösungen zum Erzeugen der Kinder stark bevorzugt werden. Dies schränkt jedoch die Vielfalt der Schemata ein und hat zum Ergebnis, daß zulässige Lösungen oft überhaupt nicht erzeugt werden können, sondern der Algorithmus auf einer geringfügig unzulässigen Lösung vorzeitig konvergiert.

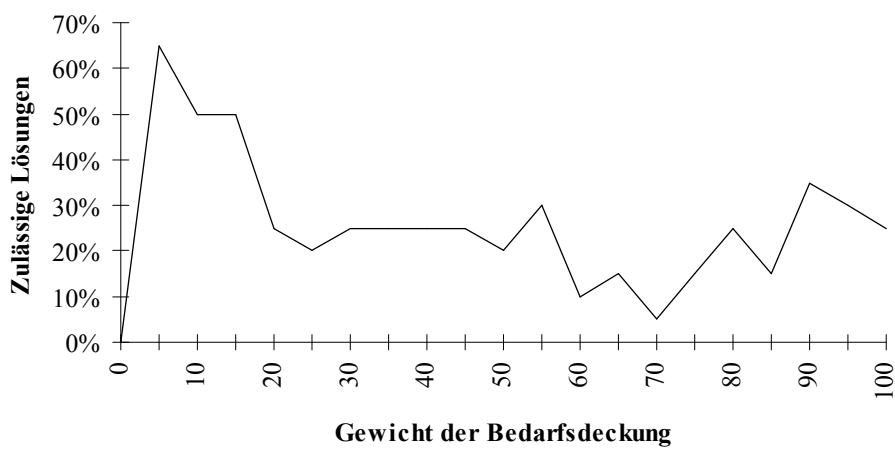

**Abb. 0-11:** Prozentualer Anteil zulässiger Lösungen des besten Individuums der letzten Generation bei Variation des Gewichtes der Bedarfsdeckung.

## 4.4   Variation der genetischen Operatoren

### 4.4.1   Art der Substitution der alten Generation

Durch die Art der Substitution wird festgelegt, wie die Eltern der alten Generation ersetzt werden: komplett durch die Kinder (total), alle bis auf die 10% Besten durch die Kinder (Besten 10%) oder durch die jeweils besseren aus einem Elternteil und dessen eigenem Kind (Zweikampf).

Wie man aus Abb. 0-12 sehen kann, schneidet das Verfahren, bei dem alle bis auf die 10% Besten der Eltern ersetzt werden, am besten ab. Diese Verfahren weist zwei Vorteile auf: Einerseits behält es die jeweils beste Lösung bei, die beim totalen Ersetzen verlorengehen kann und andererseits gibt es in jeder Generation 90% neue Individuen, was für Vielfalt sorgt. Letzteres ist bei der Zweikampf-Strategie nicht gegeben, hier dominieren sehr schnell wenige Individuen bzw. deren Schemata.

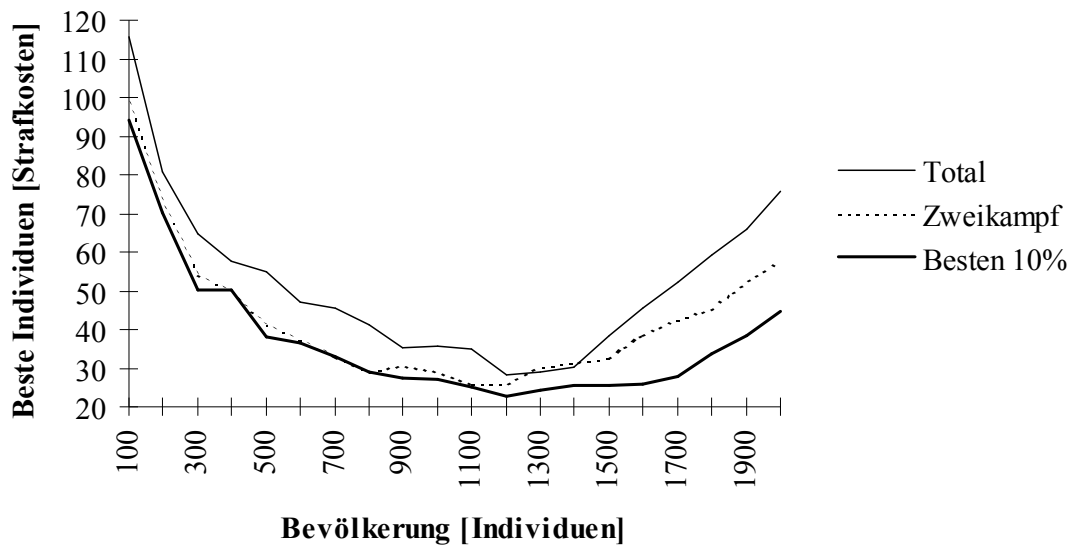

**Abb. 0-12:**   Durchschnittliche Güte des besten Individuums der letzten Generation bei verschiedenen Substitutions-Strategien.

Vergleicht man den Anteil an zulässigen Lösungen der verschiedenen Strategien in Abb. 0-13, so zeigt sich auch hier, daß das totale Ersetzen unterlegen ist. Für Bevölkerungen ab 1000 Individuen ist die 10%-Stategie, darunter die Zweikampfstrategie besser. Bei kleineren Bevölkerungen war auch die Güte der Individuen beim Zweikampfverfahren nur unwesentlich schlechter als bei der 10%-Strategie. Dies liegt wahrscheinlich daran, daß der Vorteil der 10%-Strategie, die Erhaltung der Vielfalt, sich vor allem bei großen Bevölkerungen, wo eine solche

Vielfalt erst gegeben ist, bezahlt macht. Für die Referenzbevölkerung von 1200 ist die 10%-Strategie in beiden Bereichen die bessere.

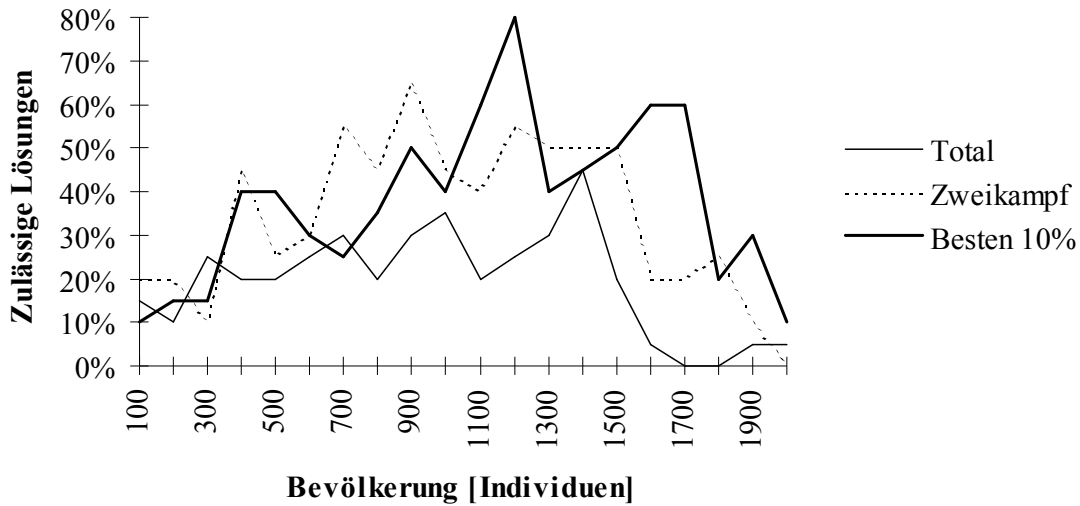

**Abb. 0-13:** Prozentualer Anteil zulässiger Lösungen der besten Individuen der letzten Generation bei verschiedenen Substitutions-Strategien.

## 4.4.2  Anzahl der Eltern

In Anlehnung an Ergebnisse aus der Literatur, wurde die Anzahl der Eltern pro Individuum nur bis vier getestet.[73] Grundsätzlich erhöht eine größere Anzahl an Eltern die Vielfalt an Kindern, was wiederum die Vielfalt an Individuen und damit Schemata erhöht. Daher ist eine größere Anzahl an Eltern, wie man aus Abb. 0-14 sehen kann, besonders bei kleineren Bevölkerungen von Vorteil.

Betrachtet man den prozentualen Anteil an zulässigen Lösungen in Abb. 0-15, so hat auch hier die Strategie einer größeren Anzahl an Eltern pro Individuum leichte Vorteile. Besonders die Strategie mit vier Eltern pro Individuum schneidet gut ab und ist auch die für die Referenzbevölkerung von 1200 die am besten geeignete.

---

[73] Vgl. Männer, 1994, S85-86.

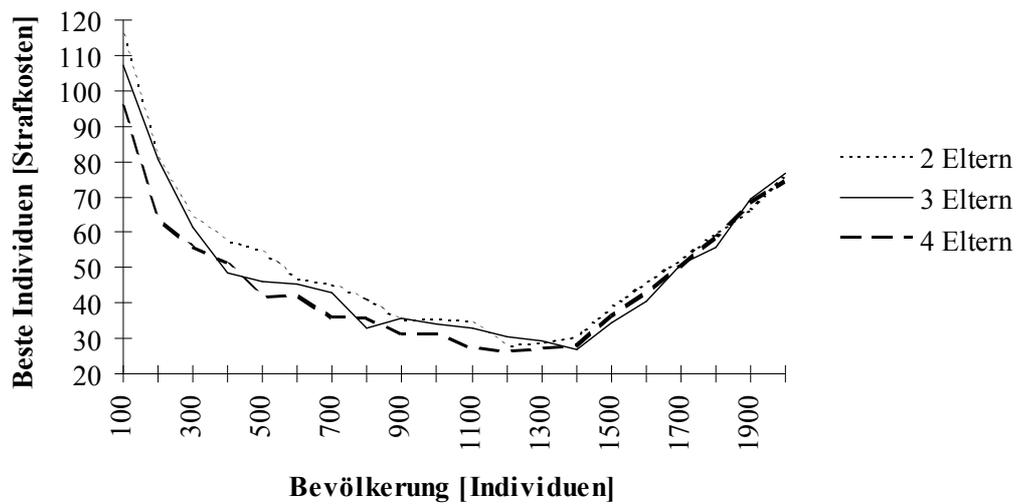

**Abb. 0-14:**   Durchschnittliche Güte des besten Individuums der letzten Generation bei Variation der Anzahl der Eltern pro Crossover.

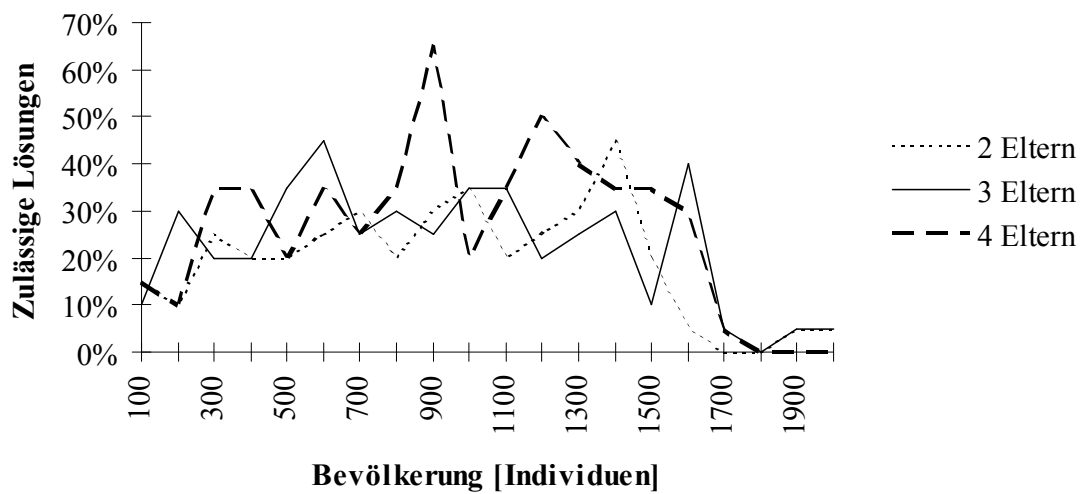

**Abb. 0-15:**   Prozentualer Anteil zulässiger Lösungen der besten Individuen der letzten Generation bei Variation der Anzahl der Eltern pro Crossover.

### 4.4.3   Art des Crossovers

In Abb. 0-16 ist ein Vergleich verschiedener Crossover-Strategien aufgezeichnet. Das uniforme Crossover schneidet klar am besten ab. Die Nachteile des 1- und 2-Punkt Crossovers sind die geringe Vielfalt der erzeugbaren Kinder. Gleiche gilt für das n-Punkt-Crossover. Hier kommt aber noch hinzu, daß die Länge eines Schematas, das von einem Elternteil an ein Kind weitergegeben werden kann, auf eins begrenzt ist. Damit werden gute Building-Blocks ständig aufgespalten und können sich nicht fortpflanzen. Das uniforme Crossover geht hier den goldenen Mittelweg, indem es sowohl kurze, wie auch lange Schematas bzw. Building-Blocks verarbeiten kann, ohne diese aufzuspalten und damit für eine maximale Vielfalt an Kindern sorgt.

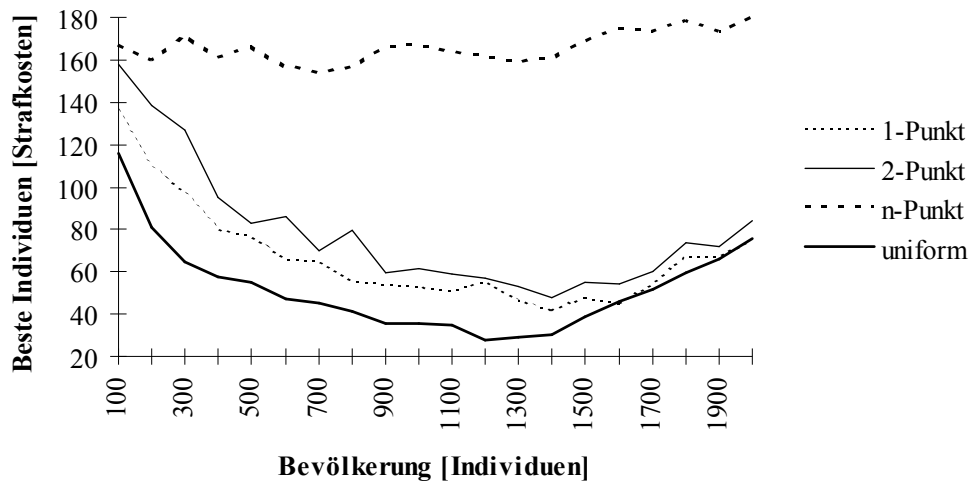

**Abb. 0-16:** Durchschnittliche Güte des besten Individuums der letzten Generation bei verschiedenen Crossover-Strategien.

Ein Vergleich des prozentualen Anteils an zulässigen Lösungen zeigt, daß uniformes und 1-Punkt Crossover am besten abschneiden. Besonders das gute Ergebnis des 1-Punkt-Crossovers deutet auf relativ große Building-Blocks hin, die nötig sind, um zulässige Lösungen zu erzielen. Dieser Gedanke wird in Kapitel 0 als segmentiertes Crossover weitergeführt.

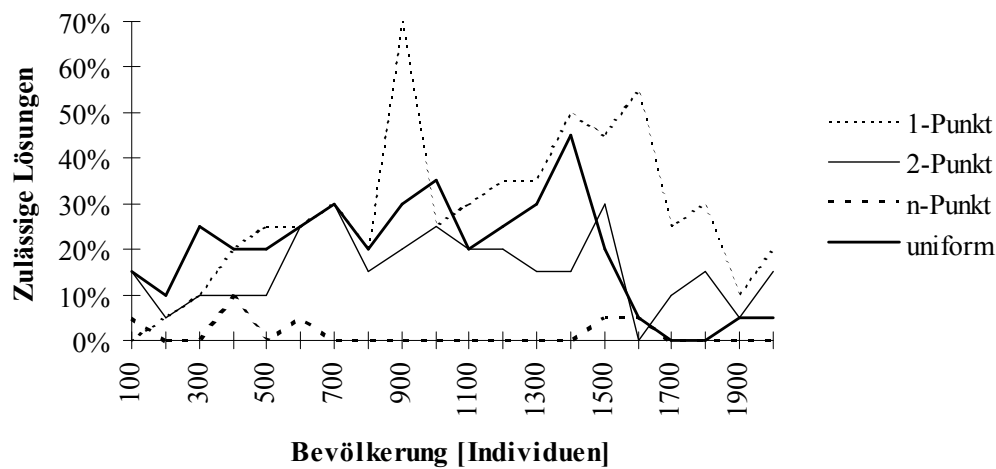

**Abb. 0-17:** Prozentualer Anteil zulässiger Lösungen der besten Individuen der letzten Generation bei verschiedenen Crossover-Strategien.

## 4.5    Zusammenfassung der bisherigen Ergebnisse

Faßt man die bisherigen Ergebnisse zusammen und setzt alle Parameter auf ihre separat ermittelten optimalen Werte, d.h. Mutationsrate p(M) = 1%, Reproduktionsrate p(C) = 2%, Gewicht der Nachfragedeckung G(Nachfrage) = 5, 10%-Substitutionsstrategie und uniformes Crossover mit vier Eltern, so erhält man die als optimiert bezeichnete Kurve in Abb. 0-18. Dieses Vorgehen ist jedoch nicht unproblematisch, da die gegenseitige Beeinflussung der Parameter bzw. Strategien nicht geklärt ist.[74]

Im Vergleich mit der alten Konfiguration schneidet die neue mit optimierten Parametern bzw. Strategien deutlich besser ab. Für Bevölkerungen zwischen 1200 und 1800 Individuen und einer Laufzeit von zwei Minuten liefert der genetische Algorithmus praktisch die optimale Lösung. Diese wird dabei oft schon nach ca. 30 bis 45 Sekunden erreicht.

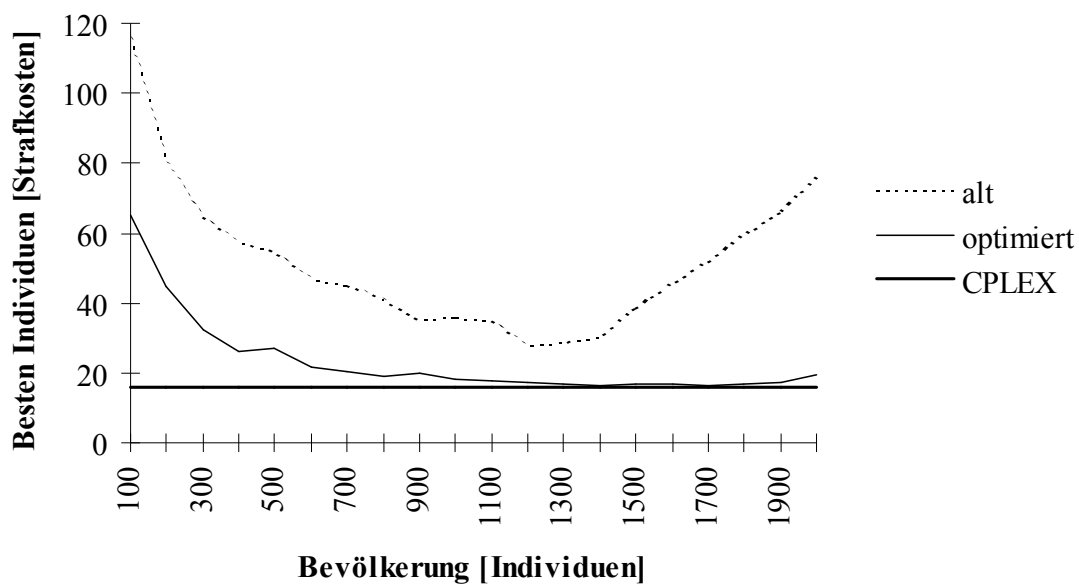

**Abb. 0-18:**    Vergleich der Güte der besten Individuen der letzten Generation mit und ohne Parameter-Optimierung mit den Ergebnissen von CPLEX.

Wie aus Abb. 0-19 ersichtlich wird bei einer Bevölkerung von 1400 Individuen in ca. 90% aller Fälle eine solche optimale und zulässige Lösung erreicht. Deutlich sieht man auch hier, daß die neue Konfiguration erheblich besser abschneidet als die alte.

---

[74] Durch die Komplexität dieses Vorhabens gibt auch keine entsprechenden Untersuchungen in der Literatur. Ein häufiger Vorschlag ist ein genetischer Meta-Algorithmus, der die Parameter bzw. Strategien für den eigentlichen genetischen Algorithmus festlegt, vgl. hierzu Bäck, 1996, 233ff. Dies verschiebt das Problem jedoch nur um eine Ebene nach oben.

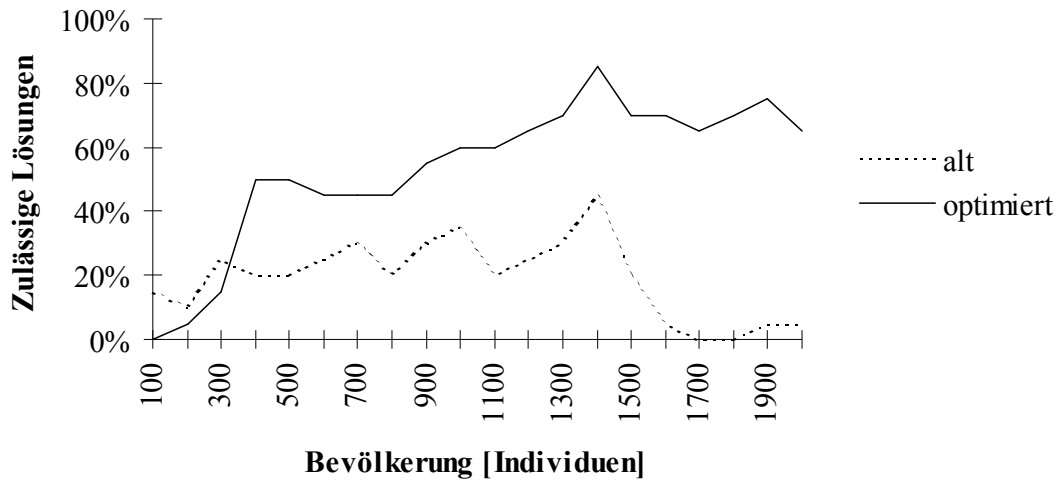

**Abb. 0-19:**   Vergleich des Prozentualen Anteils zulässiger Lösungen der besten Individuen der letzten
Generation mit und ohne Parameter-Optimierung.

Zusammenfassend läßt sich sagen, daß der genetische Algorithmus CPLEX nicht unterlegen ist,
sofern die Parameter bzw. Strategien optimiert sind. Ein Vergleich im Abschnitt 0 wird zeigen,
wie der genetische Algorithmus mit solchen Problemen, die mit CPLEX nicht lösbar sind,
umgeht.

# 4.6   Variationen der Eingabedaten

Grundsätzlich gibt es vier verschiedene Typen von Eingabedaten für eine Woche:

- Durchschnittliche Datenreihen, bei denen es mehrere mögliche Schichtpläne für die Station
  gibt. Ein solcher Fall war die in Kapitel 0 bis 0 behandelte Datenreihe. Hier ist der nicht
  modifizierte genetische Algorithmus zur Lösung ausreichend. CPLEX kann diese Probleme
  ebenfalls schnell lösen, ohne dabei jedoch mehrere gute und voneinander grundsätzlich
  verschiedene Lösungen liefern zu können. CPLEX könnte dies eventuell durch Basistausch
  von Nachbarlösungen erreichen, was nicht weiter untersucht worden ist. Lösungen innerhalb
  des effizienten Randes bleiben CPLEX dabei jedoch auf jeden Fall verborgen.

- Stark restriktive Datenreihen, die nur sehr wenige zulässige Schichtpläne für die Station
  ermöglichen. Dies kann z.B. durch Krankheit oder sonstigen Ausfall einer Krankenschwester
  geschehen. Aufgrund seiner Vorgehensweise löst CPLEX solche Probleme sehr schnell. Wie

Abb. 0-20 und Abb. 0-23 zeigen, ist der nicht modifizierte genetische Algorithmus hier nur sehr eingeschränkt erfolgreich.

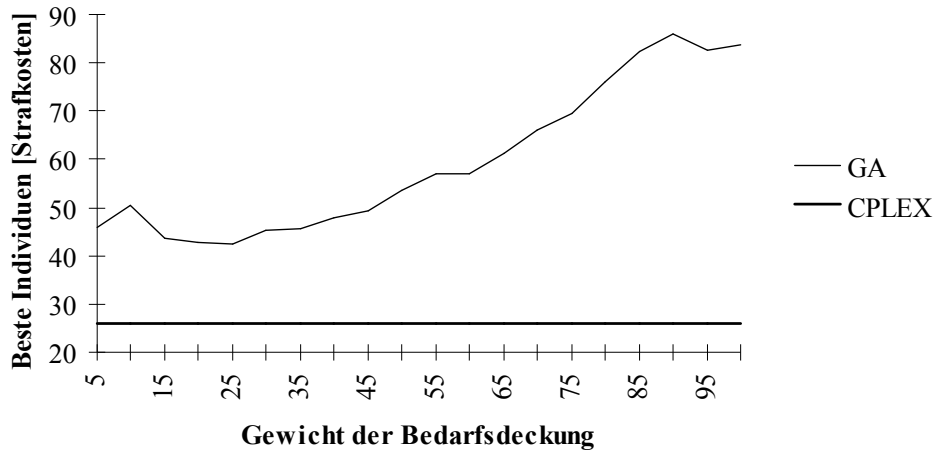

**Abb. 0-20:**    Vergleich der Lösungen durch CPLEX und durch den genetischen Algorithmus (GA) im Falle einer stark restriktiven Datenreihe und bei verschieden Gewichten der Bedarfsdeckung.

- Wenig restriktive Datenreihen, bei denen es eine große Vielzahl an zulässigen Lösungen gibt. Dies ist in Wochen mit einer leichten Überkapazität von Krankenschwestern der Fall. Wie man Abb. 0-21 und Abb. 0-23 entnehmen kann, hat der unmodifizierte genetische Algorithmus keinerlei Probleme mit diesem Typ. In fast 100% aller Fälle löst er das Problem optimal und liefert zusätzlich viele weitere gute Lösungen. CPLEX kann diese Probleme zwar auch lösen, hat aber nach zehn Minuten noch nicht den Branch-and-Bound-Vorgang beendet.

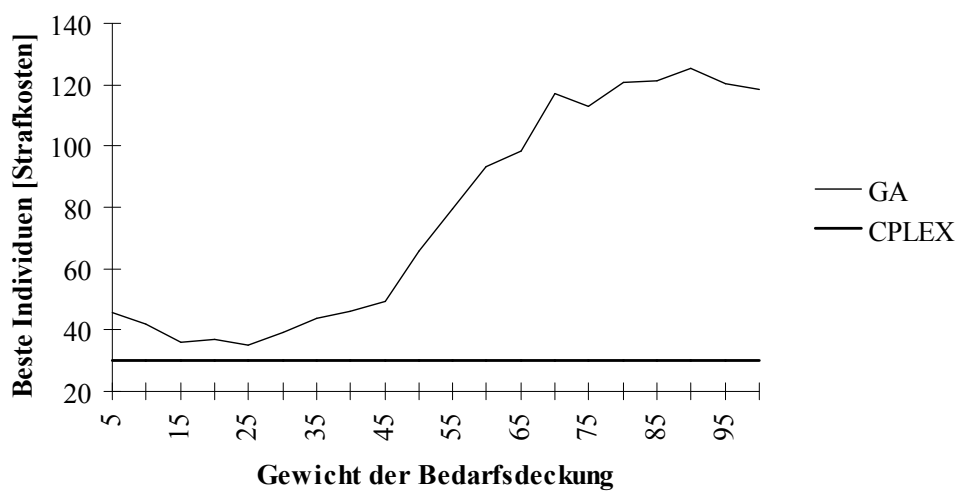

**Abb. 0-21:**    Vergleich der Lösungen durch CPLEX und durch den genetischen Algorithmus (GA) im Falle einer wenig restriktiven Datenreihe und bei verschieden Gewichten der Bedarfsdeckung.

- Datenreihen mit unterschiedlicher Nachfrage nach Krankenschwestern an unterschiedlichen Tagen. Dies kann z.B. durch große Operationen an manchen Tagen oder andere kurzfristige Nachfrageschwankungen verursacht werden. Wie man aus Abb. 0-22 und Abb. 0-23 entnehmen kann, stellt dies kein größeres Problem für den genetischen Algorithmus dar. Weitere Verbesserungen, insbesondere beim Anteil der zulässigen Lösungen, wären von Vorteil. Im Gegensatz dazu ist CPLEX nicht in der Lage, solche Probleme innerhalb von zehn Minuten zu lösen.

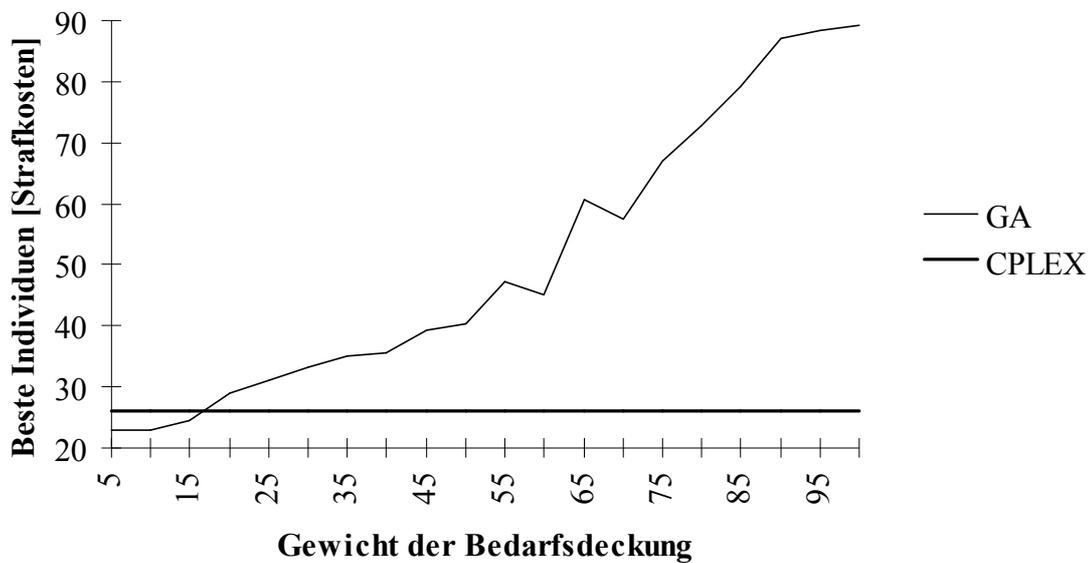

**Abb. 0-22:** Vergleich der Lösungen durch CPLEX und durch den genetischen Algorithmus (GA) im Falle einer im Krankenschwesternbedarf schwankenden Datenreihe und bei verschieden Gewichten der Bedarfsdeckung.

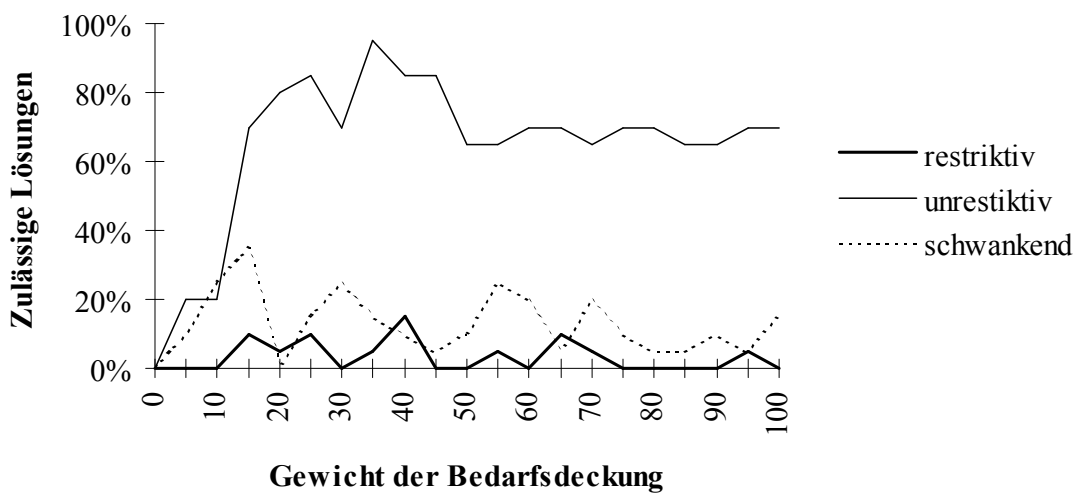

**Abb. 0-23:** Anteil zulässiger Lösungen an den besten Individuen der letzten Generation bei verschieden Datenreihentypen.

# 4.7    Weitere Vorgehensweise

Durch die in Kapitel 0 vorgenommen Modifikationen wird versucht, den genetischen Algorithmus weiter zu verbessern, damit alle in Kapitel 0 aufgeführten Fälle zufriedenstellend gelöst werden können. Dies soll in den Bereichen Lösungsqualität, Lösungsgeschwindigkeit und besonders im Bereich der Zulässigkeit erreicht werden. Ein Schwerpunkt wird dabei auf der Behandlung der Nebenbedingungen liegen.

Obwohl die besten durch den genetischen Algorithmus erzielten Lösungen die selbe oder sogar eine bessere Qualität als die von CPLEX ermittelten Lösungen besitzen, ist eine weitere Verbesserung der durchschnittlichen Lösungsqualität wünschenswert. Dies wird in Kapitel 0 erfolgen. Wie einzelne Testreihen gezeigt haben, ist der genetische Algorithmus in der Lage, optimale Ergebnisse innerhalb einer kurzen Laufzeit zu liefern. Dies soll durch weitere Verbesserungen in Kapitel 0 noch beschleunigt werden.

Wie man aus Abb. 0-20 bis Abb. 0-23 ersehen kann, hat das Gewicht der Nachfragedeckung einen hohen Einfluß auf den Anteil zulässiger Lösungen und auf die Lösungsqualität. Für die einzelnen Datenreihen ist das jeweils optimale Gewicht jedoch sehr unterschiedlich. Zur Auffindung des jeweils besten Gewichtes werden in Kapitel 0 daher sogenannte lernende Gewichte eingeführt.

Durch das bisherige zufällige Zerstückeln der Eltern wird die Erzeugung von zulässigen Kindern erschwert, indem gute größere Building-Blocks regelmäßig wieder verloren gehen. In Kapitel 0 wird daher aufbauend auf die Erfahrungen aus den verschiedenen Crossover-Strategien ein segmentiertes Crossover eingeführt, bei dem längere Building-Blocks erhalten bleiben können. Eine Weiterentwicklung dazu führt dann in Kapitel 0 zur Nischenbildung und damit zu mehreren konkurrierenden Unterbevölkerungen.

# 5 Verbesserungen des Grundprogrammes

## 5.1 Grundsätzliche Vorgehensweise der Verbesserungen

In Kapitel 0 soll durch eine Reihe von Maßnahmen die Güte des besten Individuums der letzten Generation, welche der vom Programm ausgegeben Lösung entspricht, verbessert werden. Dies läßt sich nicht vollständig von einer Verbesserung der Zulässigkeit der Lösungen (siehe Kapitel 0) abgrenzen, da durch den gewählten Strafkostenansatz eine Verbesserung der Güte der Lösungen durch eine Verbesserung deren Zulässigkeit erreicht wird.

Die Analyse der Verbesserungsvorschläge wird jeweils an zwei verschiedenen Datenreihen vorgenommen. Eine Datenreihe ist aus der Gruppe der restriktiven Datenreihen, im folgenden kurz als restriktive Datenreihe bezeichnet. In ihr sollen 22 Krankenschwestern eine Nachfrage nach neun statt acht tagsüber anwesenden Krankenschwestern der niedrigsten Qualifikationsstufe befriedigen, was nur wenig zulässige Stationsschichtpläne ermöglicht.

Die andere Datenreihe ist aus der Gruppe der Datenreihen mit schwankender Nachfrage nach Krankenschwestern und wird im folgenden kurz schwankende Datenreihe genannt. Insgesamt 21 Krankenschwestern sollen dabei Nachfragen nach tagsüber anwesenden Krankenschwestern der niedrigsten Qualifikationsstufe zwischen sieben und neun anstatt acht erfüllen. Nachts sollen entsprechen null bis zwei statt einer Krankenschwester der niedrigsten Qualifikationsstufe anwesend sein.

Alle Parameter, wie Mutations- und Reproduktionsrate, Anzahl der Eltern und die Art des Crossovers sind analog zu den Ergebnissen in Kapitel 0 gewählt. Für die Gewichte der Bedarfsdeckung gilt gemäß den Untersuchungen in Kapitel 0 $g_{restriktiv} = 25$ und $g_{schwankend} = 10$. Um zu besser vergleichbaren Ergebnissen zu gelangen, werden alle Analysen dieses Kapitels durch eine Elitismus-Strategie ergänzt. Dies bedeutet, daß die bislang beste erzielte Lösung stets in die nächste Generation übernommen wird, selbst wenn sie aufgrund der gewählten Substitutions-Strategie oder aus anderen Gründen ausscheiden würde.

Die Anwendung der Elitismus-Strategie ermöglicht ein neues Abbruchkriterium. Anstelle eines Zeitlimits, nach dem der Algorithmus bislang abgebrochen ist, läuft der Algorithmus jetzt solange weiter, bis 20 Generationen lang keine Verbesserung der bislang besten Lösung mehr erzielt wird. Das Limit von 20 Generationen wurde aufgrund bisheriger Erfahrungen mit dem Algorithmus gewählt.

Dieses neue Abbruchkriterium führt dazu, daß größere Bevölkerungen aufgrund deren Vielfalt an Individuen kleineren grundsätzlich überlegen sind. Wie man in Abb. 0-1 sehen kann, sind jedoch die Laufzeiten für größere Bevölkerungen erheblich länger. Ein geeigneter Kompromiß ist eine Bevölkerung um 1300. Hier werden bei einer vertretbaren Laufzeit beinahe die insgesamt besten Werte erzielt.

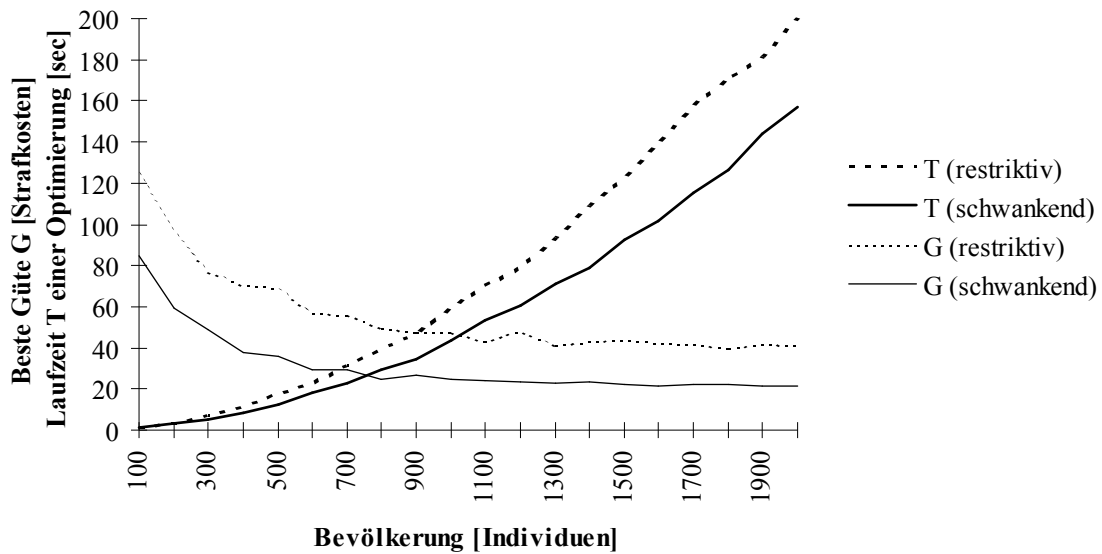

**Abb. 0-1:**   Durchschnittliche Laufzeit (T) einer kompletten Optimierung und durchschnittliche Güte (G) der besten Individuen der letzten Generation für eine restriktive und eine schwankende Datenreihe.

# 5.2   Verbesserung der Güte der Individuen

## 5.2.1   Variation der Substitutions-Strategien

Zunächst soll überprüft werden, ob eine Verbesserung der Lösungsgüte durch eine Variation der konventionellen Substitutions-Strategien möglich ist. Als erstes soll das bislang erfolgreichste Verfahren, das Beibehalten der 10% besten Eltern, variiert werden. Anstatt der 10% besten Eltern werden nun die 5% bzw. 20% besten Eltern beibehalten und der Rest der Bevölkerung durch Kinder ersetzt. Bei diesen Verfahren ist die Anwendung der Elitismus-Strategie bedeutungslos, da das beste Eltern-Individuum automatisch überlebt.

Betrachtet man die drei Strategien für die im Bedarf schwankende Datenreihe in Abb. 0-2, so zeigt sich, daß die Unterschiede in der erzielten Güte der Individuen nur sehr gering sind. Ein Vergleich des durchschnittlichen Anteils an zulässigen Lösungen über alle Bevölkerungsgrößen in Tabelle 0-1 (als Durchschn. Zulässigkeit bezeichnet) zeigt jedoch, daß die 20%-Strategie die meisten zulässigen Lösungen erzeugt.

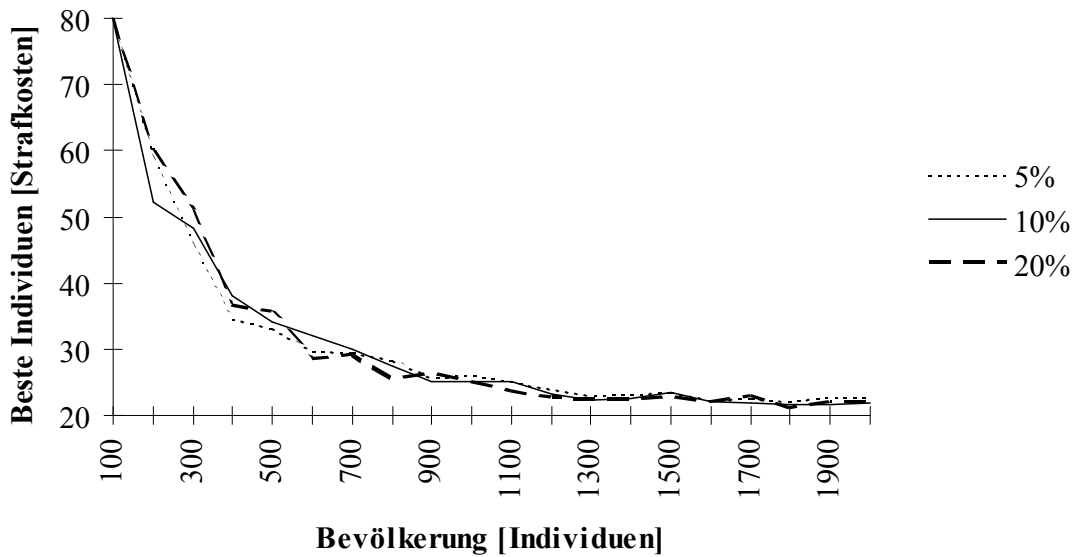

**Abb. 0-2:**   Durchschnittliche Güte der besten Individuen der letzten Generation bei einer im Bedarf schwankenden Datenreihe und Variation des Prozentsatzes der zu übernehmenden Eltern.

Wendet man die selben drei Strategien auf die restriktive Datenreihe an, so zeigt Abb. 0-3, daß die 20%-Strategie für die meisten Bevölkerungsgrößen die besten Ergebnisse liefert. Dies bestätigt auch der Durchschnitt der Güte über alle Bevölkerungen in Tabelle 0-2. Bezüglich des Anteils an zulässigen Lösungen über alle Bevölkerungen erweist sich keine der drei Strategien als den anderen überlegen.

Die zweite Gruppe der untersuchten Strategien bilden die Zweikampfstrategien. Die in Kapitel 0 angewandte Zweikampfstrategie, bei der jedes Kind um den Platz in der neuen Generation gegen ein eigenes Eltern-Individuum antritt, wird so modifiziert, daß nur ein gewisser Prozentsatz an Kindern einen Zweikampf antreten muß. Alle anderen Kinder kommen automatisch weiter. Auf diese Weise sollen vier Substitutions-Strategien gebildet werden, die einen 0%, 50%, 75% und 100% Zweikampfanteil besitzen.

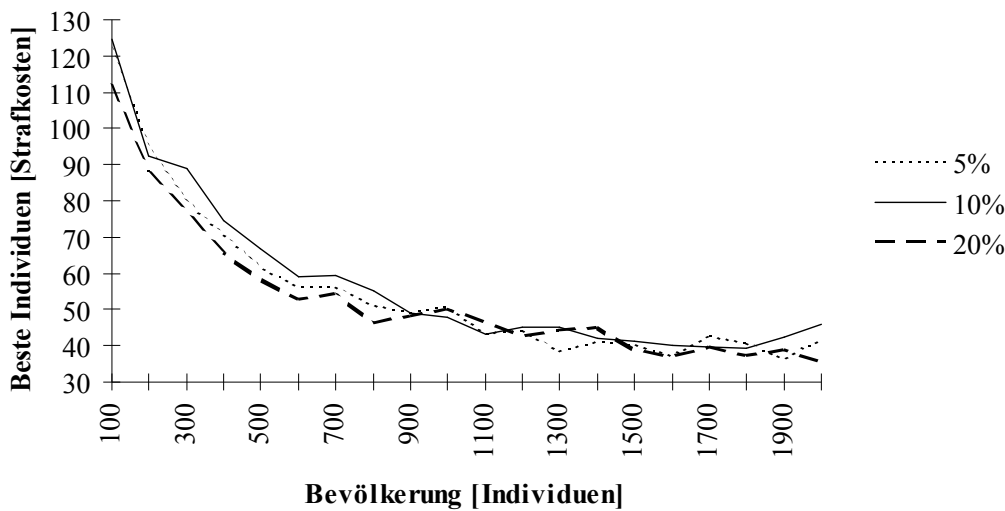

**Abb. 0-3:** Durchschnittliche Güte der besten Individuen der letzten Generation bei einer restriktiven Datenreihe und Variation des Prozentsatzes der zu übernehmenden Eltern.

Die 0% Zweikampf-Strategie entspricht dabei dem totalen Ersetzten der Eltern-Generation. Die 100% Zweikampf-Strategie entspricht der in Kapitel 0 angewendeten Zweikampf-Strategie. Außer bei dieser kommt hier die Elitismus-Strategie zu tragen. Das beste Eltern-Individuum kommt auf jeden Fall weiter, ob es an einem Zweikampf teilnimmt oder nicht. Auch falls es an einem Zweikampf teilnimmt und diesen gegen ein besseres Kind verliert, kommt es weiter.

Wie man Abb. 0-4 und Tabelle 0-1 entnehmen kann, bei der im Bedarf schwankenden Datenreihe die 100% Zweikampf-Strategie sowohl für die Güte der Individuen, wie auch für den Anteil an zulässigen Lösungen die am besten geeignete.

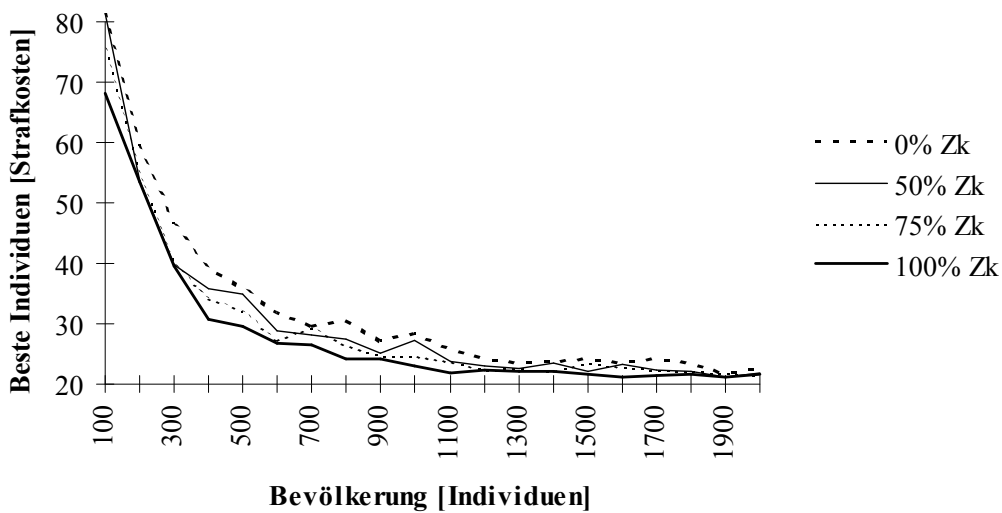

**Abb. 0-4:** Durchschnittliche Güte der besten Individuen der letzten Generation bei einer im Bedarf schwankenden Datenreihe und Variation des Prozentsatzes der am Zweikampf (Zk) teilnehmenden Eltern mit gleichzeitigem Elitismus.

Vergleicht man die vier Strategien für die restriktive Datenreihe in Abb. 0-5 und Tabelle 0-2, so ist das Ergebnis nicht eindeutig. Die 75% Zweikampf-Strategie erzielt die im Schnitt beste Güte der Individuen. Die 50% Strategie liefert hingegen im Durchschnitt etwa doppelt so viele zulässige Lösungen. Die 100%-Strategie liegt für beide Kriterien jeweils dazwischen.

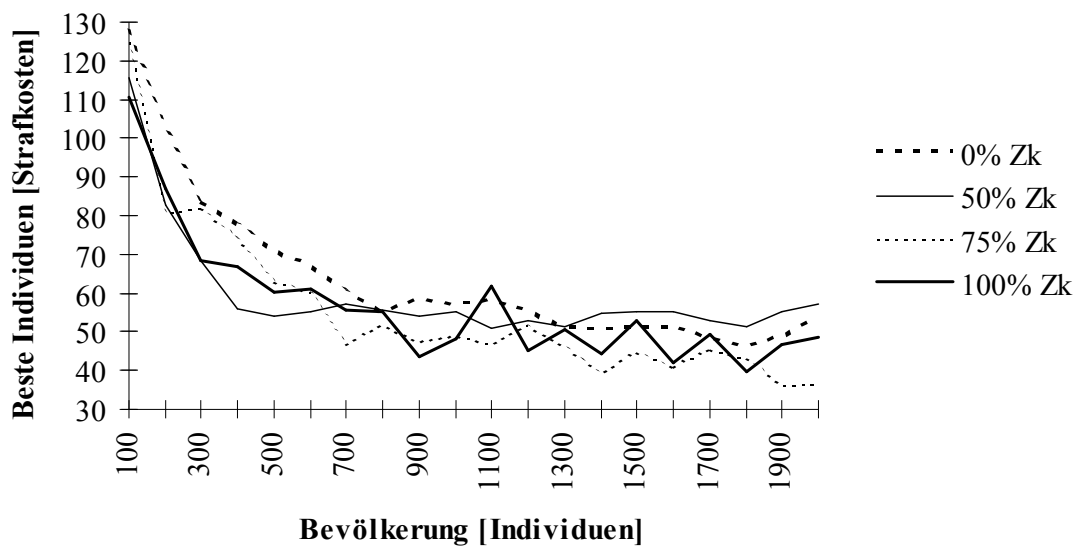

**Abb. 0-5:** Durchschnittliche Güte der besten Individuen der letzten Generation bei einer restriktiven Datenreihe und Variation des Prozentsatzes der am Zweikampf (Zk) teilnehmenden Eltern mit gleichzeitigem Elitismus.

Vergleicht man die Ergebnisse mit dem in Kapitel 0 angesprochenen Konflikt zwischen einer tief und einer breit angelegten Lösungssuche, so wird der Konflikt hier bestätigt. Die Strategien, bei denen die X% Besten der Eltern übernommen werden, übertreffen in der erzielten Lösungsgüte die Zweikampf-Strategien. Es handelt sich dabei um eine tiefe Suche, die um wenige Individuen stattfindet. Die in die Breite suchenden Zweikampf-Strategien weisen dagegen einen höheren Anteil an zulässigen Lösungen auf. Dies gilt insbesondere bei der restriktiven Datenreihe, welche insgesamt nur wenige zulässige Lösungen aufweist und zu deren Lösung deshalb eine breit angelegte Suche hilfreich ist.

|  | 0% | 50% | 75% | 100% | 5% | 10% | 20% |
|---|---|---|---|---|---|---|---|
| Durchschnittliche Güte | 32,3 | 30,4 | 29,7 | 28,2 | 31,1 | 30,9 | 31,1 |
| Durchschn. Zulässigkeit | 14% | 16% | 18% | 21% | 13% | 17% | 20% |

**Tabelle 0-1:** Zusammenfassung der Ergebnisse aller Substitutions-Strategien für die schwankende Datenreihe.

|  | 0% | 50% | 75% | 100% | 5% | 10% | 20% |
|---|---|---|---|---|---|---|---|
| Durchschnittliche Güte | 63,9 | 59,9 | 55,5 | 56,9 | 55,1 | 57,1 | 53,0 |

| Durchschn. Zulässigkeit | 4% | 17% | 9% | 14% | 11% | 10% | 10% |
|---|---|---|---|---|---|---|---|

**Tabelle 0-2:** Zusammenfassung der Ergebnisse aller Substitutions-Strategien für die restriktive Datenreihe.

Die Überlegung, daß möglichst vielseitige Eltern-Individuen bessere Kindern erzeugen, da lokale Optima dadurch besser vermieden werden können, führt zu zwei weiteren Substitutions-Strategien. Bei der „3-Leben"-Strategie wird ein Eltern-Individuum erst nach drei verlorenen Zweikämpfen ersetzt. Dies soll ein zu schnelles Konvergieren zu den besten Individuen hin verhindern. Ansonsten entspricht die Strategie der 100% Zweikampf-Strategie.

Die „Abstand"-Strategie versucht ebenfalls, möglichst unterschiedliche Individuen in die nächste Generation gelangen zu lassen. Dazu wird für alle Einzelmerkmale der Durchschnitt des Schichtmusterindex über die gesamte Bevölkerung berechnet. Alle Durchschnitte zusammen ergeben einen Art Schwerpunkt aller Individuen im Lösungsraum. Es kommen nun jeweils die Hälfte aller Eltern und Kinder in die nächste Generation, die von diesem Schwerpunkt am weitesten entfernt sind. Die Gesamtentfernung ergibt sich dabei als Summe der Einzelentfernungen vom jeweiligen Durchschnitt.

Als dritte Strategie wird eine Mischung aus einer 75% Zweikampf-Strategie und einer Strategie, bei der die Besten 5% der Eltern weiterkommen, gewählt. Die Strategie ist so aufgebaut, daß die Besten 5% der Eltern automatisch, alle anderen Eltern-Individuen mit einer Wahrscheinlichkeit von 75% nur nach einem gewonnenen Zweikampf mit einem Kind weiterkommen. Zu 25% werden sie ohne Zweikampf durch ein Kind ersetzt.

In Abb. 0-6 sind die drei Strategien für die im Bedarf schwankende Datenreihe aufgetragen. Dabei fällt auf, daß die Mischungs-Strategie sehr gut, die beiden anderen sehr schlecht abschneiden. Ein Vergleich des durchschnittlichen prozentualen Anteils der zulässigen Lösungen in Tabelle 0-3 bestätigt dies nochmals.

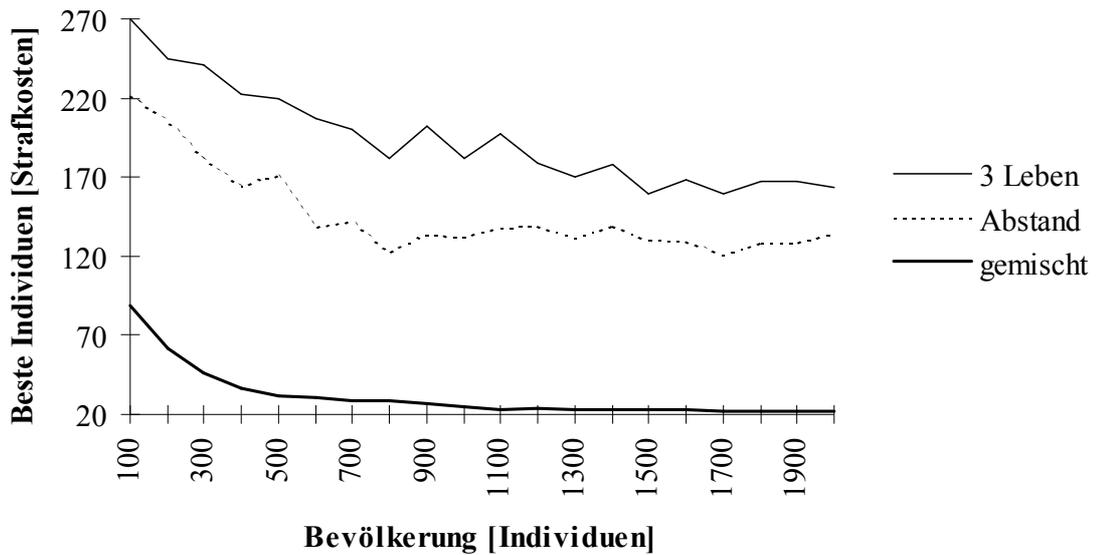

**Abb. 0-6:**   Durchschnittliche Güte der besten Individuen der letzten Generation bei einer schwankenden
Datenreihe und besonderen Substitutions-Strategien.

Für die restriktive Datenreihe ergibt sich in Abb. 0-7 ein zu obigem entsprechendes Ergebnis.
Auch hier schneidet die gemischte Strategie sehr gut, die beiden anderen Strategien sehr
schlecht ab.

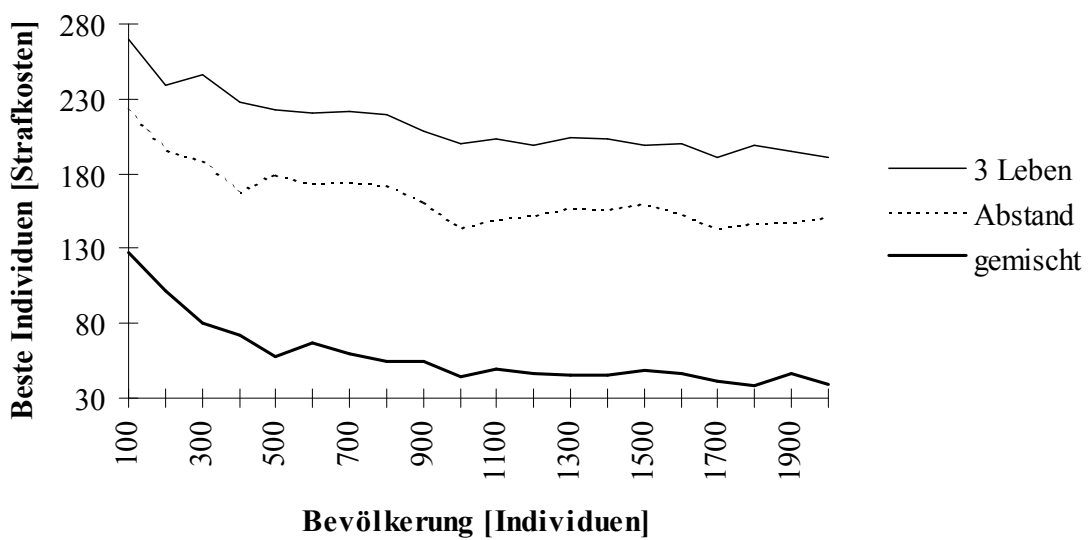

**Abb. 0-7:**   Durchschnittliche Güte der besten Individuen der letzten Generation bei einer restriktiven
Datenreihe und besonderen Substitutions-Strategien.

Zusammenfassend läßt sich sagen, daß ein Versuch, die Vielfalt über die gewählten Kriterien zu erhalten nicht zu guten Ergebnissen führt. Wahrscheinlich ist das Vielfaltskriterium weniger wichtig als zunächst angenommen, da wie in Kapitel 0 erläutert, eine Bevölkerung aus N Individuen zwischen $k^L$ und $N \cdot k^L$ verschiedene Schemata enthalten kann.

Für ein geschätztes k von k = 40 durchschnittlich möglichen Schichtmustern pro Krankenschwester und einem L in Höhe der Anzahl der Krankenschwestern pro Station von L = 20, ergibt dies auch in ungünstigen Fällen, d.h. für $k^L$, eine ausreichend große Zahl an verschiedenen Schemata. Die durch die spezielle Kodierung eingeschränkten genetischen Operatoren vermindern diese Anzahl nochmals um eine unbekannte Größe. Die verbleibende Anzahl an unterschiedlichen Schemata ist obigen Ergebnissen zufolge auch bei den herkömmlichen Substitutions-Strategien ausreichend.

Weiterhin zeigt sich, daß eine einfache Kombination zweier Strategien nicht unbedingt deren beider Vorteile vereint. So liegen bei beiden Datenreihen die durch die Kombination erzielten Werte für die durchschnittliche Güte bzw. den Anteil an zulässigen Lösungen unter den Ergebnissen der jeweils besseren Einzel-Strategie. Insgesamt sind die 100% Zweikampf- und die 20% der Besten Eltern-Strategie für beide Datenreihen am erfolgreichsten.

|  | Restriktive Datenreihe | | | Schwankende Datenreihe | | |
|---|---|---|---|---|---|---|
|  | 3 Leben | Abstand | Mix | 3 Leben | Abstand | Mix |
| Durchschnittliche Güte | 212,9 | 164,8 | 58,3 | 194,6 | 147,2 | 31,4 |
| Durchschn. Zulässigkeit | 1% | 2% | 9% | 1% | 3% | 18% |

**Tabelle 0-3:** Zusammenfassung der Ergebnisse bei besonderen Substitution-Strategien.

## 5.2.2  Intelligente Mutation

Eine weitere Verbesserung der Güte der Individuen soll durch einen veränderten Mutations-Operator erreicht werden. Schlechtere Individuen sollen dazu häufiger mutiert werden als bessere und zusätzlich soll sich die Mutationsrate dem Optimierungsprozeß anpassen. Diese neue Art der Mutation soll intelligente Mutation genannt werden.[75]

---

[75]  Vgl. Chambers, 1995, S.236 und S. 267f.

Die zugrunde liegende Idee ist, daß Individuen, die bereits viele Restriktionen erfüllen und daher niedrige Strafkosten aufweisen, durch eine zufällige Mutation selten verbessert, sondern meistens verschlechtert werden. Bei schlechteren Individuen, die mehrere Restriktionen nicht erfüllen, besteht dagegen eine bessere Chance, diese durch eine zufällige Mutation zu verbessern.

Die Mutationsrate wird somit zu einer Funktion der Bedarfsdeckung der Individuen. Für die folgenden Untersuchungen wird sie so festgelegt, daß Individuen, die alle Restriktionen erfüllen, nur noch die halbe ursprüngliche Mutationswahrscheinlichkeit, d.h. 0,5%, besitzen. Für jede Bedarfsunterdeckung um eine Krankenschwesterschicht steigt sie um 0,5% an.

Außerdem soll sich die Mutationsrate an den Verlauf der Optimierung anpassen. Dies soll dadurch erreicht werden, daß bei Konvergenz des Algorithmus die Mutationsrate ansteigt, um zu überprüfen, ob nicht doch noch bessere Individuen möglich sind. Dazu steigt die Mutationsrate für alle Individuen pro Generation, in der es keine Verbesserung des besten Individuums mehr gab um je 0,05%. Kommt es doch noch zu einer weiteren Verbesserung, so wird die Mutationsrate wieder auf ihren ursprünglichen Startwert zurückgesetzt.

In Abb. 0-8 wird die intelligente mit der normalen Mutation für eine restriktive (R) und eine schwankende (S) Datenreihe verglichen. Dabei fällt auf, daß die Unterschiede zwischen beiden Mutationsarten nur sehr gering sind.

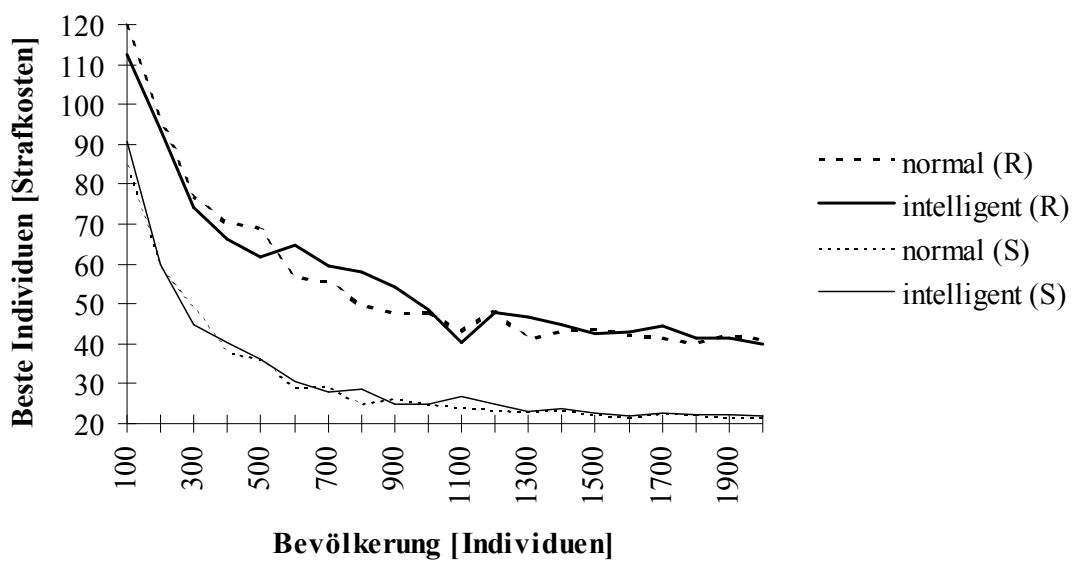

**Abb. 0-8:**    Durchschnittliche Güte der besten Individuen der letzten Generation bei intelligenter und normaler Mutation für eine restriktive (R) und eine schwankende (S) Datenreihe.

Vergleicht man die durchschnittlich erzielte Güte beider Verfahren in Tabelle 0-4, so ist die intelligente Mutation etwas schlechter. Dies ist auf den insgesamt höheren Anteil an Mutationen zurückzuführen, wodurch meistens sehr schlechte Individuen erzeugt werden. Bei der besten Güte und dem durchschnittlichen Anteil an zulässigen Lösungen weist die intelligente Mutation Vorteile auf. Dies zeigt, daß die obigen Überlegungen grundsätzlich richtig sind. Ein Ansatz zur Verbesserung der intelligenten Mutation wäre anstelle eines zufällig gewählten Elementes, gezielt dasjenige Element, das für die Verletzung der meisten Restriktionen verantwortlich ist, zu ersetzen. Zusätzlich könnte man dieses Element anstatt durch ein beliebiges durch ein möglichst gut dafür geeignetes ersetzen.

| | Restriktive Datenreihe | | Schwankende Datenreihe | |
|---|---|---|---|---|
| | intelligent | standard | intelligent | standard |
| Durchschnittliche Güte | 56,5 | 56,0 | 32,0 | 31,3 |
| Durchschn. Zulässigkeit | 11% | 10% | 15% | 14% |

**Tabelle 0-4:** Zusammenfassung der Ergebnisse bei intelligenter und normaler Mutation.

### 5.2.3  Re-Initialisierung

Eine einfache Möglichkeit, die Vielfalt an Schemata in der Bevölkerung zu erhöhen, ist das Erzeugen zufälliger Kinder. Dieses Verfahren soll Re-Initialisierung genannt werden. Dazu wird ein gewisser Prozentsatz an Kindern nicht wie üblich mittels Eltern, sondern zufällig wie die Individuen der Startbevölkerung erzeugt. Anschließend werden sie wie normale Kinder in die Bevölkerung aufgenommen.

Abb. 0-9 zeigt einen Vergleich der durchschnittlich besten Individuen mit und ohne Re-Initialisierung für eine schwankende (S) und eine restriktive Datenreihe (R). Bei der Re-Initialisierung werden 1% der Kindern zufällig eingefügt.

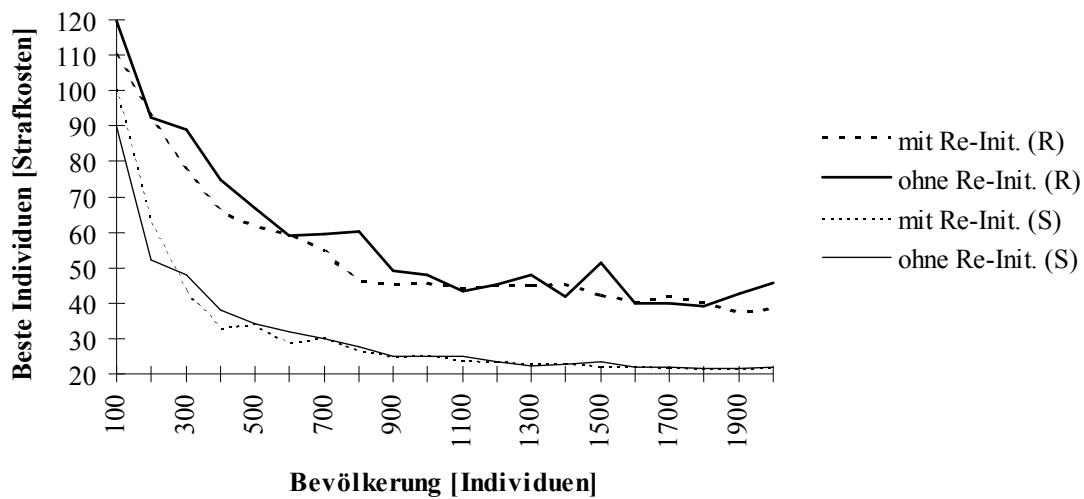

**Abb. 0-9:** Vergleich der Güte der besten Individuen der letzten Generation mit und ohne Re-Initialisierung für eine restriktive (R) und eine schwankende (S) Datenreihe.

Die Unterschiede zwischen beiden Verfahren sind für beide Datenreihen gering. Insbesondere kleine Bevölkerungen profitieren etwas durch die mittels Re-Initialisierung größere Vielfalt an Schemata. Das Kriterium der Vielfalt, wie bereits in Kapitel 0 für die besonderen Substitutions-Strategien gesehen, ist von geringer Bedeutung. Da die durchschnittliche Zulässigkeit durch Anwendung der Re-Initialisierung zurückgeht (vgl. Tabelle 0-5), ist deren Anwendung insgesamt nicht empfehlenswert.

| | Restriktive Datenreihe | | Schwankende Datenreihe | |
|---|---|---|---|---|
| | mit Re-Init. | ohne Re-Init. | mit Re-Init. | ohne Re-Init. |
| Durchschnittliche Güte | 53,9 | 57,8 | 31,8 | 31,4 |
| Durchschn. Zulässigkeit | 7% | 10% | 15% | 17% |

**Tabelle 0-5:** Zusammenfassung der Ergebnisse mit und ohne Re-Initialisierung.

## 5.2.4 Lokale Optimierung

Um eine weitere Verbesserung der Lösungsgüte der besten Individuen der letzten Generation zu erreichen, soll ein der manuellen Lösung ähnliches Verfahren angewendet werden. Bei der manuellen Lösung werden zuerst Schichtpläne erstellt, die möglichst alle Bedarfs-Restriktionen erfüllen, ohne die Wünsche der Krankenschwestern überhaupt zu berücksichtigen. Anschließend wird durch paarweises Vertauschen der Arbeitstage der Krankenschwestern versucht, deren Wünsche möglichst gut zu integrieren.

Durch die Problemrepräsentation ist ein paarweises Vertauschen von Arbeitstagen nicht möglich. Statt dessen sollen paarweise Schichtmuster getauscht werden, falls beide Krankenschwestern die selbe Qualifikationsstufe und Wochenarbeitszeit besitzen. Diese Einschränkungen sind notwendig, um die durch die spezielle Kodierung erzielte implizite Erfüllung der Wochenarbeitszeit-Nebenbedingung beizubehalten. Außerdem wird dadurch sichergestellt, daß die neue Lösung den Bedarf auf allen Qualifikationsstufen genauso gut deckt wie die alte.

Ein solches internes Vertauschen soll nur dann durchgeführt werden, wenn die Summe der Schichtmuster-Strafkosten beider am Tausch beteiligter Krankenschwestern dadurch geringer wird. Ein Ringtausch zwischen drei Krankenschwestern ist ebenfalls möglich. Damit nicht alle Individuen für eine Krankenschwester durch Vertauschungen das selbe Schichtmuster aufweisen, werden die internen Vertauschungen nur mit einer Wahrscheinlichkeit von 10% durchgeführt.

In Abb. 0-10 ist für eine schwankende und in Abb. 0-11 für eine restriktive Datenreihe ein Vergleich bei verschiedener Tauschintensitäten aufgezeichnet. Bei der mit Tausch = 0 bezeichneten Kurve wird auf interne Vertauschungen komplett verzichtet. Bei Tausch = 1 bzw. Tausch = 5 wird eine interne Vertauschung nur bei solchen Individuen in Erwägung gezogen, die maximal eine bzw. fünf Nebenbedingungen verletzen. Alle Individuen nehmen für Tausch = max. an den internen Vertauschungen teil.

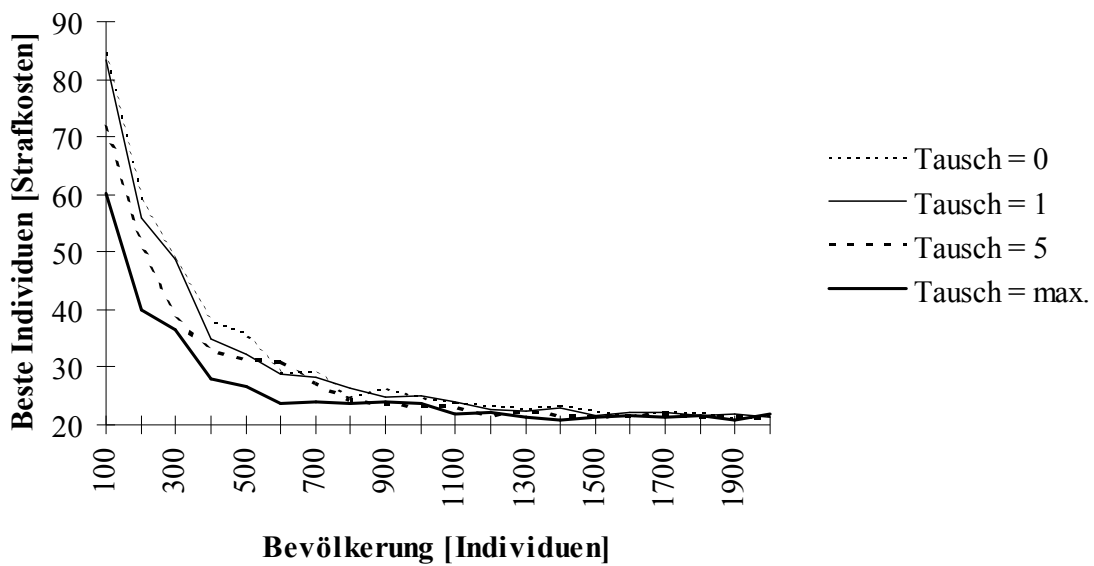

**Abb. 0-10:** Durchschnittliche Güte des besten Individuums der letzten Generation bei verschiedenen internen Tauschintensitäten für eine schwankende Datenreihe.

Für die schwankende Datenreihe ist die maximale Tauschintensität eindeutig von Vorteil. Besonders für kleine Bevölkerungen kann sie die durchschnittliche Güte der besten Individuen der letzten Generation deutlich verbessern. Bei einer um 5% gestiegen Gesamtlaufzeit erzielt der Algorithmus durchschnittlich um 20% bessere Lösungen (vgl. Tabelle 0-6). Die durchschnittliche Zulässigkeit dieser Individuen ist von 14% auf 25% gestiegen.

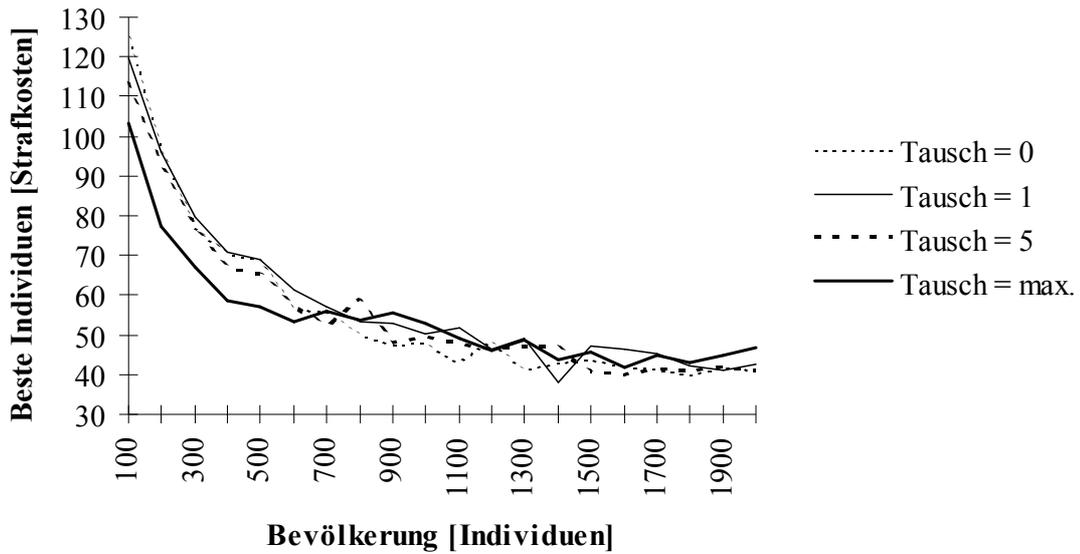

**Abb. 0-11:**    Durchschnittliche Güte des besten Individuums der letzten Generation bei verschiedenen internen Tauschintensitäten für eine restriktive Datenreihe.

Für die restriktive Datenreihe ist das Ergebnis weniger eindeutig. Zwar kann die maximale Tauschintensität die Güte der besten Individuen der letzten Generation für kleine Bevölkerungen wieder stark verbessern, für große Bevölkerungen ist dieses Verfahren jedoch schlechter als die anderen. Auch die durchschnittliche Zulässigkeit steigt nur von 10% auf 11%. Offensichtlich verhindern die insgesamt nur wenigen zulässigen Lösungen einen Erfolg des internen Vertauschens.

|  | Restriktive Datenreihe | | | | Schwankende Datenreihe | | | |
|---|---|---|---|---|---|---|---|---|
| Tauschintensität | 0 | 1 | 5 | max. | 0 | 1 | 5 | max. |
| Durchschnittliche Güte | 56,0 | 58,0 | 55,8 | 54,4 | 31,3 | 30,5 | 28,6 | 26,3 |
| Durchschn. Laufzeit | 79,7 sec | 82,3 sec | 83,3 sec | 85,1 sec | 59,0 sec | 61,4 sec | 61,7 sec | 63,2 sec |
| Durchschn. Zulässigkeit | 10% | 8% | 11% | 11% | 14% | 15% | 21% | 25% |

**Tabelle 0-6:**    Zusammenfassung der Ergebnisse für verschiedene interne Tauschintensitäten.

Ein Ansatz zur Verbesserung des internen Vertauschens wäre der Verzicht auf die Beschränkung, daß alle am Tausch beteiligten Krankenschwestern der selben Qualifikationsstufe angehören müssen. Dies verschlechtert unter Umständen jedoch die Bedarfserfüllung. Eine weitere Erweiterungsmöglichkeit wäre, analog zur manuellen Einsatzplanerstellung, nur einzelne Tage anstatt ganze Schichtmuster zu tauschen. Dazu wäre jedoch eine neue Problemkodierung nötig.

## 5.2.5   Zwei oszillierende Zielfunktionen

Als letzte Möglichkeit zur Verbesserung der Güte der Individuen soll eine Variation der bisher verwendeten Zielfunktion untersucht werden. Statt wie bisher eine Zielfunktion sollen zwei Zielfunktionen, die sich in den Gewichten für die Bedarfsdeckung unterschieden, zum Einsatz kommen. Jede der Zielfunktionen wird dabei für eine bestimmte Anzahl von Generationen benutzt. Anschließend wird auf die andere Zielfunktion gewechselt.

Dieses Oszillieren zwischen den beiden Zielfunktionen soll bewirken, daß der Algorithmus jeweils besonders stark versucht, eines der beiden Kriterien aus Bedarfserfüllung und Wünschen der Krankenschwestern zu erfüllen. Ist eine momentan gute Lösung für ein Kriterium gefunden, soll sich der Algorithmus mit dem anderen auseinandersetzten. Setzt man dazu z.B. das eine Gewicht der Bedarfsdeckung sehr niedrig und das andere sehr hoch an, so versucht der Algorithmus zuerst möglichst den Bedarf zu erfüllen. Wechselt die Zielfunktion, versucht er möglichst gut die Wünsche der Krankenschwestern zu erfüllen usw.

Durch den speziellen Charakter der oszillierenden Zielfunktionen kommt es zu einem ständigen Wechsel des besten Individuums. Abwechselnd wird dies ein Individuum, das den Bedarf gut erfüllt und ein Individuum das die Wünsche der Krankenschwestern stark berücksichtigt, sein. Daher ist das bisher verwendete Abbruchkriterium, 20 Generationen ohne einen Wechsel des besten Individuums, ungeeignet. Statt dessen wird als Abbruchkriterium zwei Minuten Laufzeit pro Optimierung festgelegt. Dies macht einen direkten Vergleich mit den anderen Verfahren dieses Kapitels schwierig.

In Abb. 0-13 und Abb. 0-12 sind die Ergebnisse der oszillierenden Zielfunktionen für eine schwankende bzw. restriktive Datenreihe dargestellt. Im Einzelnen wurde jeweils eine Oszillationsphase mit dem bisherigen Gewicht der Bedarfsdeckung, d.h. $g_{restriktiv}$ = 25 und $g_{schwankend}$ = 10, die andere mit g = 1 durchgeführt. Die erste Phase legt also Wert auf die Bedarfserfüllung, während die zweite die Wünsche stärker berücksichtigt. Für die Anzahl der Wiederholungen der einzelnen Oszillierungsphasen wurden die Verhältnisse 1 zu 1, 5 zu 1 und 5 zu 5 gewählt.

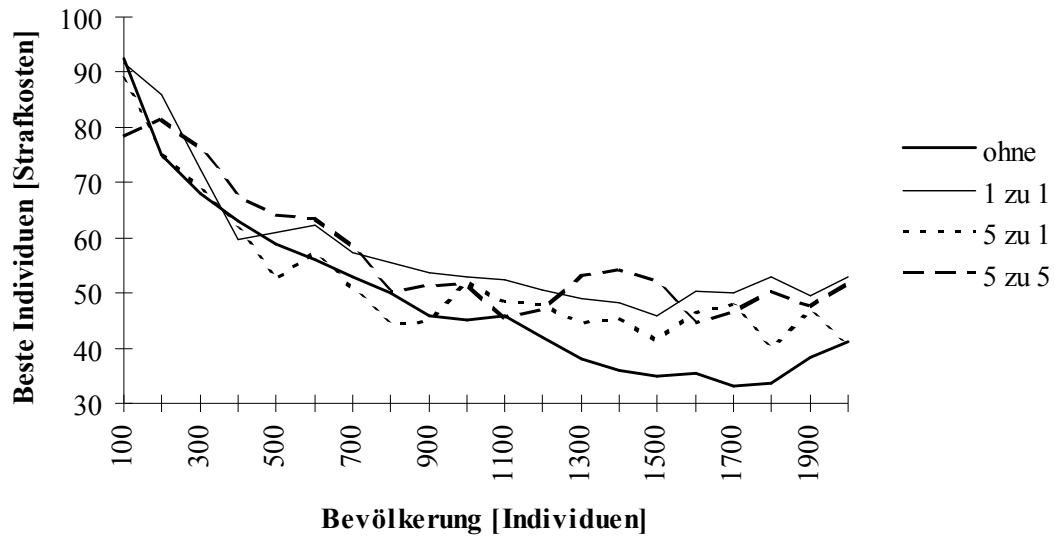

**Abb. 0-12:** Vergleich der durchschnittlichen Güte des besten Individuums der letzten Generation bei verschieden oszillierenden Zielfunktionen und zwei Minuten Gesamtlaufzeit pro Optimierung für eine schwankende Datenreihe.

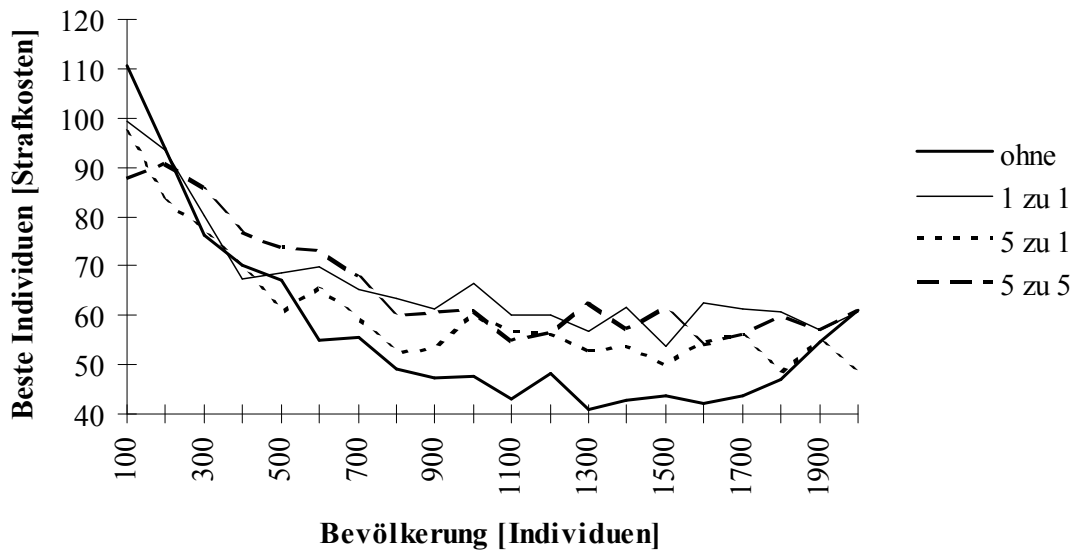

**Abb. 0-13:** Vergleich der durchschnittlichen Güte des besten Individuums der letzten Generation bei verschieden oszillierenden Zielfunktionen und zwei Minuten Gesamtlaufzeit pro Optimierung für eine restriktive Datenreihe.

Wie man den Abbildungen und auch der Zusammenfassung der Ergebnisse in Tabelle 0-7 entnehmen kann, ist für beide Datenreihen die normale Zielfunktion besser als die beiden Oszillierenden. Besonders die Verhältnisse 1 zu 1 und 5 zu 5 schneiden schlecht ab. Einzig das Verhältnis 5 zu 1 erzielt in etwa so gute Ergebnisse wie die herkömmliche Zielfunktion. Dies ist ein Zeichen dafür, daß die häufige Oszillation dem Entwicklungsprozeß der Individuen schadet. Sie brauchen offensichtlich einige Zeit, um sich an ihre Umgebung, d.h. Zielfunktion, „anzupassen". Oszillierende Zielfunktionen werden daher im folgenden nicht weiter betrachtet.

| | Restriktive Datenreihe | | | | Schwankende Datenreihe | | | |
|---|---|---|---|---|---|---|---|---|
| Oszillations-Verhältnis | ohne | 1 zu 1 | 5 zu 1 | 5 zu 5 | ohne | 1 zu 1 | 5 zu 1 | 5 zu 5 |
| Durchschnittliche Güte | 57,0 | 66,5 | 60,6 | 65,9 | 49,2 | 57,4 | 52,9 | 56,7 |
| Beste durchschn. Güte | 41,1 | 53,7 | 48,2 | 54,0 | 33,0 | 45,1 | 39,9 | 44,5 |
| Durchschn. Zulässigkeit | 9% | 1% | 8% | 2% | 18% | 3% | 14% | 3% |

**Tabelle 0-7:** Zusammenfassung der Ergebnisse für oszillierende Zielfunktionen.

# 5.3    Verbesserung der Lösungsgeschwindigkeit

Wie bereits in Kapitel 0 erläutert, ist der Haupteinflußfaktor auf die Laufzeit des genetischen Algorithmus die Bevölkerungsgröße. Wie die bisherigen Untersuchungen zeigten, liegen die Schwankungen der Laufzeit bei Variation der anderen Parameter bzw. Strategien unter 10%. Grundsätzlich führt daher nur eine Verkleinerung der Bevölkerung zu einer kürzeren Laufzeit. Andererseits ist wie bereits mehrfach gesehen, nur eine große Bevölkerung in der Lage, sehr gute Ergebnisse zu erzielen. Die Idee der dynamischen Bevölkerungsgröße ist es daher, die Bevölkerungsgröße an die einzelnen Phasen des Algorithmus anzupassen.

In der Anfangsphase, wenn eine große Vielfalt an Individuen benötigt wird, soll die Bevölkerung groß gehalten werden. Gegen Ende, wenn die meisten Individuen einander bereits sehr ähnlich sind, genügt eine kleinere Bevölkerung. Da es jedoch schwierig ist, die beiden Phasen gegeneinander abzugrenzen, soll die Bevölkerung vereinfachend von einem anfänglichen Höchststand ausgehend kontinuierlich abnehmen. Dazu wird die Bevölkerung in jeder Generation um 1% verkleinert. Damit sie nicht zu klein wird, um zulässige Lösungen zu erzeugen, ist als Untergrenze ein Drittel der Anfangsbevölkerung festgelegt.

In Abb. 0-14 ist für die schwankende bzw. in Abb. 0-15 für die restriktive Datenreihe ein Vergleich der Laufzeit und der Lösungsgüte der besten Individuen der letzten Generation für eine normale, d.h. konstante, und für eine dynamische Bevölkerung aufgetragen. Dabei fällt für beide Datenreihen auf, daß die bei einer dynamischen Bevölkerungsgröße erzielten Ergebnisse nur um etwa 5-10% schlechter als die bei einer konstanten Bevölkerungsgröße sind. Die Laufzeit hat sich jedoch für die dynamische Bevölkerungsgröße in etwa halbiert.

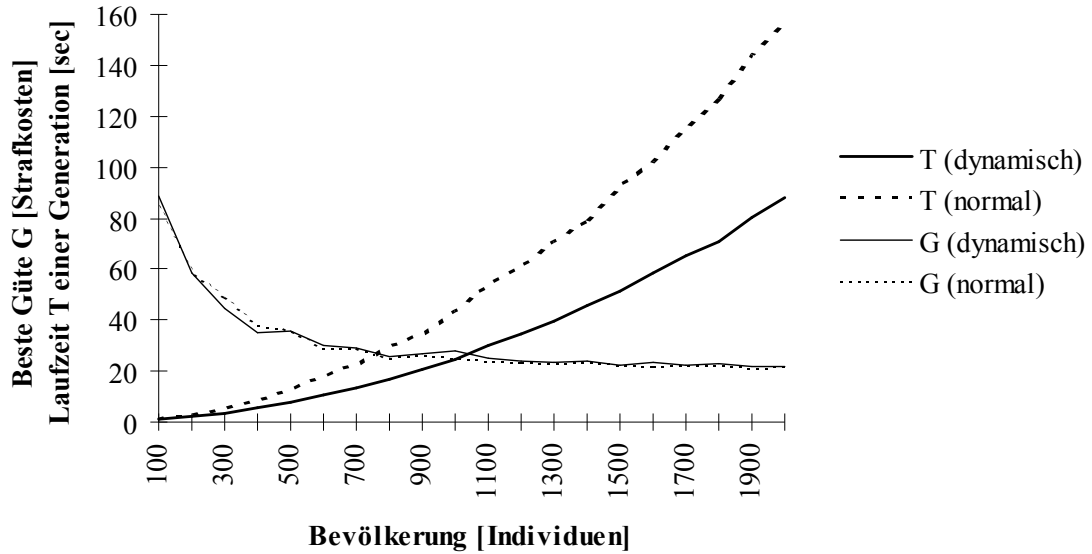

**Abb. 0-14**:    Vergleich der durchschnittlichen Optimierungszeit T und der durchschnittlichen Güte G des besten Individuums der letzten Generation im Falle einer dynamischen und einer normalen Bevölkerungsgröße für eine schwankende Datenreihe.

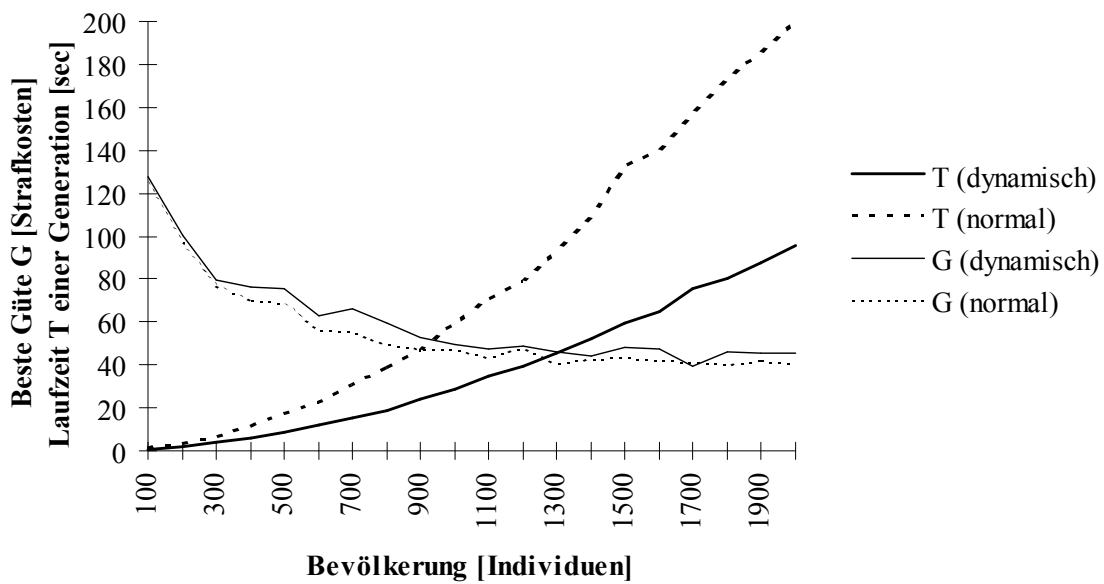

**Abb. 0-15**:    Vergleich der durchschnittlichen Optimierungszeit T und der durchschnittlichen Güte G des besten Individuums der letzten Generation im Falle einer dynamischen und einer normalen Bevölkerungsgröße für eine restriktive Datenreihe.

Auch ein Vergleich der durchschnittlich zulässigen Lösungen in Tabelle 0-8 bestätigt, daß die mit einer dynamischen Bevölkerungsgröße erzielten Ergebnisse nur geringfügig schlechter als die bei konstanter Bevölkerungsgröße sind. Das Verfahren ist daher geeignet, wenn die Lösungsgeschwindigkeit im Vordergrund steht.

|  | Restriktive Datenreihe | | Schwankende Datenreihe | |
|---|---|---|---|---|
|  | dynamisch | normal | dynamisch | normal |
| Durchschnittliche Güte | 60,5 | 56,0 | 31,6 | 31,3 |
| Durchschn. Zulässigkeit | 8% | 10% | 13% | 14% |
| Durchschn. Laufzeit | 37,7 sec | 78,9 sec | 33,5 sec | 59,0 sec |

**Tabelle 0-8:** Zusammenfassung der Ergebnisse für dynamische und normale Bevölkerungsgrößen.

## 5.4   Verbesserung der Zulässigkeit der Lösungen

### 5.4.1   Lernende Gewichte

Alle bisher in Kapitel 0 vorgestellten Verbesserungen haben es nicht geschafft, den Anteil an zulässigen Lösungen entscheidend zu verbessern. Zwar war der Anteil bei großen Bevölkerungen in der Regel wesentlich höher als bei kleineren, erreichte aber auch hier nie mehr als 50%. In diesem Kapitel soll daher versucht werden, den Anteil an zulässigen Lösungen weiter zu verbessern.

Das erste Verfahren hierzu versucht, dies über eine Anpassung des Gewichtes der Bedarfsdeckung zu erreichen. Verletzt das bislang beste Individuum noch viele Restriktionen, so wird das Gewicht der Bedarfsdeckung erhöht und damit die Bedeutung der Wünsche der Krankenschwestern verringert. Erfüllt das beste Individuum bereits alle Restriktionen, so wird das Gewicht der Bedarfsdeckung reduziert, um den Wünschen mehr Bedeutung zu geben und damit zu einer besseren Lösung zu gelangen.

Insgesamt kann sich ein solches „lernendes" Gewicht dazu zwischen einem Drittel und dem Dreifachen des Startgewichtes bewegen. Die gewählten Startgewichte entsprechen den in Kapitel 0 ermittelten von $g_{Bedarf} = 10$ für die schwankende und $g_{Bedarf} = 25$ für die restriktive Datenreihe. Zur Vereinfachung steigt das Gewicht um eine Einheit, falls das beste Individuum eine Restriktion verletzt. Erfüllt es alle Restriktionen, sinkt das Gewicht pro Generation um eine Einheit.

Abb. 0-16 zeigt den Vergleich zwischen lernenden und normalen Gewichten für beide Datenreihen. Es fällt auf, daß die Anwendung des lernenden Gewichtes die durchschnittliche Güte des besten Individuums der letzten Generation für alle Bevölkerungsgrößen verschlechtert.

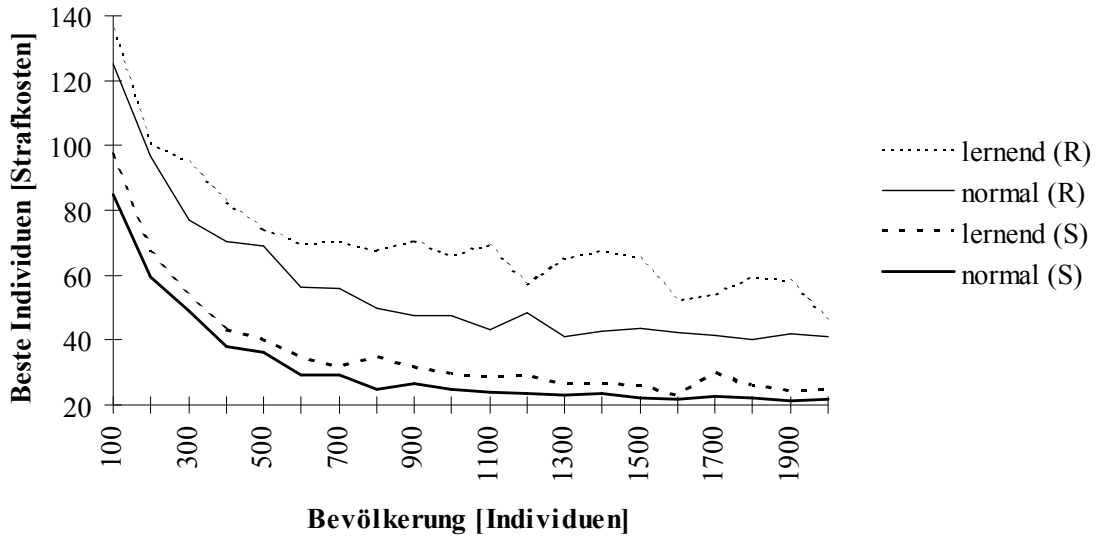

**Abb. 0-16:** Durchschnittliche Güte der besten Individuen der letzten Generation bei lernenden und normalen Gewichten der Bedarfsdeckung für eine restriktive (R) und eine schwankende (S) Datenreihe.

Wie ein Vergleich in Tabelle 0-9 zeigt, erreichen die lernenden Gewichte auch keinen höheren Anteil an zulässigen Lösungen als die normalen Gewichte. Ein Grund für das Versagen des Verfahrens ist nicht bekannt. Eine Möglichkeit hierfür ist der in Kapitel 0 erläuterte nicht monotone Zusammenhang zwischen dem Gewicht der Bedarfsdeckung und dem Anteil an zulässigen Lösungen. Es wäre jedoch auch denkbar, daß die grundlegende Idee, aufgrund der getroffenen Vereinfachungen, für diesen speziellen genetischen Algorithmus nicht anwendbar ist.

|  | Restriktive Datenreihe | | Schwankende Datenreihe | |
|---|---|---|---|---|
|  | lernend | normal | lernend | normal |
| Durchschnittliche Güte | 71,5 | 56,0 | 36,5 | 31,3 |
| Durchschn. Zulässigkeit | 8% | 10% | 14% | 14% |

**Tabelle 0-9:** Zusammenfassung der Ergebnisse für lernende und normale Gewichte der Bedarfsdeckung.

## 5.4.2 Segmentiertes Crossover

Ein zweiter Versuch, den Anteil der zulässigen Lösungen zu erhöhen, setzt beim Crossover an. Das bisher verwendete uniforme Crossover hat zwar den höchsten Freiheitsgrad, d.h. es kann die unterschiedlichsten Kinder erzeugen, es hat aber auch einen schwerwiegenden Nachteil. In Kapitel 0 wurde erläutert, daß der genetische Algorithmus gute Teil-Individuen, sogenannte Building-Blocks, benötigt, um erfolgreich zu sein. Beim uniformen Crossover ist die intakte Weitergabe von längeren Building-Blocks durch das Aufspalten der Eltern-Individuen in alle Einzelmerkmale jedoch sehr unwahrscheinlich.

Daß jedoch gerade längere Building-Blocks zur Lösung des Nurse Scheduling Problems gut geeignet sind, hat der Erfolg des 1-Punkt Crossovers in Kapitel 0 gezeigt. Darauf aufbauend soll der genetische Algorithmus, trotz generellem uniformen Crossover, durch ein sogenanntes segmentiertes Crossover zur intakten Weitergabe von längeren Building-Blocks gezwungen werden. Dazu soll folgende grundsätzliche Überlegung ausgenutzt werden: Ein Stationsschichtplan ist dann gut, wenn alle Restriktionen erfüllt und die Wünsche möglichst vieler Krankenschwestern berücksichtigt werden. Es ist möglich, den kompletten Stationsschichtplan in drei den Qualifikationsstufen entsprechende Teile aufzuspalten, für die die selbe Zielsetzung gilt.

Jeder der drei Teilschichtpläne sollte die Wünsche der entsprechenden Krankenschwestergruppe erfüllen. Im folgenden soll der Teilschichplan für die höchste Qualifikationsstufe mit (I), der für die mittlere mit (II) und der für die niedrigste mit (III) bezeichnet werden. Durch die Möglichkeit, den Bedarf von Krankenschwestern niedrigerer Qualifikationsstufen durch höher qualifizierte abzudecken, ergeben sich folgende drei Kategorien von Restriktionen:

1.  Der Teilschichtplan (I) für die höchstqualifizierten Krankenschwestern muß den Bedarf nach diesen abdecken.

2.  Die beiden Teilschichtpläne für die höchste (I) und für die mittlere (II) Qualifikationsstufe müssen entsprechend zusammen den Bedarf nach Krankenschwestern der mittleren Qualifikationsstufe abdecken.

3.  Letztlich müssen alle drei Teile (I), (II) und (III), d.h. der komplette Stationsschichtplan, den Bedarf nach Krankenschwestern der niedrigsten Qualifikationsstufe erfüllen.

Die Überlegung ist, jedem Individuum nicht eine sondern sieben verschiedene Güten zuzuweisen. Je eine Güte für die jeweils einzelnen Teil-Schichtpläne und für jede Kombinationsmöglichkeit. Zusätzlich werden nur noch 50% der Individuen über ein uniformes Crossover erzeugt, bei dessen Elternauswahl die herkömmliche Güte des gesamten

Stationsschichtplanes entscheidend ist. Die andere Hälfte wird über Individuen erzeugt, die über die neuen sechs weiteren Gütearten ausgewählt werden.

Dabei werden dem Algorithmus fixe Crossover-Punkte aufgezwungen, damit die jeweiligen Teil-Schichtpläne nicht gespaltet werden. Im Einzelnen wird auf diese Art (I) mit (II+III), (I+II) mit (III), (II) mit (I+III) und (I) mit (II) und mit (III) gekreuzt. Welches der Verfahren zum Zuge kommt bleibt dem Zufall überlassen. Wichtig ist, daß die jeweiligen Eltern nicht nach der Güte des Gesamtschichtplanes, sondern nach der Güte des entsprechenden Teil-Schichtplanes ausgewählt werden.

Der Nachteil dieses segmentierten Crossovers ist, daß das Lösungsprogramm wesentlich komplexer und langsamer wird. Für jedes Individuum müssen sieben statt bisher eine Güte berechnet werden. Außerdem müssen jedem Individuum sieben Ränge zugewiesen werden, was ein siebenfaches Sortieren der gesamten Bevölkerung nötig macht. Wie aus Abb. 0-17 und Abb. 0-18 sowie aus Tabelle 0-10 ersichtlich ist, steigt dadurch die durchschnittliche Bearbeitungszeit um etwa 25% im Vergleich zum herkömmlichen genetischen Algorithmus an.

Insgesamt ist das segmentierte Crossover sehr erfolgreich beim Lösen des Nurse Scheduling Problems für beide Datenreihen. Die erzielte Güte ist der des einfachen genetischen Algorithmus ebenbürtig bzw. leicht überlegen. Gleichzeitig beträgt der Anteil an zulässigen Lösungen das Sechsfache des bisher erzielten Wertes. Bei großen Bevölkerungen werden für beide Datenreihen sogar über 90% zulässige Lösungen erzeugt. Das Konzept, den Algorithmus bei der Erstellung der Building-Blocks zu unterstützen, hat sich ausgezahlt, lediglich die längere Optimierungszeit ist ein Nachteil. Dies könnte durch eine Dynamisierung der Bevölkerungsgröße, wie in Kapitel 0 beschrieben, korrigiert werden. Eine weitere Möglichkeit dazu, die zusätzlich noch versucht, die erzeugten Building-Blocks zu verbessern, wird im nächsten Kapitel als Nischenbildung vorgestellt.

### 5.4.3   Nischenbildung

Die Idee der Nischenbildung ist eine Weiterentwicklung des segmentierten Crossovers.[76] Die Bevölkerung wird dazu in sieben Teilbevölkerungen, sogenannte Nischen, aufgespalten. Jede dieser Teilbevölkerungen entwickelt sich grundsätzlich autark von den anderen Teilen. Es entsteht damit eine parallele Optimierung mehrerer Bevölkerungen, die unterschiedliche Ziele entsprechend den im vorherigen Kapitel eingeführten sieben Güteklassen.

Innerhalb der sechs Teilbevölkerungen, die die Optimierung von (I), (II), (III), (I+II), (I+III) und (II+III) anstreben, kommt es dabei zu einem uniformen Crossover. Die siebte Teilbevölkerung wendet zufallsabhängig ein uniformes Crossover mit Individuen aus der eigenen Bevölkerung oder eines der im vorherigen Kapitel beschriebenen segmentierten Crossover mit Individuen aus den entsprechenden Teilbevölkerungen an. Die siebte Teilbevölkerung ist aufgrund ihrer zentralen Bedeutung so groß wie alle anderen Teilbevölkerungen zusammen.

Diese Strategie der Nischenbildung weist zwei Vorteile gegenüber dem segmentierten Crossover auf. Einerseits wird die Güte er zum segmentierten Crossover nötigen Building-Blocks zusätzlich direkt optimiert. Andererseits ist die durchschnittliche Laufzeit einer Optimierung wesentlich geringer, da für jede Teilbevölkerung nur eine Güte und damit für jedes Individuum nur ein Rang berechnet werden muß. Außerdem ist die Auswahl eines Elternteils aus einer kleineren Teilbevölkerung schneller möglich als aus einer großen Bevölkerung. Wie aus Abb. 0-17 und Abb. 0-18 bzw. Tabelle 0-10 ersichtlich, sinkt die durchschnittliche Optimierungszeit auf ca. 25% des ursprünglichen Wertes. Damit ist die Optimierung mittels Nischenbildung fünf mal schneller als die Anwendung des segmentierten Crossovers.

Die Nachteile der Nischenbildung sind, daß sowohl zur Auswahl der einzelnen Building-Blocks wie auch zur eigentlichen Hauptoptimierung nicht mehr die gesamte Bevölkerung, sondern nur noch ein Teil davon zur Verfügung steht. Beides soll teilweise dadurch ausgeglichen werden, daß in jeder Generation ein geringer Prozentsatz der Individuen die Zugehörigkeit zu einer Nische mit einem anderen Individuum tauscht. Trotzdem läßt dies vermuten, daß die Nischenbildung bei gleicher Gesamtbevölkerung dem reinen segmentierten Crossover unterlegen ist. Betrachtet man Abb. 0-17 und Abb. 0-18, so zeigt sich, daß die Unterschiede in der Güte der besten Individuen nur sehr gering sind und durch die eingesparte Laufzeit bei Anwendung der Nischenbildung mehr als ausgeglichen werden.

---

[76] Das hier vorgestellte Modell der Nischenbildung ist eine Mischung aus dem klassischen Insel-Modell und dem Migrations-Modell. Zu diesen und weiteren Parallelisierungsmodellen siehe Schöneburg, 1994, S.242-255.

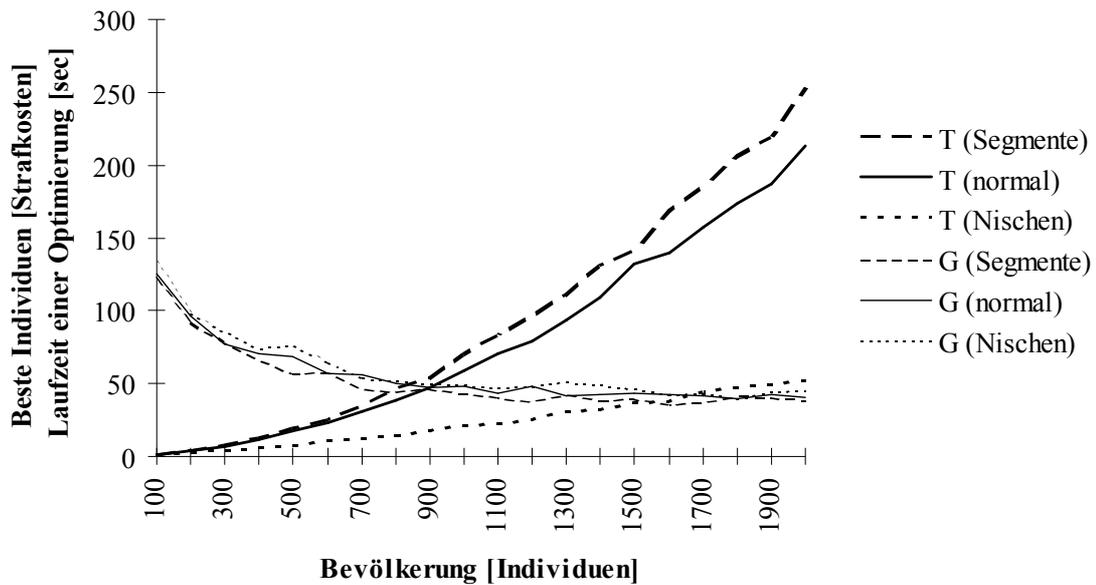

**Abb. 0-17:** Vergleich der durchschnittlichen Güte des besten Individuums der letzten Generation (G) und der durchschnittliche Laufzeit einer Optimierung (T) für eine normale Bevölkerung, eine Bevölkerung mit segmentiertem Crossover und einer Bevölkerung mit Nischenbildung für eine restriktive Datenreihe.

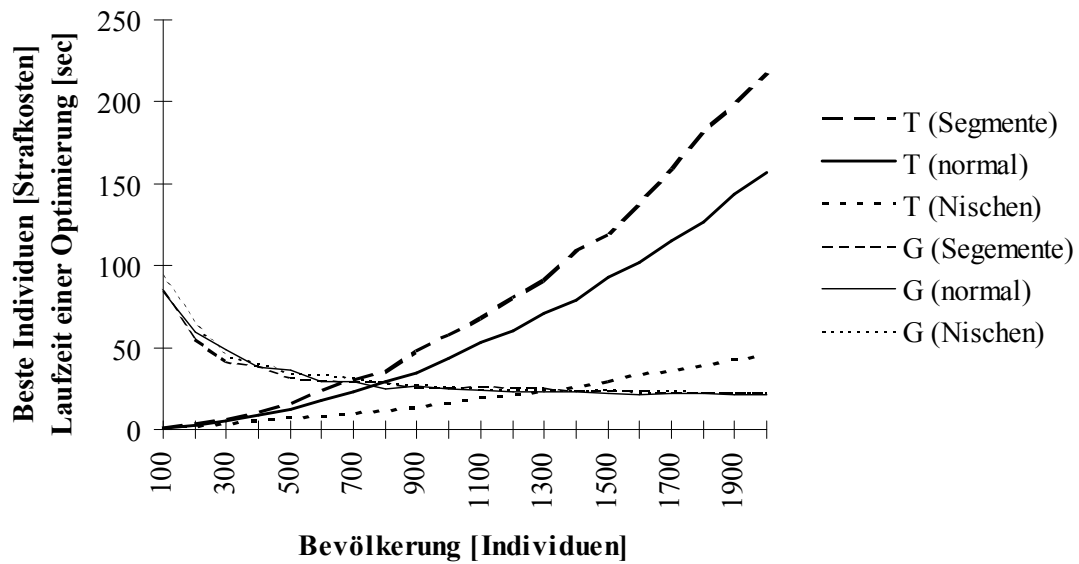

**Abb. 0-18:** Vergleich der durchschnittlichen Güte des besten Individuums der letzten Generation (G) und der durchschnittliche Laufzeit einer Optimierung (T) für eine normale Bevölkerung, eine Bevölkerung mit segmentiertem Crossover und einer Bevölkerung mit Nischenbildung für eine schwankende Datenreihe.

Auch ist der Anteil an zulässigen Lösungen bei Anwendung der Nischenbildung um etwa 10% geringer als bei der Anwendung des segmentierten Crossovers. Dies liegt ebenfalls an obiger Feststellung. Weitere Untersuchungen sind hier nötig, um zu ermitteln, ob eventuell eine andere

Verteilung der Gesamtbevölkerung auf die einzelnen Nischen von Vorteil wäre. Trotz dieses Nachteils kann die Optimierung mittels Nischenbildung durch die schnellere Laufzeit in der selben Zeit etwa die dreifache Menge an zulässigen Lösungen erzeugen.

|  | Restriktive Datenreihe | | | Schwankende Datenreihe | | |
|---|---|---|---|---|---|---|
|  | normal | Segmente | Nischen | normal | Segmente | Nischen |
| Durchschnittliche Güte | 56,0 | 52,2 | 59,8 | 31,3 | 31,8 | 33,3 |
| Durchschn. Zulässigkeit | 10% | 60% | 52% | 14% | 81% | 71% |
| Durchschn. Laufzeit | 79,7 sec | 93,5 sec | 23,7 sec | 59,0 sec | 79,8 sec | 19,5 sec |

**Tabelle 0-10:** Zusammenfassung der Ergebnisse für segmentiertes Crossover und Nischenbildung.

## 5.5   Zusammenfassung der Verbesserungen

In diesem Kapitel sollen nochmals die Ergebnisse des genetischen Algorithmus für alle vier Typen von Datenreihen denen von CPLEX gegenübergestellt werden. Die verwendeten Parameter bzw. Strategien sind die in den vorherigen Kapiteln als optimal gefundenen, d.h. die in Kapitel 0 gewählten zuzüglich einer 20%-Eltern-Substitution anstelle einer 10%-Strategie. Außerdem wird der herkömmlich genetische Algorithmus um eine maximale lokale Optimierung erweitert (vgl. Kapitel 0). Parallel dazu werden alle Datenreihen durch einen genetischen Algorithmus, der zusätzlich mit einem segmentierten Crossover bzw. mit Nischenbildung arbeitet, gelöst.

- Restriktive Datenreihen (Abb. 0-19): Dies ist der einzige Datentyp, für den CPLEX noch immer leichte Vorteile besitzt. Zwar gelingt es dem genetischen Algorithmus jetzt, mittels Nischenbildung bzw. segmentiertem Crossover bei großen Bevölkerungen bis zu 80% zulässige Lösungen zu erzielen, jedoch ist deren Lösungsgüte im Durchschnitt ca. 30% schlechter als die von CPLEX ermittelte. Im Vergleich zum herkömmlichen genetischen Algorithmus in Kapitel 0 beträgt der Anteil der zulässigen Lösungen durch das segmentierte Crossover ein Vielfaches, im Vergleich zum erweiterten genetischen Algorithmus wird er immerhin noch verdoppelt.

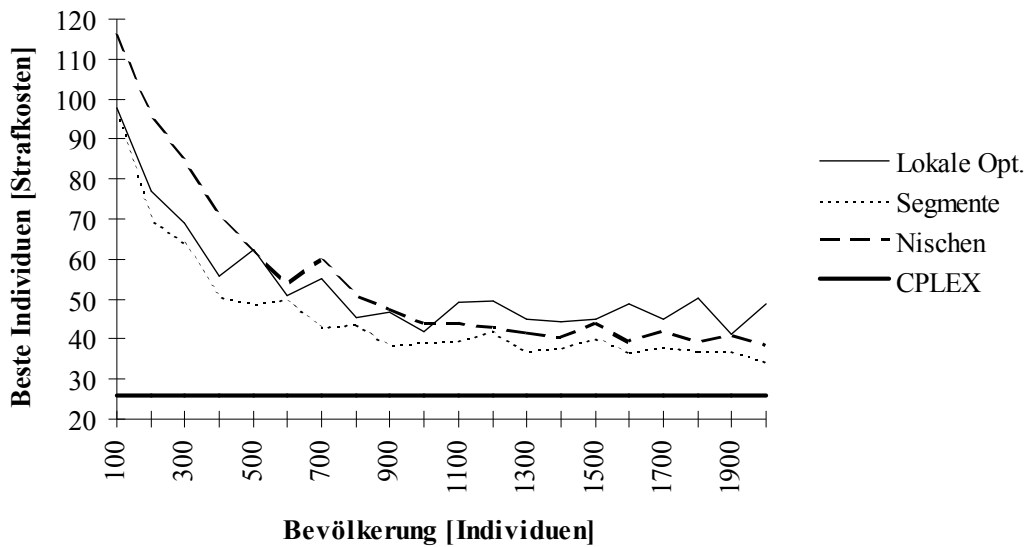

**Abb. 0-19:** Vergleich der durchschnittlichen Güte des besten Individuums der letzten Generation bei einem genetischen Algorithmus mit lokaler Optimierung, bei segmentiertem Crossover, bei Nischenbildung sowie bei Anwendung von CPLEX für eine restriktive Datenreihe.

- Wenig restriktive Datenreihen (Abb. 0-20): Wie bereits in Kapitel 0 gesehen, war zur Lösung dieser Datenreihe der nicht erweiterte genetische Algorithmus ausreichend. Die Erweiterung um die lokale Optimierung erbringt eine um ca. 10% verbesserte Lösungsgüte. Segmentiertes Crossover bzw. Nischenbildung ergeben nochmals ca. 20% mehr zulässige Lösungen. Für eine Bevölkerung von 2000 Individuen liefert der genetische Algorithmus mit Nischenbildung in weniger als einer Minute Ergebnisse, die bis zu 10% besser sind als die von CPLEX innerhalb von zehn Minuten berechneten.

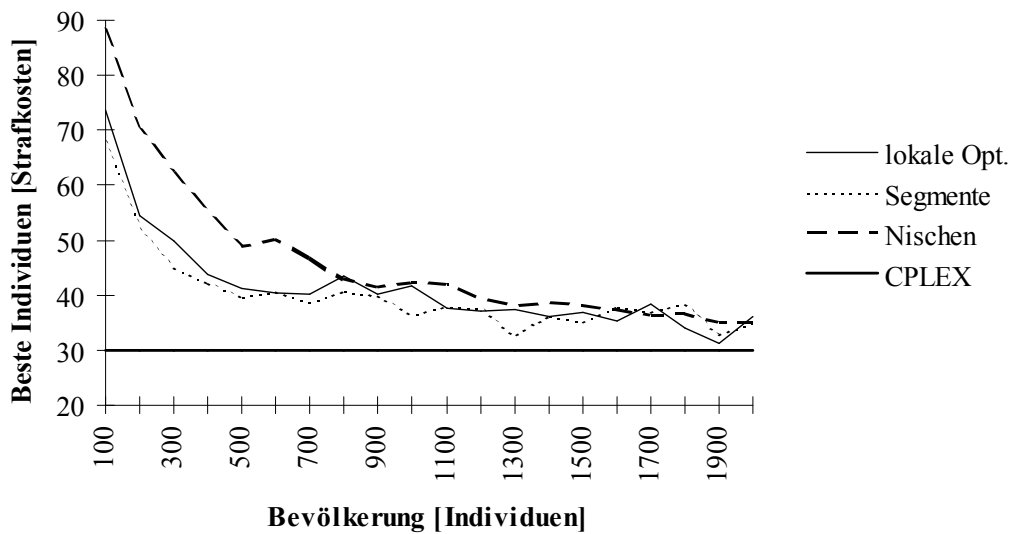

**Abb. 0-20:**    Vergleich der durchschnittlichen Güte des besten Individuums der letzten Generation bei einem genetischen Algorithmus mit lokaler Optimierung, bei segmentiertem Crossover, bei Nischenbildung sowie bei Anwendung von CPLEX für eine wenig restriktive Datenreihe.

- Durchschnittliche Datenreihen (Abb. 0-21): Die durchschnittliche Datenreihe konnte mittels des nicht erweiterten genetischen Algorithmus in Kapitel 0 zufriedenstellend gelöst werden. Der erweiterte Algorithmus findet bei einer Bevölkerung von 1500 Individuen zu 90% eine zulässige Lösung, die im Durchschnitt weniger als 2% von der optimalen Lösung abweicht. Durch Nischenbildung kann der Anteil an zulässigen Lösungen praktisch auf 100% erhöht werden. Die durchschnittlich benötigte Lösungszeit ist dabei nur ca. halb so groß wie die von CPLEX.

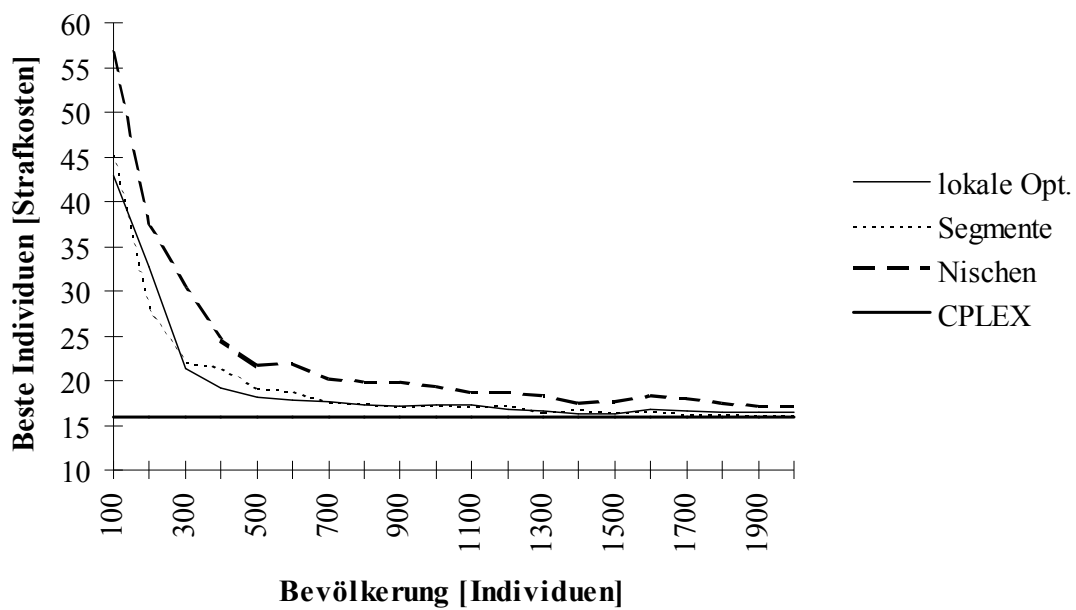

**Abb. 0-21:**   Vergleich der durchschnittlichen Güte des besten Individuums der letzten Generation bei einem genetischen Algorithmus mit lokaler Optimierung, bei segmentiertem Crossover, bei Nischenbildung sowie bei Anwendung von CPLEX für eine durchschnittliche Datenreihe.

- Schwankende Datenreihen (Abb. 0-22): Bereits bei den Ergebnissen des Kapitels 0 zeigte sich, daß der herkömmliche genetische Algorithmus in der Lage ist, wesentlich bessere Ergebnisse als CPLEX zu erzielen. So erreicht der genetische Algorithmus mit lokaler Optimierung bei einer Bevölkerung von 2000 Individuen im Durchschnitt 20% bessere Lösungen als CPLEX. Wird zusätzlich noch Nischenbildung bzw. segmentiertes Crossover verwendet, so steigt der Anteil von zulässigen Lösungen von 40% auf 100%. Die Optimierungszeit beträgt dabei bei Nischenbildung durchschnittlich nur 40 Sekunden.

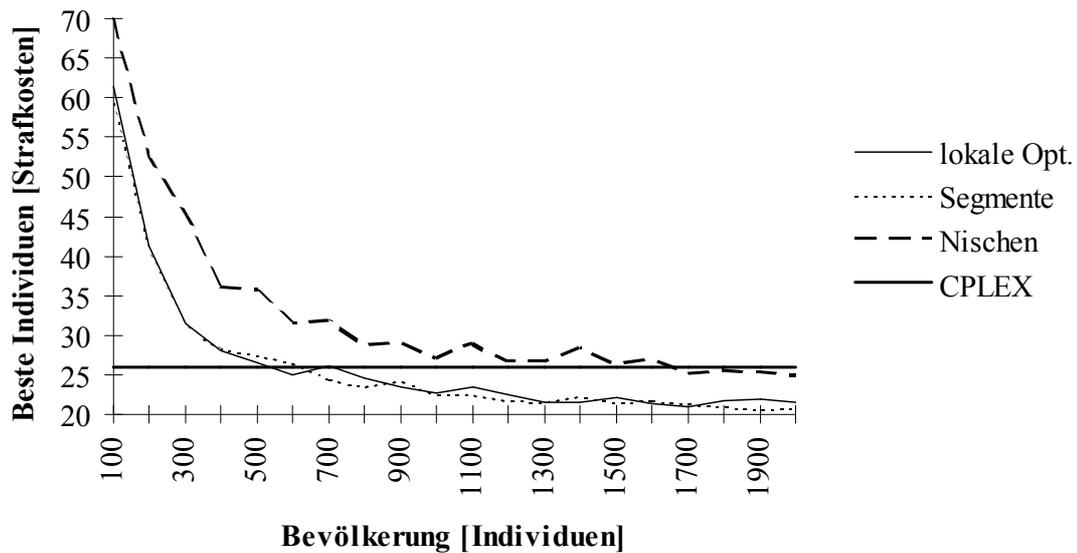

**Abb. 0-22:**   Vergleich der durchschnittlichen Güte des besten Individuums der letzten Generation bei einem genetischen Algorithmus mit lokaler Optimierung, bei segmentiertem Crossover, bei Nischenbildung sowie bei Anwendung von CPLEX für eine schwankende Datenreihe.

Zusammenfassend läßt sich sagen, daß der um die lokale Optimierung und Nischenbildung erweiterte genetische Algorithmus alle Datenreihen zuverlässig lösen kann. Es fällt auf, daß die jeweils besten Werte bei sehr hohen Bevölkerungen am Rande des betrachteten Spektrums liegen. Eine weitere Erhöhung der Bevölkerung über 2000 Individuen hinaus ist zu prüfen. Die Optimierungszeit ist bei Anwendung der Nischenbildung unkritisch. Sie liegt im Schnitt selbst bei einer Bevölkerung von 2000 Individuen unter einer Minute.

| Datentyp | untersuchte Größe | lokale Opt. | Segmente | Nischen | CPLEX |
|----------|-------------------|-------------|----------|---------|-------|
| restriktiv | Durchschn. beste Güte | 53,5 | 46,2 | 54,8 | 26 |
| | Durchschn. Optimierungszeit | 57,9 sec | 127,9 sec | 24,1 sec | 30 sec |
| | Durchschn. zulässige Lösungen | 29% | 56% | 55% | 100% |
| | Beste Bevölkerung | 1900 | 2000 | 2000 | --- |
| | Deren durchschn. beste Güte | 41,4 | 34,1 | 38,4 | --- |
| | Deren durchschn. Optimierungszeit | 133,2 sec | 349,7 sec | 52,1 sec | --- |
| | Deren Anteil zulässiger Lösungen | 35% | 80% | 80% | --- |
| | Insgesamt beste erzielte Lösung | 27 | 26 | 26 | 26 |
| unrestriktiv | Durchschn. beste Güte | 41,5 | 40,2 | 46,3 | 30 |
| | Durchschn. Optimierungszeit | 63,9 sec | 142,7 sec | 23,4 sec | > 10 min |
| | Durchschn. zulässige Lösungen | 79% | 98% | 95% | 100% |
| | Beste Bevölkerung | 1900 | 1900 | 2000 | --- |
| | Deren durchschn. beste Güte | 31,2 | 32,75 | 35,1 | --- |
| | Deren durchschn. Optimierungszeit | 166,8 sec | 372,6 sec | 48,7 sec | --- |
| | Deren Anteil zulässiger Lösungen | 90% | 100% | 100% | --- |
| | Insgesamt beste erzielte Lösung | 28 | 27 | 28 | 30 |
| normal | Durchschn. beste Güte | 19,4 | 19,5 | 22,5 | 16 |
| | Durchschn. Optimierungszeit | 36,5 sec | 78,6 sec | 16,8 sec | 60 sec |
| | Durchschn. zulässige Lösungen | 62% | 84% | 83% | 100% |
| | Beste Bevölkerung | 1500 | 1900 | 2000 | --- |
| | Deren durchschn. beste Güte | 16,3 | 16,0 | 17,0 | --- |
| | Deren durchschn. Optimierungszeit | 57,6 sec | 192,0 sec | 36,3 sec | --- |
| | Deren Anteil zulässiger Lösungen | 90% | 100% | 100% | --- |
| | Insgesamt beste erzielte Lösung | 16 | 16 | 16 | 16 |
| schwankend | Durchschn. beste Güte | 26,4 | 26,1 | 32,7 | 26 |
| | Durchschn. Optimierungszeit | 41,5 sec | 93,2 sec | 20,7 sec | > 10 min |
| | Durchschn. zulässige Lösungen | 25% | 78% | 72% | 100% |
| | Beste Bevölkerung | 2000 | 1900 | 1900 | --- |
| | Deren durchschn. beste Güte | 21,5 | 20,6 | 25,3 | --- |
| | Deren durchschn. Optimierungszeit | 103,5 sec | 238,5 sec | 43,6 sec | --- |
| | Deren Anteil zulässiger Lösungen | 40% | 100% | 100% | --- |
| | Insgesamt beste erzielte Lösung | 20 | 20 | 21 | 26 |

**Tabelle 0-11:** Zusammenfassender Vergleich zwischen den Ergebnissen des Genetischen Algorithmus und CPLEX für alle Datenreihentypen.

## 5.6    Erweiterungen des Grundproblemes

### 5.6.1    Automatische Berücksichtigung der Schichtmuster der Vorwoche

Bei allen in Kapitel 0 aufgeführten Erweiterungen handelt es sich um nachträgliche Wünsche der Krankenhausleitung, die nach Vorstellung der ersten Ergebnisse geäußert wurden. Sie sind noch nicht in das Programm implementiert. Im Rahmen diese Kapitels soll aufgezeigt werden, inwiefern dies in Zukunft geschehen könnte.

Die erste von der Krankenhausleitung gewünschte Erweiterung betrifft die Berücksichtigung des Schichtmusters, das eine Krankenschwester in der Woche vor der eigentlichen Planungswoche gearbeitet hat. Dessen Güte soll bei der Berechnung der Güte der Schichtmuster der Planungswoche nun automatisch berücksichtigt werden. Bisher geschah dies manuell durch die Stationsleitung. Außerdem soll berücksichtigt werden, daß keine Krankenschwester mehr als sieben Tage an einem Stück arbeiten darf.

Der erste Aspekt kann durch eine kumulative Berechnung der Krankenschwester-Schichtmuster-Strafkosten geschehen. Zu den Strafkosten dieser Woche werden die des in der Vorwoche gearbeiteten Musters addiert. Dies geschieht nicht, falls die Strafkosten der Vorwoche unterhalb eines festzulegenden Niveaus liegen. Damit soll berücksichtigen werden, daß eine Krankenschwester nicht erneut bevorzugt behandelt wird, falls sie in der letzten Woche ein für sie sehr gutes Schichtmuster gearbeitet hat. Hatte eine Krankenschwester andererseits ein für sie schlechtes Schichtmuster, so werden alle ihre Schichtmuster-Strafkosten entsprechend erhöht, was den Algorithmus durch die Minimierung der Summe aller Krankenschwester-Schichtmuster-Strafkosten dazu veranlassen wird, ihr ein relativ gutes Schichtmuster zu geben.

Eventuell ist auch eine andere als eine lineare Verknüpfung der Strafkosten der vergangenen Wochen, wie z.B. eine exponentielle Glättung erster Ordnung, denkbar. Ein weiterer Gedanke wäre es, mehr als nur die letzte gearbeitete Woche zu berücksichtigen, um die Verteilung ungünstiger Schichtmuster noch gerechter zu gestalten.

Um das Problem der maximal aufeinander folgenden Arbeitstage zu lösen, wird die Schichtmustertabelle (vgl. Anhang A4) um zwei Spalten erweitert. Für jedes Schichtmuster wird der erste und der letzte freie Tag berechnet. Die Zeitspanne zwischen dem letztem freien Tag des in der Vorwoche gearbeiteten und dem erstem freien Tag des in dieser Woche gearbeiteten Schichtmusters darf nun maximal sieben Tage betragen.
Das mathematische Modell in Kapitel 0 ist wie folgt zu ergänzen:

zusätzliche Modellparameter:

$e_j$        = erster freier Tag des Schichtmusters j

$l_i$        = letzter freier Tag des in der Vorwoche von Krankenschwester i gearbeiteten

         Schichtmusters

zusätzliche Restriktionen:

4. Keine Krankenschwester darf mehr als sieben Tage an einem Stück arbeiten:

$$\sum_{j=1}^{m} x_{ij}\left(7 + e_j - l_i\right) \leq 7 \quad \forall i \quad \Leftrightarrow \quad \sum_{j=1}^{m} x_{ij}\left(e_j - l_i\right) \leq 0 \quad \forall i \qquad\qquad ( \; 0\text{-}1 \; )$$

Der genetische Algorithmus kann diese Restriktion implizit in der Kodierung berücksichtigen, indem das Alphabet eines Elementes weiter eingeschränkt wird.[77] Dies muß je nach Wochenarbeitszeit bzw. dem in der vorherigen Woche gearbeiteten Schichtmuster erfolgen. Wird das Alphabet derart beschränkt, so ist die zusätzliche Restriktion jederzeit und auch nach Anwendung aller genetischen Operatoren erfüllt.

## 5.6.2   Höchstgrenze für die Wochendarbeit der Oberschwestern

Aufgrund der hohen Wochenendzuschläge möchte die Krankenhausleitung aus Kostengründen erreichen, daß am Wochenende nur maximal eine der beiden auf einer Station beschäftigten Oberschwestern gleichzeitig arbeitet. Die Oberschwestern bilden eine Untergruppe der Krankenschwestern der höchsten Qualifikationsstufe und sind ansonsten beliebig durch diese austauschbar. Dies führt zu folgenden Erweiterungen des mathematischen Modells aus Kapitel 0:

zusätzliche Modellparameter:

---

[77] Vgl. hierzu die Ausführungen bezüglich des Kodierungsproblems in Kapitel 0 und 0.

$$o_i = \begin{cases} 1 & \text{Krankenschwester } i \text{ ist eine Oberschwester} \\ 0 & \text{sonst} \end{cases}$$

<u>zusätzliche Restriktionen:</u>

5. Es darf höchstens jeweils eine Oberschwester an Samstag und Sonntag arbeiten:

$$\sum_{i=1}^{n} \sum_{j=1}^{m} o_i x_{ij} a_{jk} \leq 1 \quad k = 1, 8 \; ; \; k = 7, 14 \tag{0-2}$$

Da diese Nebenbedingung nicht durch die gewählte Kodierung berücksichtigt werden kann, muß sie mittels eines Strafkostenansatzes in die Zielfunktion aufgenommen werden.[78] Das dafür nötige Gewicht soll $g_4$ genannt werden. Der Term, um den die Zielfunktion ( 3-1 ) zu erweitern ist, entspricht dann:

$$g_4 \left( \max \left[ \sum_{k=1,8} \sum_{i=1}^{n} \sum_{j=1}^{m} o_i a_{jk} x_{ij} - 1 \; ; \; 0 \right] + \max \left[ \sum_{k=7,14} \sum_{i=1}^{n} \sum_{j=1}^{m} o_i a_{jk} x_{ij} - 1 \; ; \; 0 \right] \right) \tag{0-3}$$

### 5.6.3  Einführung von Betreuungsgruppen

Weiterhin möchte die Krankenhausleitung, daß jeder Patient jederzeit einen Ansprechpartner hat, den er persönlich kennt. Dazu ist jedem Patienten eine Betreuungsgruppe von Krankenschwestern zugeordnet, von denen rund um die Uhr mindestens eine anwesend sein sollte. Um dies zu ermöglichen, müssen aus jeder Betreuungsgruppe nachts eine und tagsüber zwei Krankenschwestern, eine für die Früh- und die andere für die Spätschicht, anwesend sein. Diese Anforderungen führen zu folgenden Erweiterungen des mathematischen Modells aus Kapitel 0:

<u>zusätzliche Indizes:</u>

---

[78]  Vgl. zum Strafkostenansatz die Ausführungen in Kapitel 0 und 0.

h         = 1,..., $h_{max}$        Patientenindex

<u>zusätzliche Modelparameter:</u>

$h_{max}$      = insgesamt auf einer Station untergebrachte Patienten

$b_{ih}$    $= \begin{cases} 1 & \text{Krankenschwester i gehört zur Betreuungsgruppe von Patient h} \\ 0 & \text{sonst} \end{cases}$

<u>zusätzliche Restriktionen:</u>

6. Es muß jederzeit mindestens eine Krankenschwester aus jeder Betreuungsgruppe anwesend sein:

$$\sum_{i=1}^{n}\sum_{j=1}^{m} a_{jk} b_{ih} x_{ij} \geq 2 \qquad \forall h; \quad k = 1,...,7 \tag{0-4}$$

$$\sum_{i=1}^{n}\sum_{j=1}^{m} a_{jk} b_{ih} x_{ij} \geq 1 \qquad \forall h; \quad k = 8,...,14 \tag{0-5}$$

Bislang wurde noch kein geeigneter Ansatz gefunden, diese zusätzlichen Restriktionen innerhalb des genetischen Algorithmus zu berücksichtigen. Da eine implizite Berücksichtigung durch die gewählte Kodierung ausscheidet, bleibt nur ein Strafkosten-Ansatz. Doch auch dieser ist bei der gewählten Repräsentation des Problems nicht ohne weiteres durchführbar, da die Güte des Schichtmusters einer Krankenschwester von den jeweils gearbeiteten Schichtmustern der anderen Krankenschwestern der selben Betreuungsgruppe abhängig ist. Dies bedeutet, daß die Krankenschwester-Schichtmuster-Strafkosten $p_{ij}$ zu einer Funktion der jeweils von anderen Krankenschwestern in der Betreuungsgruppe gearbeiteten Schichtmuster werden, was zu einer sehr komplexen Zielfunktion führt.

# 6 Zusammenfassung und Ausblick

Kapitel 0 führt in die Problematik des Nurse Scheduling ein. Anhand von Datenmaterial eines britischen Krankenhauses soll ein einwöchiger Stationsschichtplan mittels eines Computerprogrammes erstellt werden. Bisher geschah dies manuell. Für die Krankenhausleitung ist außer der Erfüllung der Nachfrage nach Krankenschwestern wichtig, daß möglichst viele der Arbeitswünsche der Krankenschwestern berücksichtigt werden. In Kapitel 0 wird dazu ein entsprechendes mathematisches Modell formuliert. Da die Krankenhausleitung eine sehr flexible Lösungsmethode wünscht, kommen zur Lösung nur Heuristiken in Frage. Aufgrund des Erfolges von Genetischen Algorithmen für das ähnliche Stundenplanproblem sollen diese auch hier zur Anwendung kommen.

In Kapitel 0 werden genetische Algorithmen vorgestellt und deren Unterschiede zu klassischen Optimierungstechniken behandelt. Genetische Algorithmen sind stochastische Suchalgorithmen, die von einer Bevölkerung von Lösungen ausgehend die Optimierung betreiben. Um zu neuen Lösungen zu gelangen werden alte Lösungen mutiert (Mutation) oder miteinander gekreuzt (Crossover). Nach Darwins Grundsatz des „survival of the fittest" erzeugen bessere Lösungen mehr Kinder als schlechtere. Um Skalierungsproblemen aus dem Weg zu gehen werden Lösungen jedoch nicht nach ihrem Zielfunktionswert, sondern nach deren Rang innerhalb der Bevölkerung ausgewählt. Sind genügend Kinder erzeugt, ersetzt ein Teil von ihnen je nach gewählter Substitutions-Strategie die Eltern. Die theoretische Grundlage der genetischen Algorithmen bilden das Schemata-Theorem und die Building-Block-Hypothese.

Das eigentliche Programm zur Lösung des Nurse Scheduling Problems wird in Kapitel 0 vorgestellt, es ist in Turbo Pascal implementiert. Die Schwierigkeit besteht darin, eine für die Anwendung des genetischen Algorithmus geeignete Kodierung der Lösungen zu finden. Dies ist so gelöst, daß zwei der drei Nebenbedingungen des mathematischen Modells implizit durch die Kodierung erfüllt sind. Die dritte Nebenbedingung wird über einen Strafkostenansatz in die Zielfunktion eingebracht. Für alle wichtigen Programmteile enthält das Kapitel einen beispielhaften Pseudocode.

In Kapitel 0 werden erste Analysen durchgeführt, um zu guten Werten für die zahlreichen Parameter des genetischen Algorithmus zu gelangen. Im einzelnen werden akzeptable Werte für die Mutationsrate, die Reproduktionsrate und das für den Strafkostenansatz nötige Gewicht der Bedarfsdeckung ermittelt. Außerdem wird der Einfluß verschiedener Crossover- und Substitutions-Strategien auf die Lösungsqualität untersucht. Anschließend werden die Ergebnisse des genetischen Algorithmus mit denen von CPLEX, einer kommerziellen Optimierungssoftware, verglichen. Für die Hälfte aller Datenreihen schneiden beide Verfahren ungefähr gleich gut ab. Für Datenreihen, die durch Personalengpässe nur wenige zulässige Schichtpläne ermöglichen, ermittelt CPLEX wesentlich bessere Lösungen. Für Datenreihen, die durch schwankenden Bedarf komplexere Nebenbedingungen aufweisen, ist der genetische Algorithmus erfolgreicher. Als ein grundsätzliches Problem der bisherigen Vorgehensweise erweist sich jedoch der geringe Anteil an zulässigen Lösungen.

Durch Erweiterungen des klassischen Genetischen Algorithmus in Kapitel 0 sollen die bisher schlecht lösbaren Datenreihen besser optimiert werden. Eine Variation der herkömmlichen Strategien erweist sich dabei als nicht sehr erfolgreich. Eine zusätzliche lokale Optimierung kann die Lösungen geringfügig verbessern, während eine intelligente Mutation, zwei oszillierende Zielfunktionen und lernende Gewichte keinerlei Verbesserungen erbringen. Eine Dynamisierung der Bevölkerungsgröße ermöglicht bei ca. 10% Einbuße an Lösungsqualität die Optimierungszeit zu halbieren.

Eine Rückbesinnung auf die Building-Block-Hypothese und ein daraus abgeleitetes spezielles Crossover ermöglicht schließlich den Durchbruch. Für alle Datenreihen werden bei geeigneter Bevölkerungsgröße zu mehr als 90% zulässige Lösungen erzeugt. Eine zusätzliche interne Parallelisierung durch Nischenbildung beschleunigt den Algorithmus auf das Sechsfache. Ein abschließender Vergleich mit CPLEX für verschiedene Datenreihen zeigt, daß der genetische Algorithmus außer für die schwankenden Datenreihen ebenbürtig ist bzw. sogar leichte Vorteile besitzt. Für letztere erzielt er ca. 20% bessere Lösungen in weniger als der halben Zeit.

Als nächster Schritt ist nun eine Verifizierung der Analysen anhand weiterer Datenreihen notwendig. Danach sollten die in Kapitel 0 angesprochenen Erweiterungen in das Programm aufgenommen werden. Anschließend sollten die gewonnen Lösungen der Krankenhausleitung vorgestellt werden, um sicherzustellen, daß deren praktische Umsetzung keine weiteren Probleme verursacht. Auf der Seite des Algorithmus wäre zusätzlich zu den bereits in Kapitel 0 und 0 erwähnten Verbesserungsvorschlägen eine Weiterentwicklung des erfolgreichen segmentierten Crossovers denkbar. Dies könnte derart erfolgen, daß noch kleinere Building-Blocks als die durch die Qualifikationsstufen vorgegebenen optimiert und zusammengefügt werden. Eventuell wäre z.B. ein frühzeitiges Aufspalten in Krankenschwestern, die Tagschichten bzw. Nachtschichten arbeiten, möglich. Insgesamt zeigte sich der modifizierte

Genetische Algorithmus in der Lage, das Nurse Scheduling Problem schnell und gut zu lösen, was für die Flexibilität und Zukunft dieses Optimierungsverfahrens spricht.

# Literaturverzeichnis

[1]  Bäck, T. (1993): *Applications of Evolutionary Algorithms*, 5. Edition, Dortmund, 1993.

[2]  Bäck, T. (1996): *Evolutionary Algorithms in Theory and Practice*, New York, 1996.

[3]  Banzhaf, W. (1993): *Genetic Programming for Pedestrians*, International Conference on Genetic Algorithms, Urbana, 1993.

[4]  Brusco, M. (1993): *Constrained Nurse Staffing Analysis*, OMEGA, Vol. 21, S. 175-186.

[5]  Chambers, L. (1995): *Practical Handbook of Genetic Algorithms Volume I*, Boca Raton, 1995.

[6]  Domschke, W. (1991): *Einführung in Operations Research*, Heidelberg, 1991.

[7]  Dowsland, K. (1996a): *Genetic Algorithms - a Tool for OR?*, Journal of the Operations Research Society 47, S. 550-561.

[8]  Dowsland, K. (1996b): *Nurse Scheduling with Tabu Search and Strategic Oscillation*, S. 1-21, European Business Management School, Swansea, 1996.

[9]  Easton, F. (1992): *Analysis of Alternative Scheduling Policies for Hospital Nurses*, Production and Operations Management, Vol. 1 No. 2, S. 159-174.

[10]  Eiben, A. (1994): *Repairing, adding constraints and learning as a means of improving GA [Genetic Algorithms] performance on CSPs [Constrained Satisfaction Problems]*, 6th Belgian-Dutch Conference on Machine Learning, 1994.

[11]  Eiben, A. (1995a): *Heuristic Genetic Algorithms for Constrained Problems*, Artificial Intelligence Group Amsterdam, 1995.

[12]  Eiben, A. (1995b): *Solving Constraint Satisfaction Problems Using Genetic Algorithms*, Artificial Intelligence Group Amsterdam, 1995.

[13]  Fogarty, T. (1994): *Evolutionary Computing*, Berlin, 1994.

[14]  Geyer-Schulz, A. (1995): *Fuzzy Rule-Based Expert Systems and Genetic Machine Learning*, Heidelberg, 1995.

[15]  Goldberg, D. (1989): *Genetic Algorithms in Search, Optimisation, and Machine Learning*, Reading, 1989.

[16]  Grütz, M. (1982): *Personalplanung für den Pflegebereich im Allgemeinkrankenhaus mit Methoden des Operations Research*, WiSt 1982 Heft 2, S. 88-94.

[17]  Heistermann, J. (1994): *Genetische Algorithmen*, Stuttgart, 1994.

[18]  Hennefeld, J. (1995): *Using Turbo Pascal 6.0 - 7.0*, Third Edition, Boston, 1995.

[19]  Kennedy, S. (1993): *Five Ways to a Smarter Genetic Algorithm*, AI Expert 1993, S.35-38.

[20]  Koza, J. (1993): *Genetic Programming: on the programming computers by means of natural selection*, Cambridge, 1993.

[21] Kurbel, K. (1995): *Ein Vergleich von Verfahren zur Maschinenbelegungsplanung: Simulated Annealing, Genetische Algorithmen und mathematische Optimierung*, Wirtschaftsinformatik 37, S. 581-593.

[22] Männer, R. (1994): *Parallel Problem Solving from Nature - PPSN III*, Heidelberg, 1994.

[23] Michalewicz, Z. (1992): *Genetic Algorithms + Data Structures = Evolution Programs*, Berlin, 1992.

[24] Miller, E. (1976): *Nurse Scheduling Using Mathematical Programming*, Operations Research, Vol. 24, S. 857-870.

[25] Nooriafshar, M. (1995): *A heuristic approach to improving the design of nurse training schedules*, European Journal of Operational Research 81, S. 50-61.

[26] Randhawa, S. (1993): *A Heuristic Based Computerised Nurse Scheduling System*, Computers Operational Research, Vol. 20, S. 837-844.

[27] Rosenbloom, E. (1987): *Cyclic nurse Scheduling*, European Journal of Operational Research 31, S. 19-23.

[28] Ross, P. (1994): *Applications of genetic algorithms*, Edinburgh, 1994.

[29] Schneeweiß, C. (1991): *Planung 1*, Heidelberg, 1991.

[30] Schneeweiß, C. (1992): *Einführung in die Produktionswirtschaft*, 4. Auflage, Heidelberg, 1992.

[31] Schneeweiß, C. (1996): *Festlegung einer Jahresarbeitszeitverteilung durch Optimierung von Schichtplänen*, OR Spektrum 18, S. 15-27.

[32] Schöneburg, E. (1994): *Eine Einführung in Theorie und Praxis der simulierten Evolution*, Bonn, 1994.

[33] Schwefel, H.-P (1991): *Parallel Problem Solving from Nature*, Heidelberg, 1991.

[34] Sitford, S. (1992): *Workforce staffing and scheduling : Hospital nursing specific models*, European Journal of Operational Research 60, S. 233-246.

[35] Soucek, B. (1992): *Dynamic, Genetic, and Chaotic Programming: The Sixth Generation*, New York, 1992.

[36] Stender, J. (1995): *Genetic Algorithms in Optimisation, Simulation and Modelling*, Amsterdam, 1995.

[37] Worthington, D. (1988): *Allocating nursing staff to hospital wards - A case study*, European Journal of Operational Research 33, S. 174-182.

[38] Yao, X. (1995): *Progress in Evolutionary Computation*, Berlin, 1995.

# Anhang A Datenfiles

## Anhang A.1 Nachfrage nach Krankenschwestern

Es folgt eine Tabelle mit der beispielhaften Nachfrage nach Krankenschwestern auf einer Station für eine Woche.

- So - Sa seien die sieben Tage bzw. Nächte einer Woche.
- Q1, Q2 und Q3 seien die benötigte Anzahl von Krankenschwestern der entsprechenden Qualifikationsstufe, wobei Krankenschwestern einer niederen Stufe durch die einer höheren ersetzt werden können.

| | Tag-Schichten | | | | | | | Nacht-Schichten | | | | | | |
|---|---|---|---|---|---|---|---|---|---|---|---|---|---|---|
| | So | Mo | Di | Mi | Do | Fr | Sa | So | Mo | Di | Mi | Do | Fr | Sa |
| Q1 | 2 | 2 | 2 | 2 | 2 | 2 | 2 | 1 | 1 | 1 | 1 | 1 | 1 | 1 |
| Q2 | 2 | 2 | 2 | 2 | 2 | 2 | 2 | 1 | 1 | 1 | 1 | 1 | 1 | 1 |
| Q3 | 5 | 5 | 5 | 5 | 5 | 5 | 5 | 1 | 1 | 1 | 1 | 1 | 1 | 1 |

**Tabelle 0-1:** Wochenbedarf an Krankenschwestern für eine Station.

## Anhang A.2 Qualifikationen der Krankenschwestern

Es folgt eine Tabelle mit den beispielhaften Qualifikationen, Wochenarbeitszeiten und allgemeiner Tag- bzw. Nachtschicht-Präferenz der Krankenschwestern einer Station für eine Woche.

- i sei der Index der Krankenschwestern.
- $q(i)$ sei die Qualifikationsstufe von Krankenschwester i.
- $w(i)$ sei die Wochenarbeitszeit von Krankenschwester i, wobei 1 = 100%, 2 = 80%, 4 = 50%, 5 = 40%, 6 = 20% und 7 = 0% (Urlaub oder Krankheit) einer Vollzeitstelle entspricht.
- $p(i)$ sei die allgemeine Tag- bzw. Nachtschichtpräferenz von Krankenschwester i, wobei 1 Tage werden bevorzugt, 2 Nächte werden bevorzugt und 0 Indifferenz bedeuten.

| i | 1 | 2 | 3 | 4 | 5 | 6 | 7 | 8 | 9 | 10 | 11 | 12 | 13 | 14 | 15 | 16 | 17 | 18 | 19 | 20 | 21 |
|------|---|---|---|---|---|---|---|---|---|----|----|----|----|----|----|----|----|----|----|----|----|
| q(i) | 1 | 1 | 1 | 1 | 1 | 1 | 1 | 1 | 2 | 2 | 2 | 2 | 2 | 3 | 3 | 3 | 3 | 3 | 3 | 3 | 3 |
| w(i) | 7 | 1 | 2 | 1 | 2 | 3 | 2 | 3 | 2 | 1 | 1 | 1 | 1 | 7 | 1 | 1 | 1 | 1 | 1 | 4 | 1 |
| p(i) | 1 | 1 | 0 | 0 | 0 | 2 | 2 | 0 | 0 | 0 | 2 | 0 | 0 | 0 | 0 | 0 | 0 | 0 | 0 | 0 | 1 |

**Tabelle 0-2:** Qualifikationen und Arbeitszeiten von Krankenschwestern einer Station.

## Anhang A.3 Wünsche der Krankenschwestern

Nachfolgend eine Tabelle mit den Arbeitswünschen der Krankenschwestern einer Station für eine Woche. Die Werte reichen von 0 (Indifferenz) bis 4 (starker Wunsch, nicht an diesem Tag bzw. dieser Nacht zu arbeiten).

- i sei der Index für die Krankenschwestern.
- So - Sa seien die sieben Tage bzw. Nächte der Woche.

| i | Tag-Schichten | | | | | | | Nacht-Schichten | | | | | | |
|---|----|----|----|----|----|----|----|----|----|----|----|----|----|----|
|   | So | Mo | Di | Mi | Do | Fr | Sa | So | Mo | Di | Mi | Do | Fr | Sa |
| 1 | 0 | 0 | 0 | 0 | 0 | 2 | 1 | 0 | 0 | 0 | 0 | 0 | 0 | 0 |
| 2 | 0 | 1 | 0 | 0 | 0 | 0 | 0 | 0 | 0 | 0 | 0 | 0 | 0 | 0 |
| 3 | 0 | 0 | 0 | 0 | 0 | 0 | 0 | 0 | 0 | 0 | 0 | 0 | 0 | 0 |
| 4 | 0 | 0 | 0 | 1 | 0 | 0 | 0 | 2 | 0 | 0 | 0 | 0 | 0 | 0 |
| 5 | 0 | 0 | 0 | 0 | 0 | 0 | 0 | 0 | 0 | 0 | 0 | 0 | 0 | 2 |
| 6 | 0 | 0 | 0 | 0 | 0 | 0 | 0 | 0 | 0 | 1 | 0 | 0 | 0 | 0 |
| 7 | 3 | 2 | 0 | 0 | 0 | 0 | 2 | 0 | 0 | 0 | 0 | 0 | 0 | 0 |
| 8 | 0 | 0 | 0 | 0 | 0 | 0 | 1 | 0 | 0 | 0 | 0 | 0 | 0 | 0 |
| 9 | 0 | 0 | 0 | 1 | 0 | 0 | 0 | 0 | 0 | 0 | 0 | 1 | 0 | 0 |
| 10 | 0 | 0 | 0 | 0 | 0 | 0 | 0 | 0 | 0 | 0 | 0 | 0 | 0 | 0 |
| 11 | 2 | 0 | 1 | 0 | 0 | 2 | 3 | 2 | 0 | 0 | 0 | 0 | 0 | 0 |
| 12 | 0 | 0 | 0 | 0 | 0 | 0 | 0 | 0 | 0 | 0 | 0 | 0 | 0 | 0 |
| 13 | 0 | 0 | 0 | 1 | 0 | 0 | 0 | 0 | 0 | 0 | 0 | 1 | 0 | 0 |
| 14 | 0 | 0 | 0 | 0 | 2 | 0 | 0 | 0 | 0 | 0 | 0 | 0 | 0 | 0 |
| 15 | 2 | 2 | 0 | 0 | 0 | 0 | 0 | 0 | 0 | 1 | 0 | 0 | 0 | 0 |
| 16 | 0 | 0 | 0 | 0 | 0 | 0 | 0 | 0 | 0 | 0 | 0 | 0 | 0 | 0 |
| 17 | 0 | 0 | 0 | 0 | 0 | 0 | 0 | 4 | 0 | 0 | 0 | 0 | 0 | 0 |
| 18 | 0 | 0 | 1 | 0 | 0 | 3 | 3 | 3 | 0 | 0 | 0 | 0 | 0 | 0 |
| 19 | 2 | 0 | 0 | 0 | 0 | 0 | 0 | 0 | 0 | 0 | 0 | 0 | 0 | 0 |
| 20 | 0 | 0 | 0 | 0 | 0 | 0 | 1 | 0 | 0 | 0 | 0 | 0 | 0 | 0 |
| 21 | 0 | 0 | 0 | 0 | 0 | 0 | 0 | 0 | 0 | 0 | 0 | 0 | 0 | 0 |

**Tabelle 0-3:** Arbeitswünsche von Krankenschwestern einer Station.

# Anhang A.4 Liste der Schichtmuster

Es folgt eine komplette Liste aller theoretisch möglicher Schichtmuster. Aus Platzgründen ist sie beispielhaft auf Vollzeit arbeitende Krankenschwestern beschränkt. Insgesamt gibt es 219 verschiedene Schichtmuster.

- i sei der Index der Schichtmuster.
- So - Sa sei 1, wenn bei Schichtmuster i an diesem Tag bzw. in dieser Nacht gearbeitet wird, sonst 0.
- $v(i)$ ist die von der Krankenhausleitung festgelegte allgemeine Güte von Schichtmuster i.
- $e(i)$ ist der erste freie, $l(i)$ der letzte freie Tag von Schichtmuster i. Diese Werte werden für eine Berücksichtigung des in der Vorwoche gearbeiteten Schichtmusters benötigt.
- $w(i)$ gibt an, zu welcher Wochenarbeitszeit in Prozent einer Vollzeitstelle Schichtmuster i gehört.

|   | Tag-Schichten | | | | | | | Nacht-Schichten | | | | | | | | | | |
|---|---|---|---|---|---|---|---|---|---|---|---|---|---|---|---|---|---|---|
| i | So | Mo | Di | Mi | Do | Fr | Sa | So | Mo | Di | Mi | Do | Fr | Sa | v(i) | e(i) | l(i) | w(i) |
| 1 | 1 | 1 | 1 | 1 | 1 | 0 | 0 | 0 | 0 | 0 | 0 | 0 | 0 | 0 | 1 | 6 | 7 | 100 |
| 2 | 1 | 1 | 1 | 1 | 0 | 1 | 0 | 0 | 0 | 0 | 0 | 0 | 0 | 0 | 2 | 5 | 7 | 100 |
| 3 | 1 | 1 | 1 | 1 | 0 | 0 | 1 | 0 | 0 | 0 | 0 | 0 | 0 | 0 | 2 | 5 | 6 | 100 |
| 4 | 1 | 1 | 1 | 0 | 1 | 1 | 0 | 0 | 0 | 0 | 0 | 0 | 0 | 0 | 2 | 4 | 7 | 100 |
| 5 | 1 | 1 | 1 | 0 | 1 | 0 | 1 | 0 | 0 | 0 | 0 | 0 | 0 | 0 | 3 | 4 | 6 | 100 |
| 6 | 1 | 1 | 1 | 0 | 0 | 1 | 1 | 0 | 0 | 0 | 0 | 0 | 0 | 0 | 2 | 4 | 5 | 100 |
| 7 | 1 | 1 | 0 | 1 | 1 | 1 | 0 | 0 | 0 | 0 | 0 | 0 | 0 | 0 | 2 | 3 | 7 | 100 |
| 8 | 1 | 1 | 0 | 1 | 1 | 0 | 1 | 0 | 0 | 0 | 0 | 0 | 0 | 0 | 3 | 3 | 6 | 100 |
| 9 | 1 | 1 | 0 | 1 | 0 | 1 | 1 | 0 | 0 | 0 | 0 | 0 | 0 | 0 | 3 | 3 | 5 | 100 |
| 10 | 1 | 1 | 0 | 0 | 1 | 1 | 1 | 0 | 0 | 0 | 0 | 0 | 0 | 0 | 2 | 2 | 4 | 100 |
| 11 | 1 | 0 | 1 | 1 | 1 | 1 | 0 | 0 | 0 | 0 | 0 | 0 | 0 | 0 | 2 | 2 | 7 | 100 |
| 12 | 1 | 0 | 1 | 1 | 1 | 0 | 1 | 0 | 0 | 0 | 0 | 0 | 0 | 0 | 3 | 2 | 6 | 100 |
| 13 | 1 | 0 | 1 | 1 | 0 | 1 | 1 | 0 | 0 | 0 | 0 | 0 | 0 | 0 | 3 | 2 | 5 | 100 |
| 14 | 1 | 0 | 1 | 0 | 1 | 1 | 1 | 0 | 0 | 0 | 0 | 0 | 0 | 0 | 3 | 2 | 4 | 100 |
| 15 | 1 | 0 | 0 | 1 | 1 | 1 | 1 | 0 | 0 | 0 | 0 | 0 | 0 | 0 | 2 | 2 | 3 | 100 |
| 16 | 0 | 1 | 1 | 1 | 1 | 1 | 0 | 0 | 0 | 0 | 0 | 0 | 0 | 0 | 1 | 1 | 7 | 100 |
| 17 | 0 | 1 | 1 | 1 | 1 | 0 | 1 | 0 | 0 | 0 | 0 | 0 | 0 | 0 | 2 | 1 | 6 | 100 |
| 18 | 0 | 1 | 1 | 1 | 0 | 1 | 1 | 0 | 0 | 0 | 0 | 0 | 0 | 0 | 2 | 1 | 5 | 100 |
| 19 | 0 | 1 | 1 | 0 | 1 | 1 | 1 | 0 | 0 | 0 | 0 | 0 | 0 | 0 | 2 | 1 | 4 | 100 |
| 20 | 0 | 1 | 0 | 1 | 1 | 1 | 1 | 0 | 0 | 0 | 0 | 0 | 0 | 0 | 2 | 1 | 3 | 100 |
| 21 | 0 | 0 | 1 | 1 | 1 | 1 | 1 | 0 | 0 | 0 | 0 | 0 | 0 | 0 | 1 | 1 | 2 | 100 |
| 22 | 0 | 0 | 0 | 0 | 0 | 0 | 0 | 1 | 1 | 1 | 1 | 0 | 0 | 0 | 1 | 5 | 7 | 100 |
| 23 | 0 | 0 | 0 | 0 | 0 | 0 | 0 | 1 | 1 | 1 | 0 | 1 | 0 | 0 | 2 | 4 | 7 | 100 |
| 24 | 0 | 0 | 0 | 0 | 0 | 0 | 0 | 1 | 1 | 1 | 0 | 0 | 1 | 0 | 2 | 4 | 7 | 100 |
| 25 | 0 | 0 | 0 | 0 | 0 | 0 | 0 | 1 | 1 | 1 | 0 | 0 | 0 | 1 | 2 | 4 | 6 | 100 |
| 26 | 0 | 0 | 0 | 0 | 0 | 0 | 0 | 1 | 1 | 0 | 1 | 1 | 0 | 0 | 2 | 3 | 7 | 100 |
| 27 | 0 | 0 | 0 | 0 | 0 | 0 | 0 | 1 | 1 | 0 | 1 | 0 | 1 | 0 | 3 | 3 | 7 | 100 |
| 28 | 0 | 0 | 0 | 0 | 0 | 0 | 0 | 1 | 1 | 0 | 1 | 0 | 0 | 1 | 3 | 3 | 6 | 100 |
| 29 | 0 | 0 | 0 | 0 | 0 | 0 | 0 | 1 | 1 | 0 | 0 | 1 | 1 | 0 | 2 | 3 | 7 | 100 |
| 30 | 0 | 0 | 0 | 0 | 0 | 0 | 0 | 1 | 1 | 0 | 0 | 1 | 0 | 1 | 3 | 3 | 6 | 100 |
| 31 | 0 | 0 | 0 | 0 | 0 | 0 | 0 | 1 | 1 | 0 | 0 | 0 | 1 | 1 | 2 | 3 | 5 | 100 |
| 32 | 0 | 0 | 0 | 0 | 0 | 0 | 0 | 1 | 0 | 1 | 1 | 1 | 0 | 0 | 2 | 2 | 7 | 100 |
| 33 | 0 | 0 | 0 | 0 | 0 | 0 | 0 | 1 | 0 | 1 | 1 | 0 | 1 | 0 | 3 | 2 | 7 | 100 |
| 34 | 0 | 0 | 0 | 0 | 0 | 0 | 0 | 1 | 0 | 1 | 1 | 0 | 0 | 1 | 3 | 2 | 6 | 100 |
| 35 | 0 | 0 | 0 | 0 | 0 | 0 | 0 | 1 | 0 | 1 | 0 | 1 | 1 | 0 | 3 | 2 | 7 | 100 |
| 36 | 0 | 0 | 0 | 0 | 0 | 0 | 0 | 1 | 0 | 1 | 0 | 1 | 0 | 1 | 4 | 2 | 6 | 100 |
| 37 | 0 | 0 | 0 | 0 | 0 | 0 | 0 | 1 | 0 | 1 | 0 | 0 | 1 | 1 | 3 | 2 | 5 | 100 |
| 38 | 0 | 0 | 0 | 0 | 0 | 0 | 0 | 1 | 0 | 0 | 1 | 1 | 1 | 0 | 2 | 2 | 7 | 100 |
| 39 | 0 | 0 | 0 | 0 | 0 | 0 | 0 | 1 | 0 | 0 | 1 | 1 | 0 | 1 | 3 | 2 | 6 | 100 |
| 40 | 0 | 0 | 0 | 0 | 0 | 0 | 0 | 1 | 0 | 0 | 1 | 0 | 1 | 1 | 3 | 2 | 5 | 100 |
| 41 | 0 | 0 | 0 | 0 | 0 | 0 | 0 | 1 | 0 | 0 | 0 | 1 | 1 | 1 | 2 | 2 | 4 | 100 |
| 42 | 0 | 0 | 0 | 0 | 0 | 0 | 0 | 0 | 1 | 1 | 1 | 1 | 0 | 0 | 1 | 1 | 7 | 100 |
| 43 | 0 | 0 | 0 | 0 | 0 | 0 | 0 | 0 | 1 | 1 | 1 | 0 | 1 | 0 | 2 | 1 | 7 | 100 |
| 44 | 0 | 0 | 0 | 0 | 0 | 0 | 0 | 0 | 1 | 1 | 1 | 0 | 0 | 1 | 2 | 1 | 7 | 100 |
| 45 | 0 | 0 | 0 | 0 | 0 | 0 | 0 | 0 | 1 | 1 | 0 | 1 | 1 | 0 | 2 | 1 | 7 | 100 |
| 46 | 0 | 0 | 0 | 0 | 0 | 0 | 0 | 0 | 1 | 1 | 0 | 1 | 0 | 1 | 3 | 1 | 6 | 100 |
| 47 | 0 | 0 | 0 | 0 | 0 | 0 | 0 | 0 | 1 | 1 | 0 | 0 | 1 | 1 | 2 | 1 | 5 | 100 |
| 48 | 0 | 0 | 0 | 0 | 0 | 0 | 0 | 0 | 1 | 0 | 1 | 1 | 1 | 0 | 2 | 1 | 7 | 100 |
| 49 | 0 | 0 | 0 | 0 | 0 | 0 | 0 | 0 | 1 | 0 | 1 | 1 | 0 | 1 | 3 | 1 | 6 | 100 |
| 50 | 0 | 0 | 0 | 0 | 0 | 0 | 0 | 0 | 1 | 0 | 1 | 0 | 1 | 1 | 3 | 1 | 5 | 100 |
| 51 | 0 | 0 | 0 | 0 | 0 | 0 | 0 | 0 | 1 | 0 | 0 | 1 | 1 | 1 | 2 | 1 | 4 | 100 |
| 52 | 0 | 0 | 0 | 0 | 0 | 0 | 0 | 0 | 0 | 1 | 1 | 1 | 1 | 0 | 1 | 1 | 7 | 100 |
| 53 | 0 | 0 | 0 | 0 | 0 | 0 | 0 | 0 | 0 | 1 | 1 | 1 | 0 | 1 | 2 | 1 | 6 | 100 |
| 54 | 0 | 0 | 0 | 0 | 0 | 0 | 0 | 0 | 0 | 1 | 1 | 0 | 1 | 1 | 2 | 1 | 5 | 100 |
| 55 | 0 | 0 | 0 | 0 | 0 | 0 | 0 | 0 | 0 | 1 | 0 | 1 | 1 | 1 | 2 | 1 | 4 | 100 |
| 56 | 0 | 0 | 0 | 0 | 0 | 0 | 0 | 0 | 0 | 0 | 1 | 1 | 1 | 1 | 1 | 1 | 3 | 100 |

**Tabelle 0-4:** Alle theoretisch möglicher Schichtmuster für eine Vollzeit arbeitende Krankenschwester.

# Anhang A.5 Strafkosten pro Schichtmuster

Es folgt eine Tabelle für die endgültigen Strafkosten einer Krankenschwester-Schichtmuster Kombination beispielhaft für zwei Krankenschwestern einer Station.

- Die jeweils erste Zeile ist der Index der Krankenschwester.
- Die jeweils zweite Zeile gibt den Bereich der erlaubten Schichtmuster an, wobei zuerst die untere und obere Schranke für Tag-Schichtmuster, dann die jeweiligen Schranken für Nacht-Schichtmuster angegeben werden.
- Alle folgenden Zeilen sind die Strafkosten der jeweiligen Krankenschwester-Schichtmuster-Kombination, zuerst für alle zulässigen Tag-Schichtmuster, dann für alle zulässigen Nacht-Schichtmuster.

| Krankenschwesterindex | 1 | | | | | | | | | | | | |
|---|---|---|---|---|---|---|---|---|---|---|---|---|---|
| Schichtmusterbereich | 1 | 21 | 22 | 56 | | | | | | | | | |
| Strafkosten | 0 | 2 | 0 | 1 | 2 | 0 | 1 | 1 | 2 | 0 | 2 | 2 | 2 |
| | 2 | 1 | 0 | 1 | 1 | 1 | 2 | 0 | 18 | 20 | 20 | 20 | 19 |
| | 19 | 20 | 20 | 19 | 20 | 19 | 20 | 20 | 20 | 20 | 21 | 20 | 19 |
| | 20 | 20 | 18 | 18 | 20 | 20 | 19 | 20 | 19 | 20 | 20 | 20 | 19 |
| | 2 | 19 | 18 | 20 | | | | | | | | | |
| Krankenschwesterindex | 2 | | | | | | | | | | | | |
| Schichtmusterbereich | 57 | 91 | 92 | 126 | | | | | | | | | |
| Strafkosten | 16 | 10 | 10 | 9 | 9 | 10 | 10 | 9 | 10 | 9 | 10 | 10 | 9 |
| | 10 | 11 | 10 | 9 | 9 | 10 | 9 | 8 | 10 | 9 | 9 | 10 | 9 |
| | 6 | 10 | 10 | 10 | 10 | 8 | 9 | 9 | 9 | 8 | 0 | 5 | 5 |
| | 6 | 0 | 6 | 7 | 7 | 6 | 7 | 7 | 5 | 6 | 5 | 0 | 6 |
| | 6 | 6 | 6 | 6 | 5 | 6 | 6 | 6 | 0 | 7 | 7 | 7 | 6 |
| | 7 | 7 | 6 | 5 | 0 | | | | | | | | |

**Tabelle 0-5:** Endgültige Krankenschwester-Schichtmuster Strafkosten.

# Anhang B Turbo Pascal Listing

Es folgt das Listing der Hauptroutine des Turbo Pascal Programmes zur Lösung des Nurse Scheduling Problem mittels genetische Algorithmen. Das vollständige Programm ist ca. 100 KByte groß.

```
begin {main}

getinput (mutarate,niching,average,seed,reinit,original,smart,sevendays,test,elitism,dynamic,
parents,substitution,loop1,loop2,swapnumber,crossing,week,weights,population,dweight1,
pweight1,dweight2,pweight2,generations,outfilename);

getfilename (week,original,filename,demandfilename,qualfilename,preferfilename,shiftfilename,
solutionfilename,previousfilename);

getdemand (demandfilename,daydemand,totaldemand);

getshiftpattern (shiftfilename,shiftpattern,shiftpatternvalue,firstdayoff,lastdayoff,daynightshift);

getoldsolution (oldsolutionprefer,oldsolutionschedule,oldnurseshiftvalue,week,solutionfilename,
previousfilename);

getqual (qualfilename,nursenumber,qualification,hours,daynight);

getprefer (nursenumber,preferfilename,prefer);

possibleshifts (sevendays,nursenumber,oldsolutionschedule,hours,daynight,daynightshift,
firstdayoff,loday,hiday,lonight,hinight);

getnurseshiftvalue (sevendays,nursenumber,oldnurseshiftvalue,loday,hiday,lonight,hinight,
prefer,shiftpattern,shiftpatternvalue,nurseshiftvalue);

mark (heapstart);

for pop := 1 to population do
        begin
                individual[pop] := nil;
                newindividual[pop] := nil;
        end;

if (seed) then randseed := 543210 + runs else randomize;

writeln ('To stop the algorithm press  > space <  anytime ');

dweight         := dweight1;
pweight         := pweight1;
total           := 0;
delta           := 0;
solutionprefer  := 255;

getpops (startpop,endpop,population,niching);
for pop:= 1 to population do
```

```
            begin
                    new (individual[pop]);
                    new (newindividual[pop]);
                    randomstart (individual,nursenumber,loday,hiday,lonight,hinight,daynight,
                    pop,niching,startpop,endpop);
                    newindividual[pop]^.tribe := individual[pop]^.tribe;
            end;

demandvaluefunc (individual,totaldemand,nursenumber,point1,point2,shiftpattern,daydemand,
population);

prefervaluefunc (individual,nurseshiftvalue,nursenumber,population);

repeat

inc (total);

if (weights > 1) then setweights (dweight,pweight,dweight1,pweight1,dweight2,pweight2,
counter1,counter2,loop1,loop2,weights,feasible);

ranking (individual,value,nursenumber,dweight,pweight,best,startpop,endpop,population,niching);

if (average) then averagecalculation (averagedemand,averageprefer,individual,best,startpop,
endpop,population,niching);

crossover (mutarate,individual,newindividual,smart,reinit,niching,nursenumber,parents,
crossing,point1,point2,population,startpop,endpop);

if (niching) then nicheswap (newindividual,best);

demandvaluefunc (newindividual,totaldemand,nursenumber,point1,point2,shiftpattern,daydemand,
population);

if (swapnumber > 0) then internalswap (newindividual,population,nurseshiftvalue,qualification,
daynight,hours,nursenumber,swapnumber,swapchance);

prefervaluefunc (newindividual,nurseshiftvalue,nursenumber,population);

swapp (individual,newindividual,dweight,pweight,population,substitution,nursenumber,
delta,elitism,niching,value,best,startpop,endpop);

solution (niching,feasible,individual,nursenumber,population,solutionschedule,solutionprefer,
oldsolutionprefer,startpop);

if (dynamic) and (population > minpop) then population := (999 * population) div 1000;

until (delta >= generations) or (keypressed);

release (heapstart);

if (not original) then transform (original,nursenumber,holidaynumber,solutionschedule,
holidaynurse,holidayschedule,nurseshiftvalue);

finaloutput (original,nursenumber,holidaynumber,total,outfilename,solutionschedule,
nurseshiftvalue,solutionprefer);

end.
```

# Anhang C CPLEX Eingabefile

Es folgt ein beispielhaftes CPLEX Eingabefile für 5 Krankenschwestern und zwei Tage. Ein vollständiges Eingabefile für ca. 20 Krankenschwestern und sieben Tage bzw. Nächte ist je nach Datenkonstellation zwischen 100 KByte und 150 KByte groß.

minimize

4 x1,1 + 14 x1,2 + 14 x1,3 + 14 x1,4 + 21 x1,5 + 20 x1,6 + 14 x1,7 + 21 x1,8
+ 30 x1,9 + 20 x1,10 + 12 x1,11 + 18 x1,12 + 27 x1,13 + 27 x1,14 + 18 x1,15
+ 5 x1,16 + 10 x1,17 + 16 x1,18 + 16 x1,19 + 16 x1,20 + 7 x1,21 + 5 x2,1
+ 12 x2,2 + 10 x2,3 + 4 x2,4 + 3 x2,5 + 4 x2,6 + 12 x2,7 + 15 x2,8 + 18 x2,9
+ 4 x2,10 + 12 x2,11 + 15 x2,12 + 18 x2,13 + 6 x2,14 + 12 x2,15 + 6 x2,16
+ 10 x2,17 + 12 x2,18 + 4 x2,19 + 12 x2,20 + 6 x2,21 + 7 x3,1 + 10 x3,2
+ 16 x3,3 + 10 x3,4 + 24 x3,5 + 12 x3,6 + 10 x3,7 + 24 x3,8 + 18 x3,9
+ 12 x3,10 + 14 x3,11 + 30 x3,12 + 24 x3,13 + 24 x3,14 + 16 x3,15 + 7 x3,16
+ 20 x3,17 + 16 x3,18 + 16 x3,19 + 16 x3,20 + 10 x3,21 + 1 x4,1 + 2 x4,2
+ 2 x4,3 + 2 x4,4 + 3 x4,5 + 2 x4,6 + 2 x4,7 + 3 x4,8 + 3 x4,9 + 2 x4,10
+ 2 x4,11 + 3 x4,12 + 3 x4,13 + 3 x4,14 + 2 x4,15 + 1 x4,16 + 2 x4,17 + 2 x4,18
+ 2 x4,19 + 2 x4,20 + 1 x4,21 + 1 x5,1 + 2 x5,2 + 2 x5,3 + 2 x5,4 + 3 x5,5
+ 2 x5,6 + 2 x5,7 + 3 x5,8 + 3 x5,9 + 2 x5,10 + 2 x5,11 + 3 x5,12 + 3 x5,13
+ 3 x5,14 + 2 x5,15 + 1 x5,16 + 2 x5,17 + 2 x5,18 + 2 x5,19 + 2 x5,20 + 1 x5,21

subject to

x1,1 + x1,2 + x1,3 + x1,4 + x1,5 + x1,6 + x1,7 + x1,8 + x1,9 + x1,10 + x1,11
+ x1,12 + x1,13 + x1,14 + x1,15 + x2,1 + x2,2 + x2,3 + x2,4 + x2,5 + x2,6
+ x2,7 + x2,8 + x2,9 + x2,10 + x2,11 + x2,12 + x2,13 + x2,14 + x2,15 + x3,1
+ x3,2 + x3,3 + x3,4 + x3,5 + x3,6 + x3,7 + x3,8 + x3,9 + x3,10 + x3,11
+ x3,12 + x3,13 + x3,14 + x3,15 >= 1

x1,1 + x1,2 + x1,3 + x1,4 + x1,5 + x1,6 + x1,7 + x1,8 + x1,9 + x1,10 + x1,11
+ x1,12 + x1,13 + x1,14 + x1,15 + x2,1 + x2,2 + x2,3 + x2,4 + x2,5 + x2,6
+ x2,7 + x2,8 + x2,9 + x2,10 + x2,11 + x2,12 + x2,13 + x2,14 + x2,15 + x3,1
+ x3,2 + x3,3 + x3,4 + x3,5 + x3,6 + x3,7 + x3,8 + x3,9 + x3,10 + x3,11
+ x3,12 + x3,13 + x3,14 + x3,15 + x4,1 + x4,2 + x4,3 + x4,4 + x4,5 + x4,6
+ x4,7 + x4,8 + x4,9 + x4,10 + x4,11 + x4,12 + x4,13 + x4,14 + x4,15 >= 2

x1,1 + x1,2 + x1,3 + x1,4 + x1,5 + x1,6 + x1,7 + x1,8 + x1,9 + x1,10 + x1,11
+ x1,12 + x1,13 + x1,14 + x1,15 + x2,1 + x2,2 + x2,3 + x2,4 + x2,5 + x2,6
+ x2,7 + x2,8 + x2,9 + x2,10 + x2,11 + x2,12 + x2,13 + x2,14 + x2,15 + x3,1
+ x3,2 + x3,3 + x3,4 + x3,5 + x3,6 + x3,7 + x3,8 + x3,9 + x3,10 + x3,11
+ x3,12 + x3,13 + x3,14 + x3,15 + x4,1 + x4,2 + x4,3 + x4,4 + x4,5 + x4,6
+ x4,7 + x4,8 + x4,9 + x4,10 + x4,11 + x4,12 + x4,13 + x4,14 + x4,15 + x5,1
+ x5,2 + x5,3 + x5,4 + x5,5 + x5,6 + x5,7 + x5,8 + x5,9 + x5,10 + x5,11
+ x5,12 + x5,13 + x5,14 + x5,15 >= 3

x1,1 + x1,2 + x1,3 + x1,4 + x1,5 + x1,6 + x1,7 + x1,8 + x1,9 + x1,10 + x1,16
+ x1,17 + x1,18 + x1,19 + x1,20 + x2,1 + x2,2 + x2,3 + x2,4 + x2,5 + x2,6
+ x2,7 + x2,8 + x2,9 + x2,10 + x2,16 + x2,17 + x2,18 + x2,19 + x2,20 + x3,1
+ x3,2 + x3,3 + x3,4 + x3,5 + x3,6 + x3,7 + x3,8 + x3,9 + x3,10 + x3,16 + x3,17

+ x3,18 + x3,19 + x3,20 >= 1

x1,1 + x1,2 + x1,3 + x1,4 + x1,5 + x1,6 + x1,7 + x1,8 + x1,9 + x1,10 + x1,16
+ x1,17 + x1,18 + x1,19 + x1,20 + x2,1 + x2,2 + x2,3 + x2,4 + x2,5 + x2,6
+ x2,7 + x2,8 + x2,9 + x2,10 + x2,16 + x2,17 + x2,18 + x2,19 + x2,20 + x3,1
+ x3,2 + x3,3 + x3,4 + x3,5 + x3,6 + x3,7 + x3,8 + x3,9 + x3,10 + x3,16
+ x3,17 + x3,18 + x3,19 + x3,20 + x4,1 + x4,2 + x4,3 + x4,4 + x4,5 + x4,6
+ x4,7 + x4,8 + x4,9 + x4,10 + x4,16 + x4,17 + x4,18 + x4,19 + x4,20 >= 2

x1,1 + x1,2 + x1,3 + x1,4 + x1,5 + x1,6 + x1,7 + x1,8 + x1,9 + x1,10 + x1,16
+ x1,17 + x1,18 + x1,19 + x1,20 + x2,1 + x2,2 + x2,3 + x2,4 + x2,5 + x2,6
+ x2,7 + x2,8 + x2,9 + x2,10 + x2,16 + x2,17 + x2,18 + x2,19 + x2,20 + x3,1
+ x3,2 + x3,3 + x3,4 + x3,5 + x3,6 + x3,7 + x3,8 + x3,9 + x3,10 + x3,16 + x3,17
+ x3,18 + x3,19 + x3,20 + x4,1 + x4,2 + x4,3 + x4,4 + x4,5 + x4,6 + x4,7 + x4,8
+ x4,9 + x4,10 + x4,16 + x4,17 + x4,18 + x4,19 + x4,20 + x5,1 + x5,2 + x5,3
+ x5,4 + x5,5 + x5,6 + x5,7 + x5,8 + x5,9 + x5,10 + x5,16 + x5,17 + x5,18
+ x5,19 + x5,20 >= 3

x1,1 + x1,2 + x1,3 + x1,4 + x1,5 + x1,6 + x1,7 + x1,8 + x1,9 + x1,10 + x1,11
+ x1,12 + x1,13 + x1,14 + x1,15 + x1,16 + x1,17 + x1,18 + x1,19 + x1,20
+ x1,21 = 1

x2,1 + x2,2 + x2,3 + x2,4 + x2,5 + x2,6 + x2,7 + x2,8 + x2,9 + x2,10 + x2,11
+ x2,12 + x2,13 + x2,14 + x2,15 + x2,16 + x2,17 + x2,18 + x2,19 + x2,20
+ x2,21 = 1

x3,1 + x3,2 + x3,3 + x3,4 + x3,5 + x3,6 + x3,7 + x3,8 + x3,9 + x3,10 + x3,11
+ x3,12 + x3,13 + x3,14 + x3,15 + x3,16 + x3,17 + x3,18 + x3,19 + x3,20
+ x3,21 = 1

x4,1 + x4,2 + x4,3 + x4,4 + x4,5 + x4,6 + x4,7 + x4,8 + x4,9 + x4,10 + x4,11
+ x4,12 + x4,13 + x4,14 + x4,15 + x4,16 + x4,17 + x4,18 + x4,19 + x4,20
+ x4,21 = 1

x5,1 + x5,2 + x5,3 + x5,4 + x5,5 + x5,6 + x5,7 + x5,8 + x5,9 + x5,10 + x5,11
+ x5,12 + x5,13 + x5,14 + x5,15 + x5,16 + x5,17 + x5,18 + x5,19 + x5,20
+ x5,21 = 1

integers

x1,1 x1,2 x1,3 x1,4 x1,5 x1,6 x1,7 x1,8 x1,9 x1,10 x1,11 x1,12 x1,13 x1,14
x1,15 x1,16 x1,17 x1,18 x1,19 x1,20 x1,21

x2,1 x2,2 x2,3 x2,4 x2,5 x2,6 x2,7 x2,8 x2,9 x2,10 x2,11 x2,12 x2,13 x2,14
x2,15 x2,16 x2,17 x2,18 x2,19 x2,20 x2,21

x3,1 x3,2 x3,3 x3,4 x3,5 x3,6 x3,7 x3,8 x3,9 x3,10 x3,11 x3,12 x3,13 x3,14
x3,15 x3,16 x3,17 x3,18 x3,19 x3,20 x3,21

x4,1 x4,2 x4,3 x4,4 x4,5 x4,6 x4,7 x4,8 x4,9 x4,10 x4,11 x4,12 x4,13 x4,14
x4,15 x4,16 x4,17 x4,18 x4,19 x4,20 x4,21

x5,1 x5,2 x5,3 x5,4 x5,5 x5,6 x5,7 x5,8 x5,9 x5,10 x5,11 x5,12 x5,13 x5,14
x5,15 x5,16 x5,17 x5,18 x5,19 x5,20 x5,21

end

# Ehrenwörtliche Erklärung

Ich versichere hiermit, daß ich die beiliegende Diplomarbeit ohne Hilfe Dritter und ohne Benutzung anderer als der angegebenen Quellen und Hilfsmittel angefertigt und die den benutzen Quellen wörtlich oder inhaltlich entnommen Stellen als solche kenntlich gemacht habe. Diese Arbeit hat in gleicher oder ähnlicher Form noch keiner Prüfungsbehörde vorgelegen.

Heilbronn, den 19. Dezember 1996